\documentclass[runningheads,pdftex]{llncs}
\usepackage{graphicx}
\usepackage{amsmath,amssymb} 
\usepackage{color}
\usepackage[width=122mm,left=12mm,paperwidth=146mm,height=193mm,top=12mm,paperheight=217mm]{geometry}
\usepackage{arydshln}
\usepackage{pdflscape}
\usepackage{multirow}
\usepackage{url}
\usepackage{rotating}

\usepackage[labelformat=simple]{subcaption}
\graphicspath{{figure/}}

\newcommand{\lw}[1]{\smash{\lower2.0ex\hbox{#1}}}
\newcommand{\lwm}[1]{\smash{\lower1.5ex\hbox{#1}}}
\newcommand{\lwh}[1]{\smash{\lower1.0ex\hbox{#1}}}
\newcommand{\lwmm}[1]{\smash{\lower-1.5ex\hbox{#1}}}

\begin{document}
\pagestyle{headings}
\mainmatter
\def\ECCV16SubNumber{IWRR-14}  

\title{Advances of Scene Text Datasets} 

\titlerunning{Advances of Scene Text Datasets}

\authorrunning{M. Iwamura}

\author{Masakazu Iwamura}
\institute{Department of Computer Science and Intelligent Systems\\
  Graduate School of Engineering, Osaka Prefecture University\\
\email{masa@cs.osakafu-u.ac.jp} }

\maketitle

\begin{abstract}
This article introduces publicly available datasets in scene text detection and recognition. The information is as of 2017.

\end{abstract}

\section{Introduction}
\label{sec:intro}

Advances in pattern recognition and computer vision researches
are often brought by advances in both techniques and datasets;
a new technique requires a new dataset to prove its effectiveness, and
a new dataset motivates researches to develop new techniques.
In the research field of scene text detection and recognition, it is also true.
Particularly in the field,
representative datasets have been provided through competitions held in conjunction with the series of International Conference on Document Analysis and Recognition (ICDAR).
But, not limited to them, various datasets have been released.
This article focuses on these publicly available datasets in scene text detection and recognition and gives an overview.

\subsection{Roles of datasets}
The most important role of datasets is to well represent the recognition targets as they are (which is often referred as ``in the wild'').
Due to a variety of looks of recognition targets, large datasets are generally desired.
In the era of deep learning, demand for larger training datasets is stronger.
However, constructing a large dataset is not an easy task due to large cost in labor and money.
Hence, there is a gap between ideal and real.
As a workaround, data synthesis has been considered a very useful and important technique.
Effectiveness of data synthesis in scene text detection and recognition is shown in \cite{Jaderberg_NIPS_DLWorkshop2014,Gupta_CVPR2016}.
However, use of datasets containing synthesized data for evaluation is arguable because synthesized data are considered not to completely represent the nature of real recognition targets.

Another important role of datasets is to provide an opportunity to fairly and easily compare techniques.
In the research field, datasets provided for the series of ICDAR Robust Reading Competition (RRC) and some other datasets are often used.
Only with an experiment of the proposed method following the protocol and evaluation criterion determined for the selected dataset and task, a proposed method can be fairly compared with the state-of-the-art methods.
Hence, publicly available datasets contribute to encourage development of new methods.


\subsection{Tasks and Evaluation}
\begin{figure}[tb]
  \centering
  \includegraphics[width=\hsize]{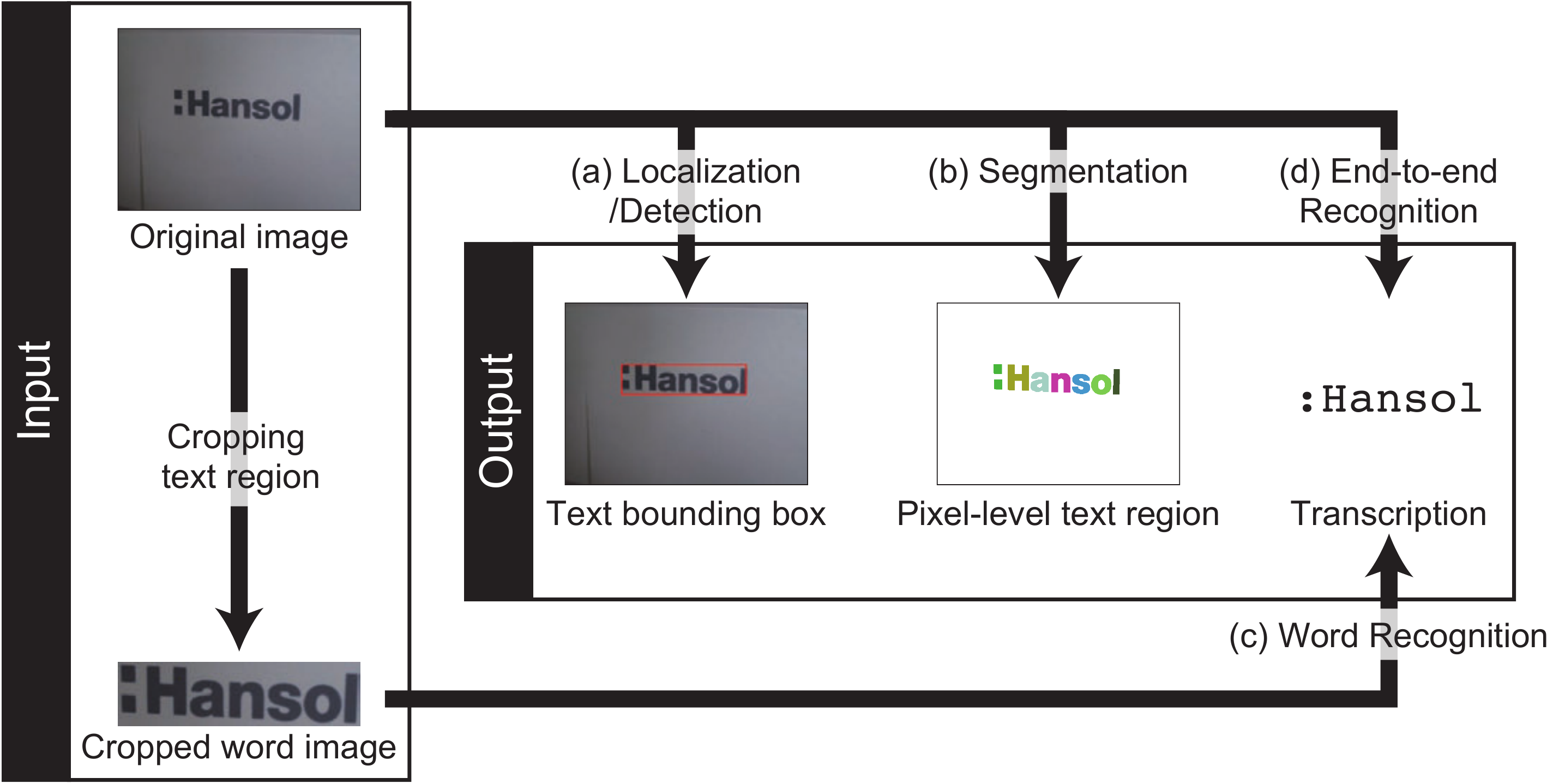}
  \caption{Tasks of scene text detection and recognition.}
  \label{fig:tasks}
\end{figure}

Four tasks are generally considered in the research field of scene text detection and recognition.
See Fig.~\ref{fig:tasks} for illustration of the tasks.
Typical evaluation criteria of the tasks can be found in
\cite{ICDAR2013_RRC,ICDAR2015_RRC}.

\renewcommand\theenumi{(\alph{enumi})}
\begin{enumerate}
\item Text Localization/Detection\\
  This task requires to output text regions of a given image in the form of \textit{bounding boxes}.
  Usually the bounding boxes are expected to be as tight to the detected text as possible.
  For evaluation of static images, a standard precision and recall metric~\cite{ICDAR_RRC2003,ICDAR_RRC2003_IJDAR2005,ICDAR_RRC2005},      
  DetEval~\cite{Wolf_IJDAR2006}\footnote{ICDAR Robust Reading Competition ``Born Digital Images'' and ``Focused Scene Text'' use a slightly different implementation from the original {\scriptsize (\url{http://liris.cnrs.fr/christian.wolf/software/deteval/})}. See more detail at {\scriptsize \url{http://rrc.cvc.uab.es/?com=faq}}.} and
  intersection-over-union (IoU) overlap method~\cite{Everingham_IJCV2014} are used.
  For evaluation of videos, CLEAR-MOT~\cite{Bernardin_EJIVP2008} and VACE~\cite{Kasturi_TPAMI2009} are used in ICDAR RRC ``Text in Videos''~\cite{ICDAR2013_RRC,ICDAR2015_RRC}.
  In addition, ``video precision and recall'' is proposed in \cite{Nguyen_WACV2014}.

\item Text Segmentation\\
  This task requires to output text regions of a given image by \textit{pixels}.
  For evaluation,
  a standard pixel-level precision and recall metric is used in \cite{ICDAR2011_RRC_challenge1,Jung_ETRI2011} and
  an atom-based metric~\cite{Clavelli_DAS2010}
  is used in ICDAR RRC ``Born Digital Images''~\cite{ICDAR2011_RRC_challenge1,ICDAR2013_RRC} and ``Focused Scene Text''~\cite{ICDAR2013_RRC}.
  
\item (Cropped) Word Recognition\\
  This task requires to output the transcription of a given cropped word image.
  For evaluation, recognition accuracy and a standard edit distance metric are often used~\cite{Wang_ICCV2011,ICDAR2013_RRC}.
  Sometimes case is ignored.
  
\item End-to-end Recognition\\
  This task requires to output the transcriptions of text regions of a given image.
  The result is evaluated by the same way as the localization task first and then wrongly recognized words are excluded~\cite{ICDAR2015_RRC}.
  
\end{enumerate}
\renewcommand\theenumi{\arabic{enumi}}


\begin{sidewaystable}[p]
  \centering
  \caption{Summary of publicly available datasets.
    \#Image represents the total number of images, mostly
    of detection tasks
    (for a video dataset, the total number of frames).
    \#Word represents the number of word regions ground truthed.
    Tasks indicate
    Text Localization/Detection (L), Text Segmentation (S), Word Recognition (R) and End-to-end Recognition (E).
    \#WS represents the number of word sequences in a video dataset.
  } 
  \label{tbl:db_summary}

  \vspace*{-3mm}
  \begin{tabular}{p{3mm}|c|c|r|r|l|l|p{88mm}}
    \hline
    \multicolumn{3}{c|}{Name} & \multicolumn{1}{c|}{\#Image} &
    \multicolumn{1}{c|}{\#Word} & Languages & Tasks & Note \\ \hline
    \multirow{18}{*}{\rotatebox[origin=c]{-90}{\scriptsize ICDAR Robust Reading Competition / Challenge}} & \multicolumn{2}{c|}{2003~\cite{ICDAR_RRC2003,ICDAR_RRC2003_IJDAR2005}, 2005~\cite{ICDAR_RRC2005}} & 529 & 2,434 & Eng. & LR & \\ \cline{2-8}
    %
    & Born & 2011~\cite{ICDAR2011_RRC_challenge1} & 522 & 4,501 & & \lwm{LSR} & \\ \cline{3-5}
    & Digital & 2013~\cite{ICDAR2013_RRC} & \lwm{561} & \lwm{5,003} & Eng. & & \\ \cline{3-3} \cline{7-7}
    & Images & 2015~\cite{ICDAR2015_RRC} &  & & & E & \\ \cline{2-8}
    %
    %
    & Focused & 2011~\cite{ICDAR2011_RRC_challenge2} & 484 & 2,037 & & LR & \\ \cline{3-5} \cline{7-7}
    & Scene & 2013~\cite{ICDAR2013_RRC} & \lwm{462} & \lwm{2,524} & Eng. & LSR &\\ \cline{3-3} \cline{7-7}
    & Text & 2015~\cite{ICDAR2015_RRC} & & & & E & \\ \cline{2-8}
    %
    & Text in & 2013~\cite{ICDAR2013_RRC} & 15,277 & 93,598 & \lwm{Eng., Fre., Spa.} & L & Video (\#WS=1,962).\\ \cline{3-5} \cline{7-8}
    &  Videos & 2015~\cite{ICDAR2015_RRC} & 27,824 & 125,141 & & LE & Video (\#WS=3,562).\\ \cline{2-8}
    %
    & \multicolumn{2}{c|}{2015 Incidental} & \lwm{1,670} & \lwm{17,548} & \lwm{Eng.} & \lwm{LRE} \\
    & \multicolumn{2}{c|}{Scene Text~\cite{ICDAR2015_RRC}} & & & & \\ \cline{2-8}
    %
    & \multicolumn{2}{c|}{2017} & \lwm{63,686} & \lwm{173,589} & Eng., Ger., Fre., & \lwm{LRE} & \\
    & \multicolumn{2}{c|}{COCO-Text~\cite{Veit_arXiv2016,ICDAR_COCO2017}} & & & Spa., etc. & & \lwmm{Text annotation of MS COCO Dataset~\cite{Lin_arXiv2014}.} \\ \cline{2-8}
    %
    & \multicolumn{2}{c|}{2017 FSNS~\cite{Smith_IWRR2016}} & 1,081,422 & \multicolumn{1}{c|}{-} & Fre. & E & Each image contains up to 4 views of a street name sign.\\ \cline{2-8}
    %
    & \multicolumn{2}{c|}{2017 DOST~\cite{Iwamura_IWRR2016,ICDAR_DOST2017}} & 32,147 & 797,919 & Jap., etc. & LRE & Video (\#WS=22,398). 5 views in most frames. \\ \cline{2-8}
    %
    & \multicolumn{2}{c|}{} & & & Ara., Ban., Chi., & & \lwm{Tasks also include script identification.} \\
    & \multicolumn{2}{c|}{2017 MLT~\cite{ICDAR_MLT2017}} & 18,000 & 107,547& Eng., Fre., Ger., & LR & \lwm{\#Word counts training and validation sets.} \\
    & \multicolumn{2}{c|}{} & & & Ita., Jap., Kor. & & \\ \hline \hline
    %
    %
    \multirow{11}{*}{\rotatebox[origin=c]{-90}{\scriptsize General}} & \multicolumn{2}{c|}{\lwm{Chars74k~\cite{deCampos_VISAPP2009}}} & \lwm{74,107} & \lwm{74,107} & Eng.,& \lwm{R} & Character image DB (natural, hand drawn and synthesised). \\
    & \multicolumn{2}{c|}{}  & & & Kannada & & \#Word represents the number of English characters. \\ \cline{2-8}
    %
    & \multicolumn{2}{c|}{SVT~\cite{Wang_ECCV2010,Wang_ICCV2011}} & 349 & 904 & Eng. & LRE & \\ \cline{2-8}
    %
    & \multicolumn{2}{c|}{NEOCR~\cite{Nagy_CBDAR2012}} & 659 & 5,238 & Eng., Ger. & LR & Text with various degradation (blur, perspective distortion+). \\ \cline{2-8}
    %
    & \multicolumn{2}{c|}{KAIST~\cite{Jung_ETRI2011}} & 3,000 & 3,000 & Eng., Kor. & LS & \\ \cline{2-8}
    %
    & \multicolumn{2}{c|}{SVHN~\cite{svhn}} & 248,823 & 630,420 & Digit & LR & Digit image DB. \#Word represents the number of digits. \\ \cline{2-8}
    %
    & \multicolumn{2}{c|}{MSRA-TD500~\cite{Yao_CVPR2012}} & 500 & 500 & Eng., Chi. & L & Text bounding boxes are in various angles. \\ \cline{2-8}
    %
    & \multicolumn{2}{c|}{IIIT5K~\cite{Mishra_BMVC2012}} & 5,000 & 5,000 & Eng. & R & Cropped word image DB. \\ \cline{2-8}
    %
    & \multicolumn{2}{c|}{YouTube Video Text~\cite{Nguyen_WACV2014}} & 11,791 & 16,620 & Eng. & LR & Videos from YouTube (\#WS=245).\\ \cline{2-8}
    & \multicolumn{2}{c|}{ICDAR2015 TRW~\cite{ICDAR_TRW2015}} & 1,271 & 6,291 & Eng., Chi. & LR & \\ \cline{2-8}
    %
    & \multicolumn{2}{c|}{ICDAR2017 RCTW~\cite{ICDAR_RCTW17}} & 12,263 & 64,248 & Chi. & LE & \#Word counts training data. \\ \hline \hline
    %
    \multirow{2}{*}{\rotatebox[origin=c]{-90}{\scriptsize Synth}} & \multicolumn{2}{c|}{MJSynth~\cite{Jaderberg_NIPS_DLWorkshop2014}} & 8,919,273 & 8,919,273 & Eng. & \multicolumn{1}{c|}{-} & Synthesized cropped word image DB. \\ \cline{2-8}
    %
    & \multicolumn{2}{c|}{SynthText~\cite{Gupta_CVPR2016}} & 800,000 & 800,000 & Eng. & \multicolumn{1}{c|}{-} & Synthesized scene text image DB. \\ \hline
  \end{tabular}
\end{sidewaystable}

\begin{figure}[tbp]
  \centering

  \begin{minipage}[b]{\hsize}
    \centering    
    \includegraphics[width=.115\hsize]{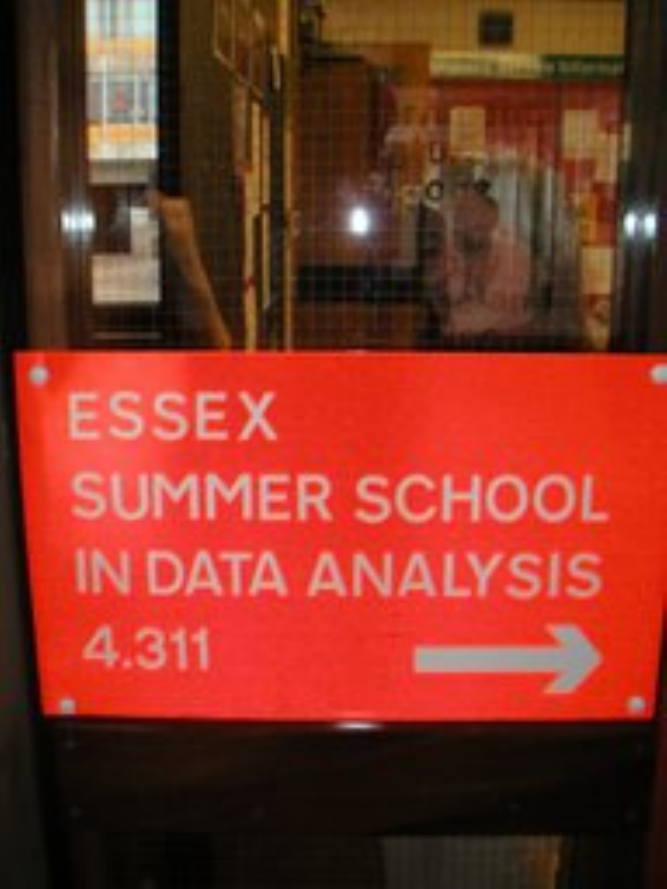}
    \includegraphics[width=.115\hsize]{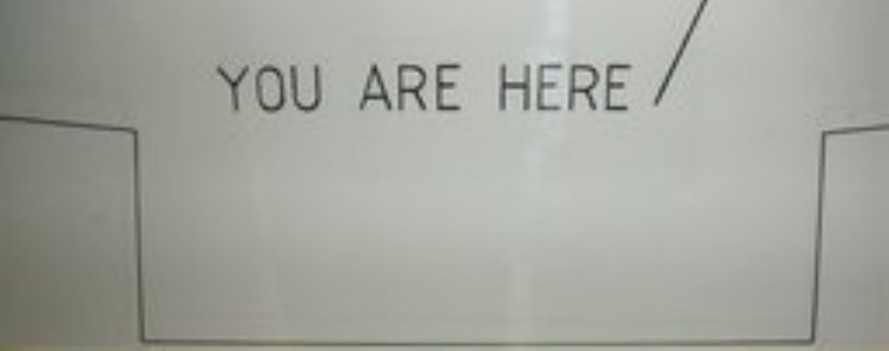}
    \includegraphics[width=.115\hsize]{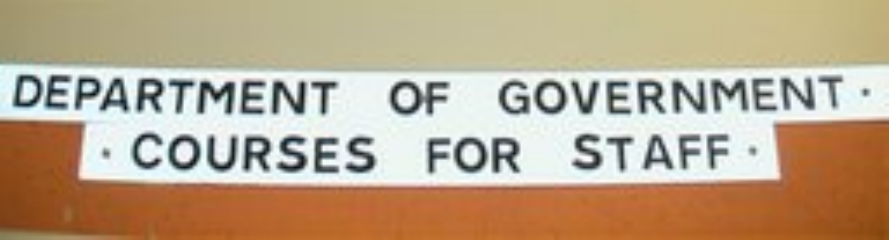}
    \includegraphics[width=.115\hsize]{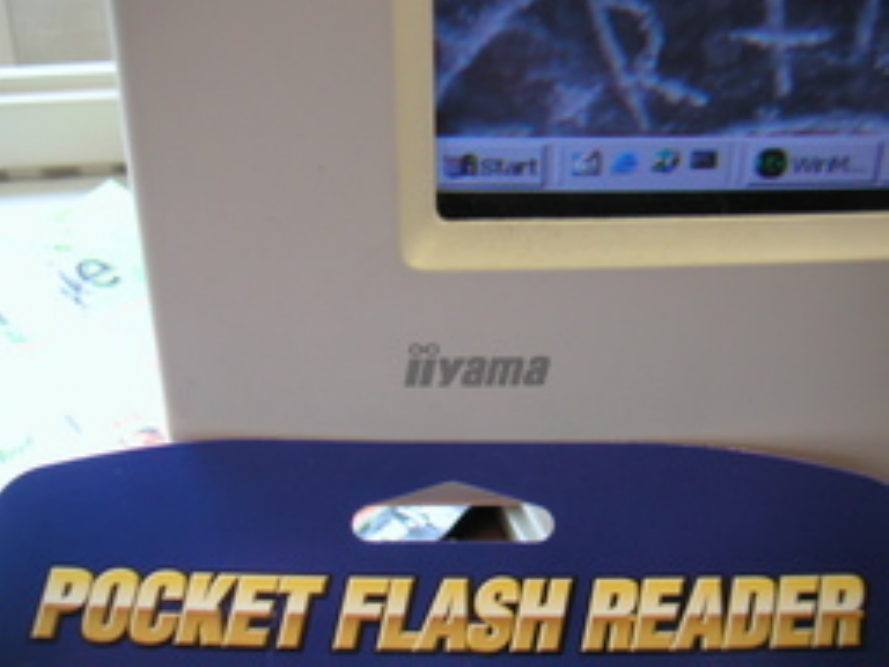}
    \includegraphics[width=.115\hsize]{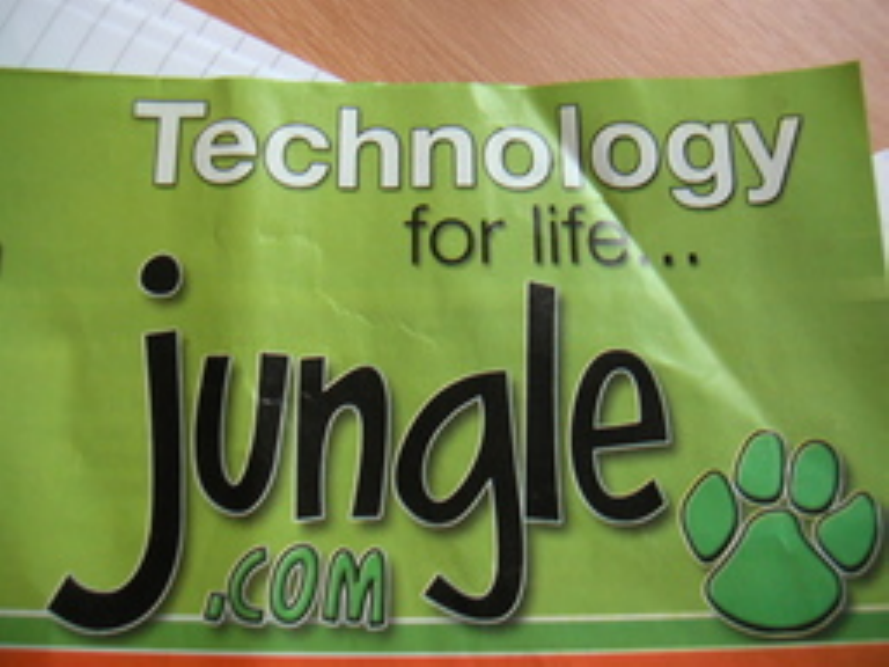}
    \includegraphics[width=.115\hsize]{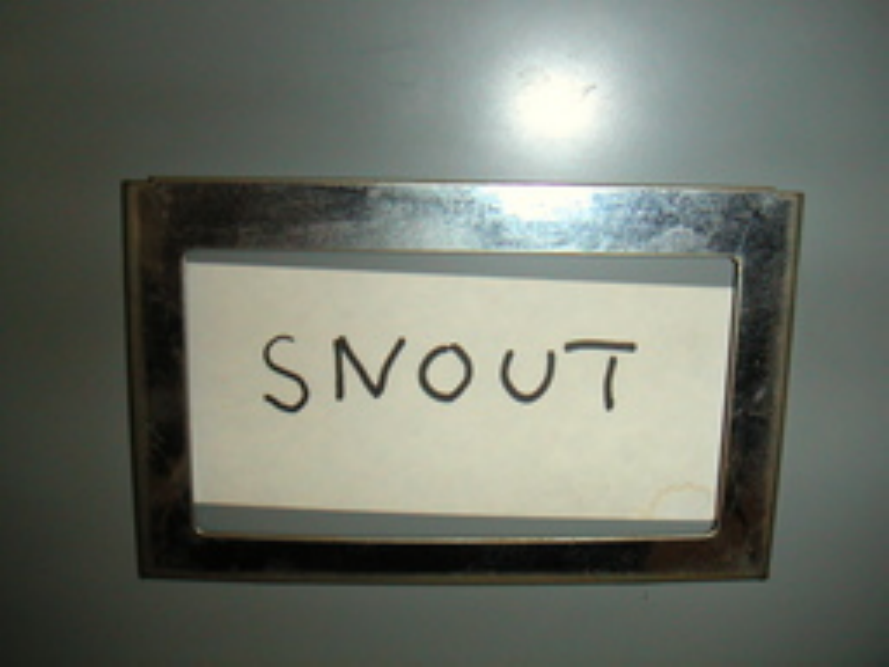}
    \includegraphics[width=.115\hsize]{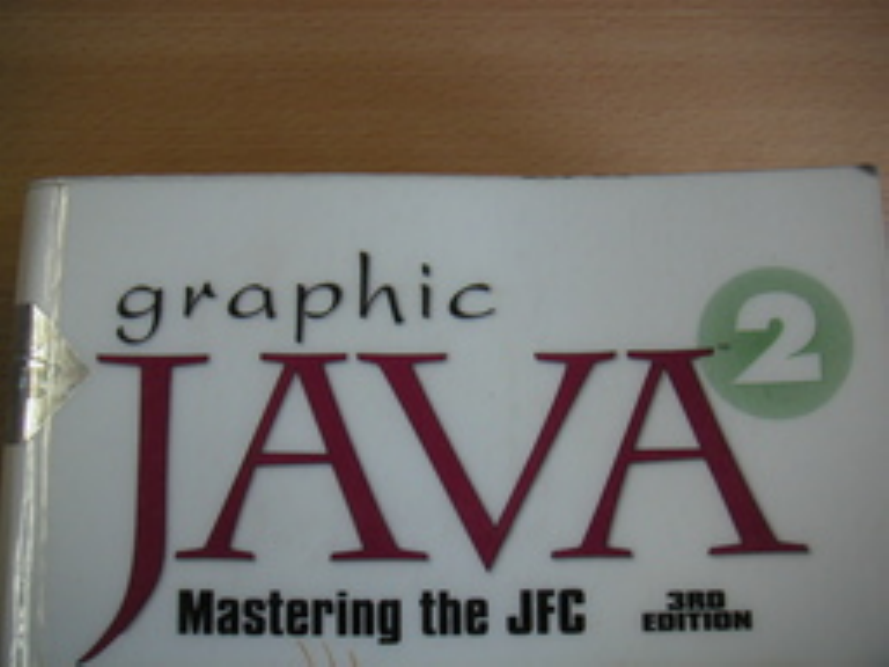}
    \includegraphics[width=.115\hsize]{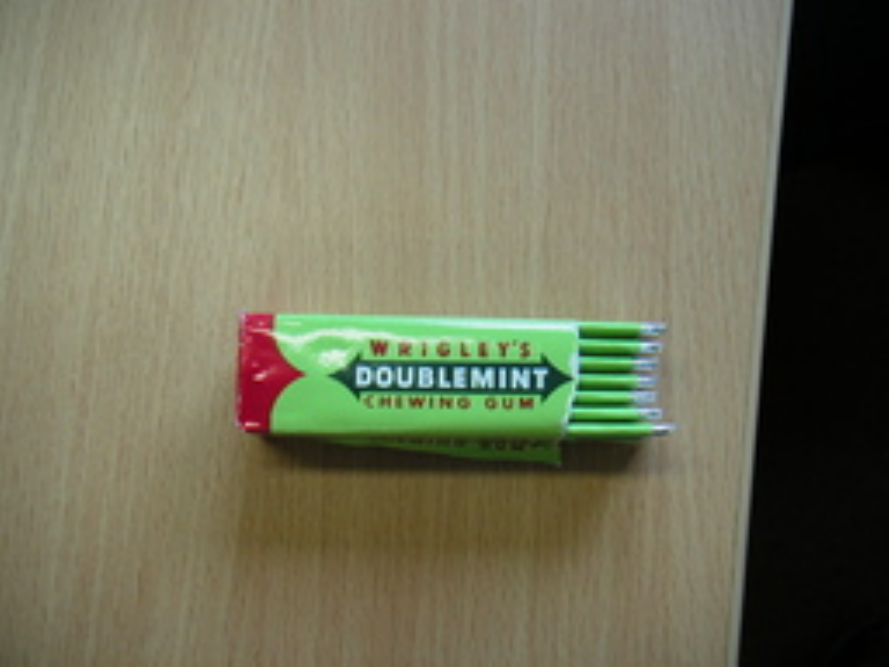}
    \subcaption{ICDAR Robust Reading Competitions (RRC) Dataset in 2003~\cite{ICDAR_RRC2003,ICDAR_RRC2003_IJDAR2005} and 2005~\cite{ICDAR_RRC2005}}
    \label{fig:ICDAR_RRC_2003}
  \end{minipage}
  \begin{minipage}[b]{\hsize}
    \centering    
    \includegraphics[width=.115\hsize]{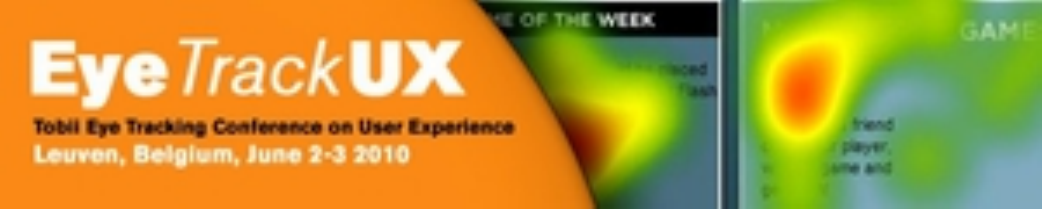}
    \includegraphics[width=.115\hsize]{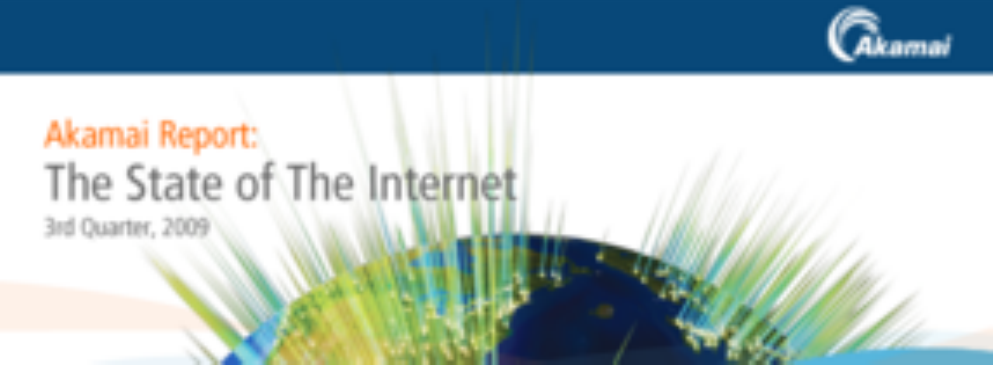}
    \includegraphics[width=.115\hsize]{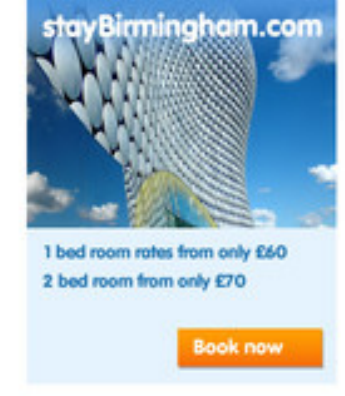}
    \includegraphics[width=.115\hsize]{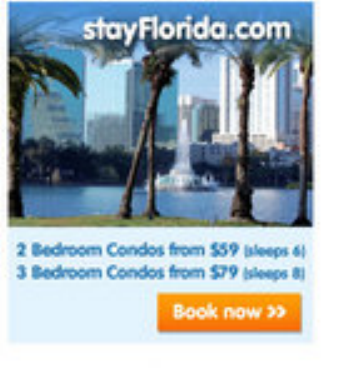}
    \includegraphics[width=.115\hsize]{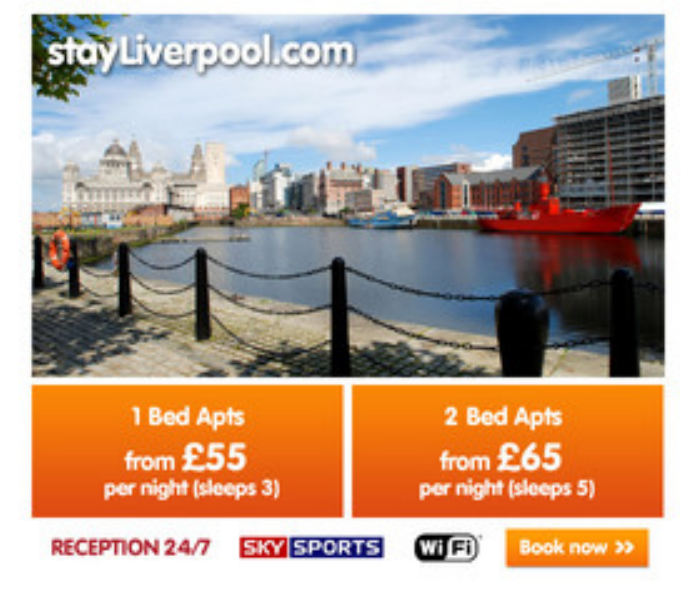}
    \includegraphics[width=.115\hsize]{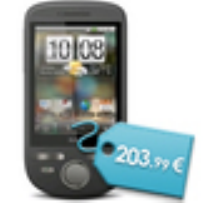}
    \includegraphics[width=.115\hsize]{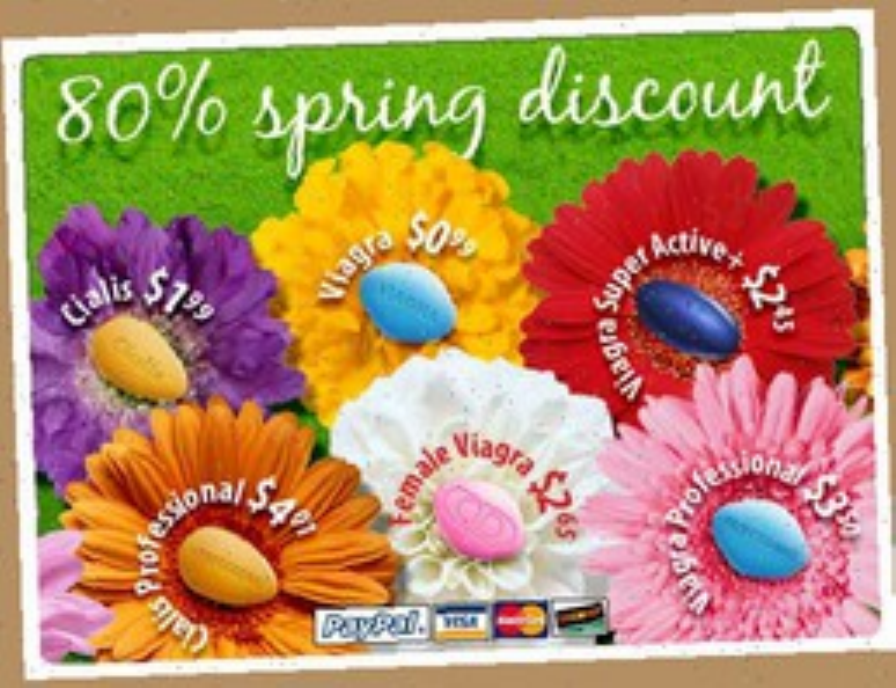}
    \includegraphics[width=.115\hsize]{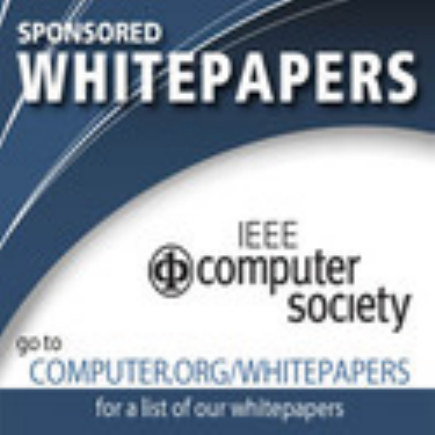}
    \subcaption{ICDAR RRC ``Born Digital Images'' (Challenge 1) Dataset
      in 2011~\cite{ICDAR2011_RRC_challenge1},
      2013~\cite{ICDAR2013_RRC} and 2015~\cite{ICDAR2015_RRC}}
    \label{fig:ICDAR_RRC_C1}
  \end{minipage}
  \begin{minipage}[b]{\hsize}
    \centering    
    \includegraphics[width=.115\hsize]{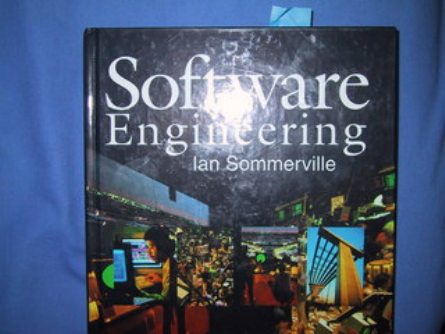}
    \includegraphics[width=.115\hsize]{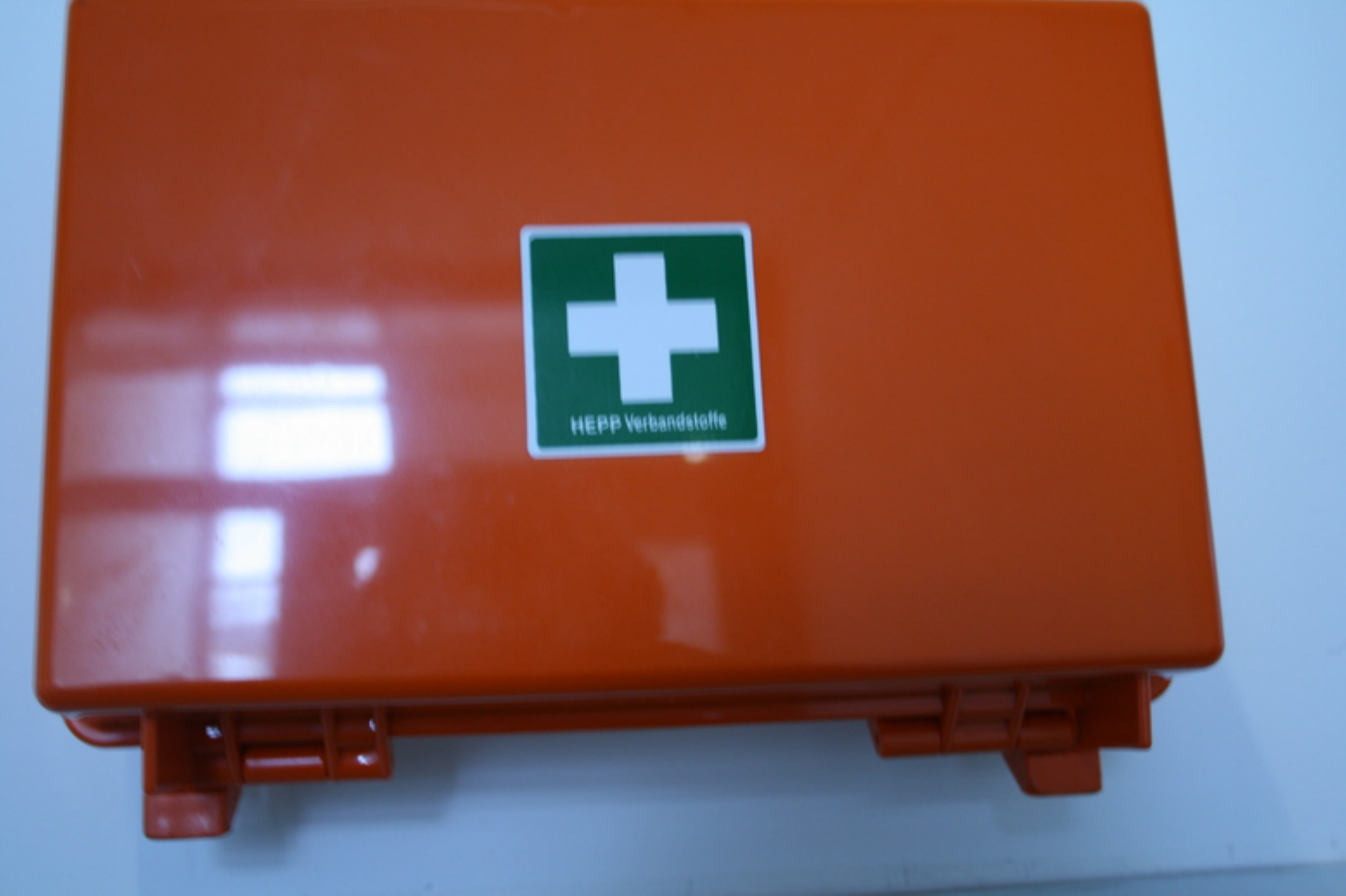}
    \includegraphics[width=.115\hsize]{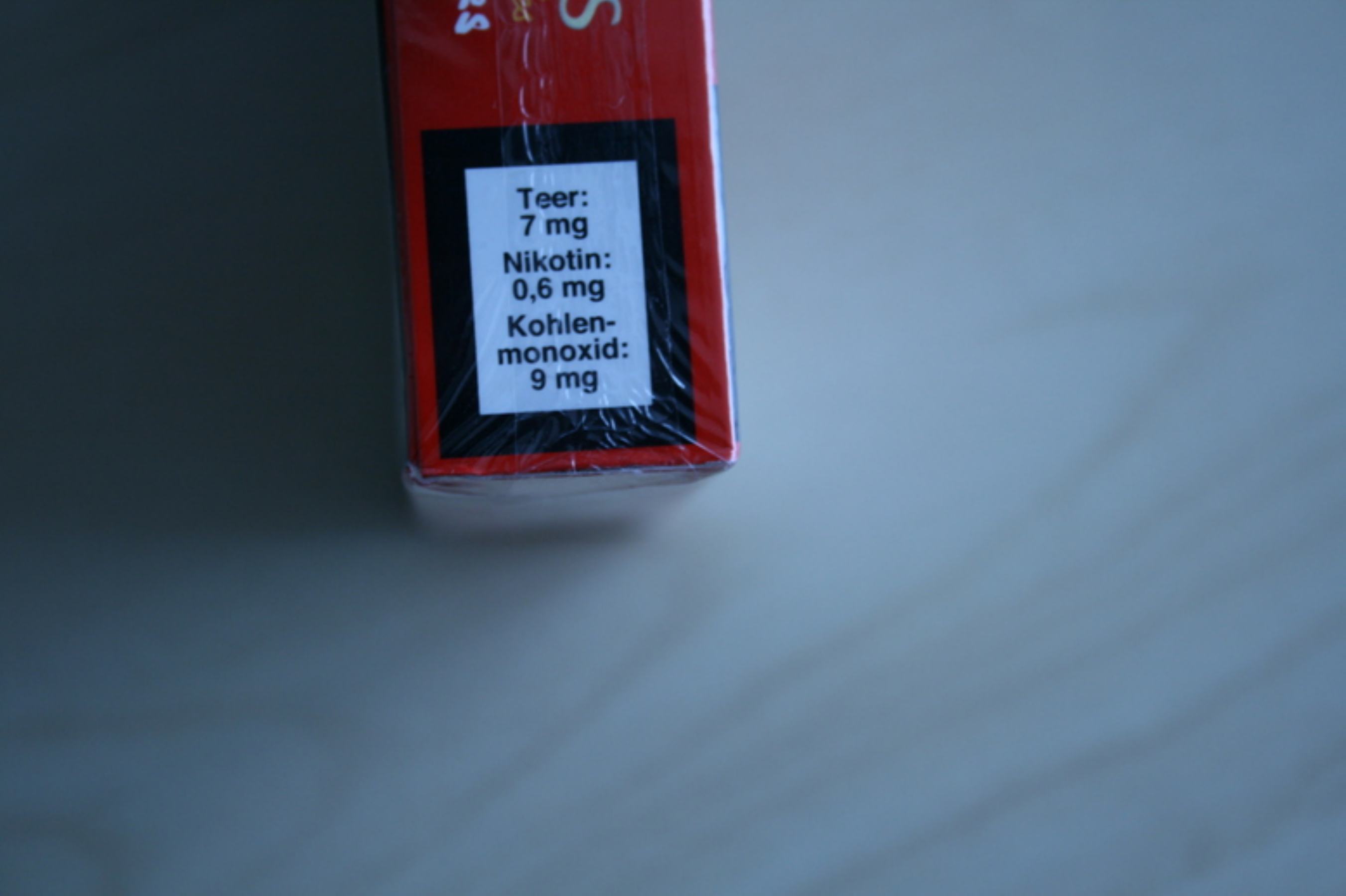}
    \includegraphics[width=.115\hsize]{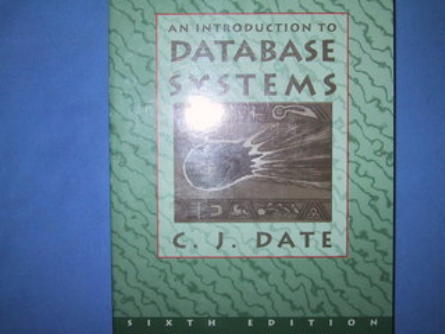}
    \includegraphics[width=.115\hsize]{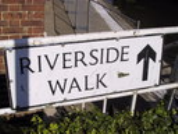}
    \includegraphics[width=.115\hsize]{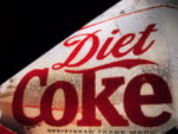}
    \includegraphics[width=.115\hsize]{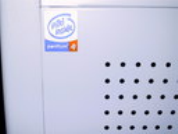}
    \includegraphics[width=.115\hsize]{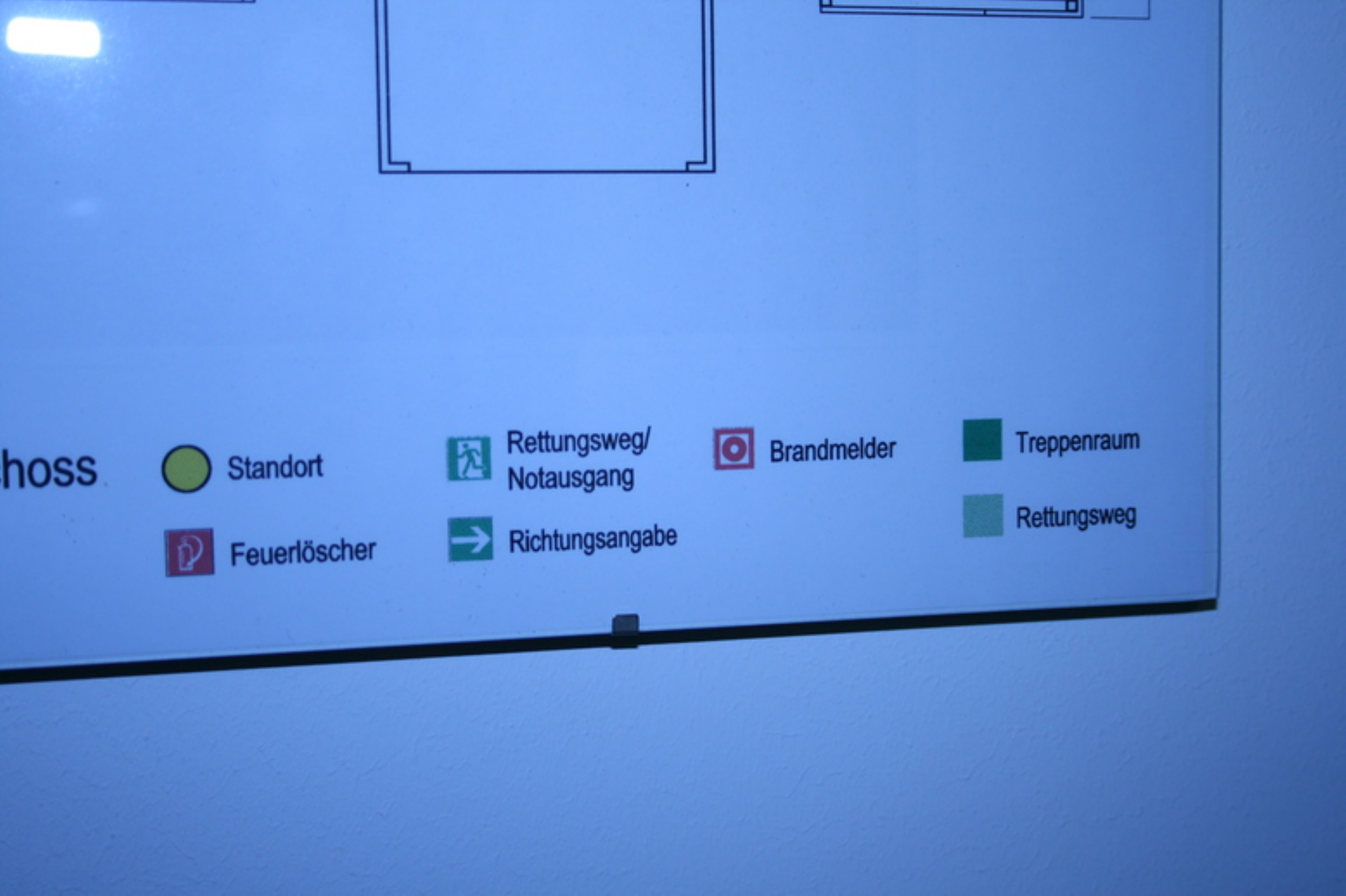}
    \subcaption{ICDAR RRC ``Focused Scene Text'' (Challenge 2) Dataset
      in 2011~\cite{ICDAR2011_RRC_challenge2},
      2013~\cite{ICDAR2013_RRC} and 2015~\cite{ICDAR2015_RRC}}
    \label{fig:ICDAR_RRC_C2}
  \end{minipage}
  \begin{minipage}[b]{\hsize}
    \centering    
    \includegraphics[width=.115\hsize]{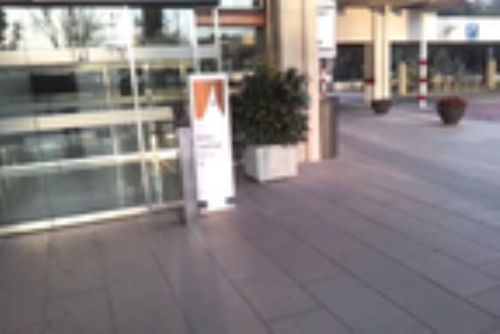}
    \includegraphics[width=.115\hsize]{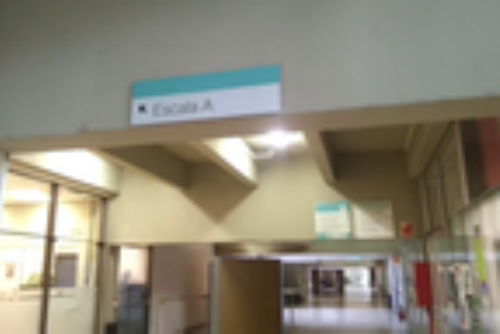}
    \includegraphics[width=.115\hsize]{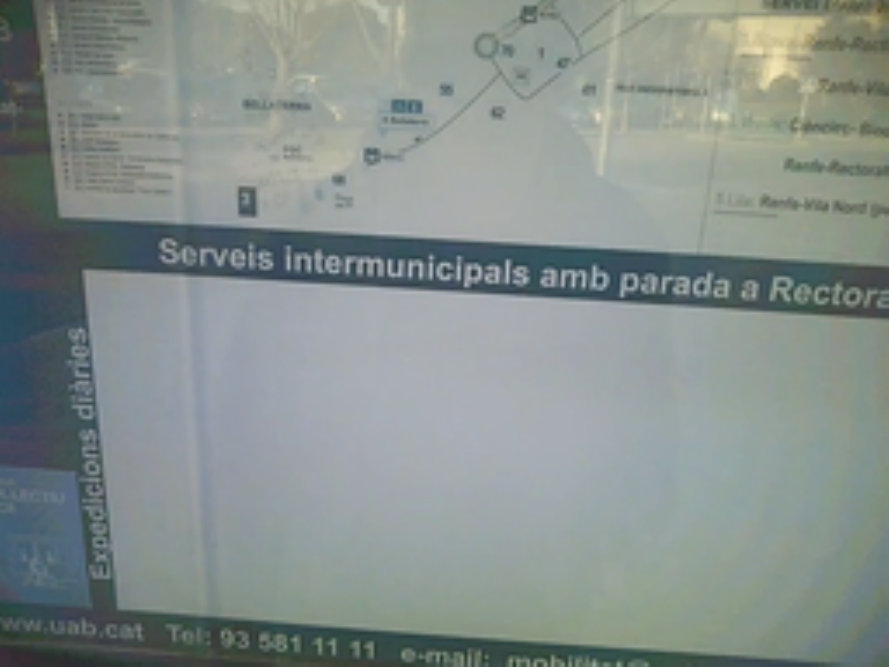}
    \includegraphics[width=.115\hsize]{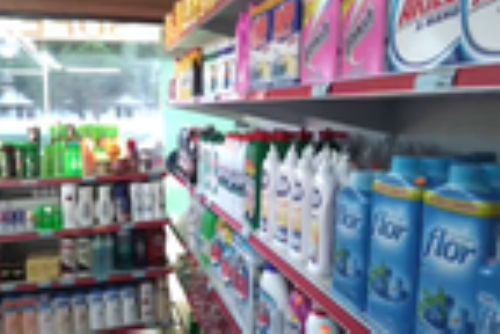}
    \includegraphics[width=.115\hsize]{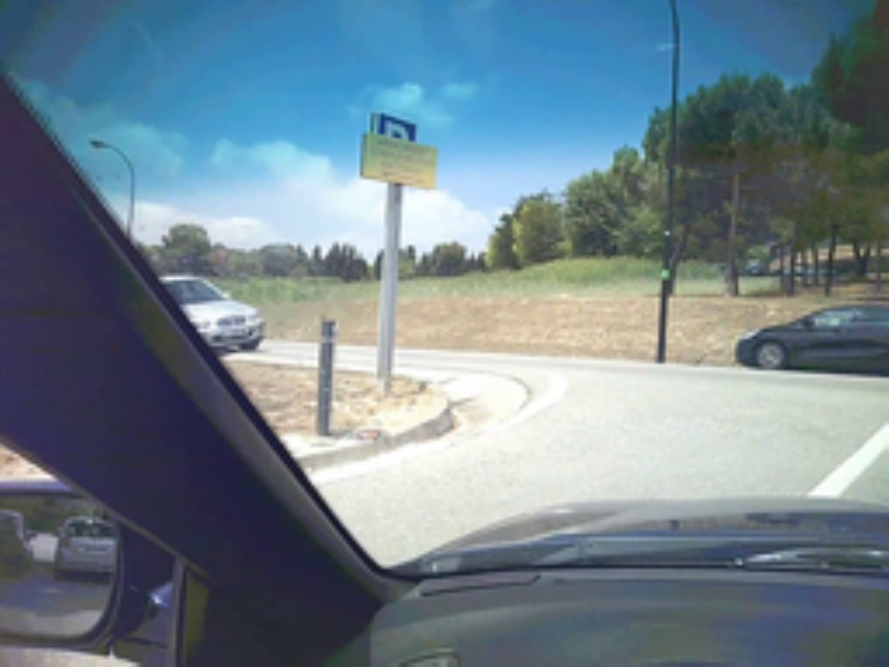}
    \includegraphics[width=.115\hsize]{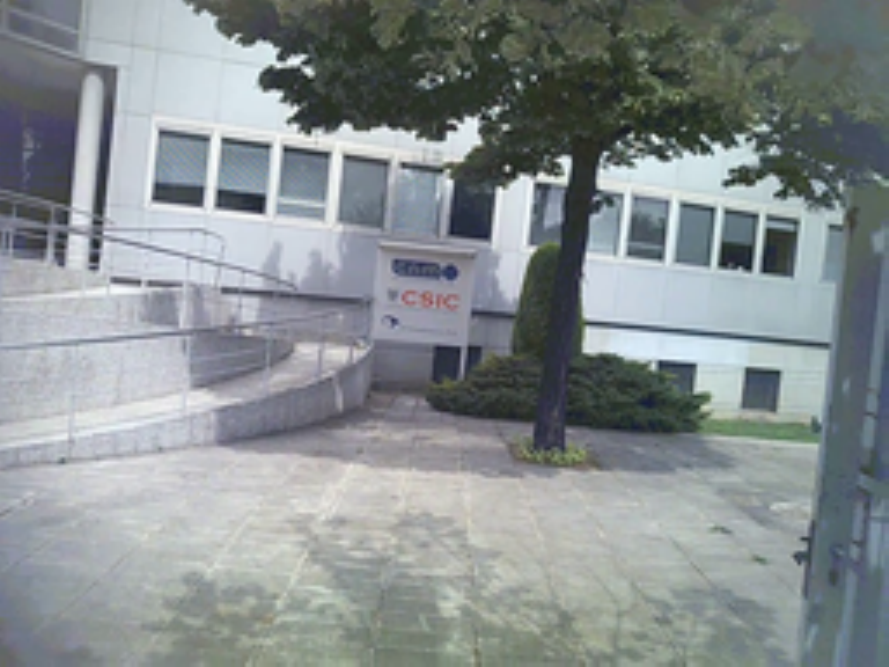}
    \includegraphics[width=.115\hsize]{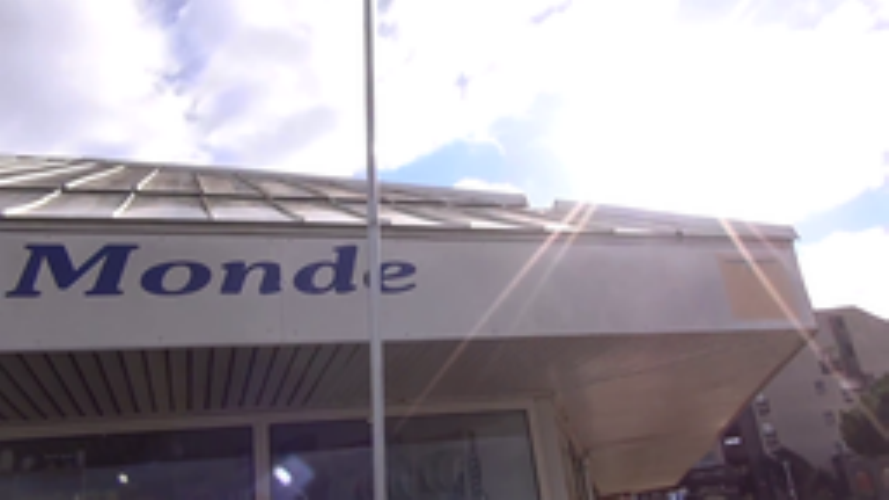}
    \includegraphics[width=.115\hsize]{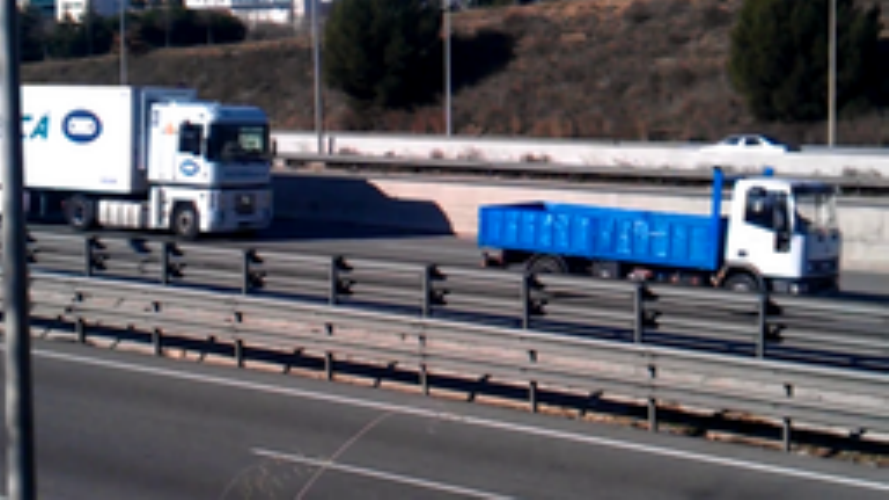}
    \subcaption{ICDAR RRC ``Text in Videos'' (Challenge 3) Dataset
      in 2013~\cite{ICDAR2013_RRC} and
      2015~\cite{ICDAR2015_RRC}}
    \label{fig:ICDAR_RRC_C3}
  \end{minipage}
  \begin{minipage}[b]{\hsize}
    \centering    
    \includegraphics[width=.115\hsize]{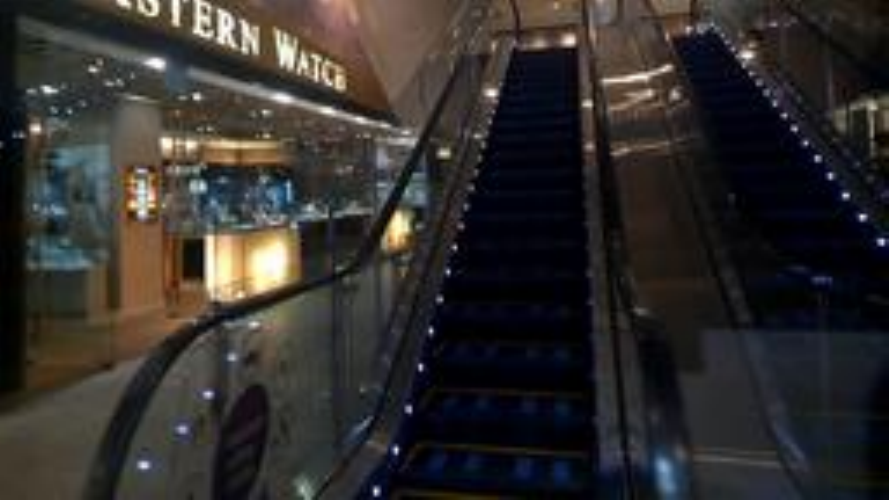}
    \includegraphics[width=.115\hsize]{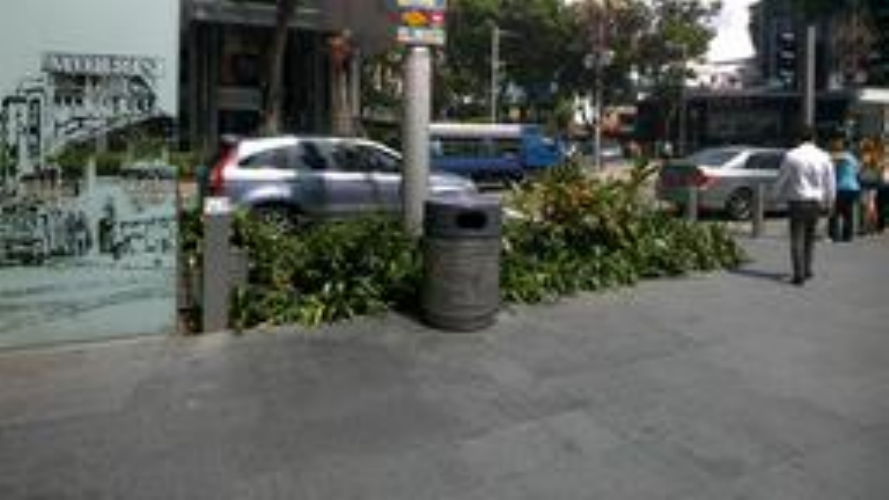}
    \includegraphics[width=.115\hsize]{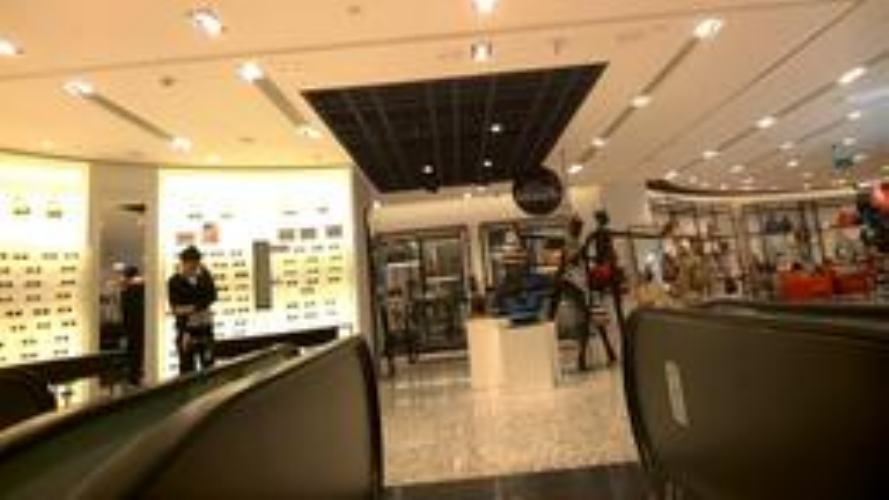}
    \includegraphics[width=.115\hsize]{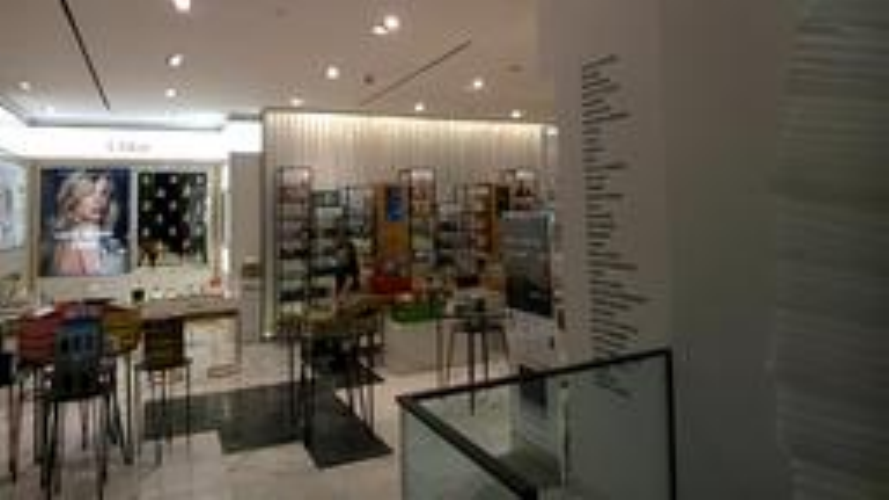}
    \includegraphics[width=.115\hsize]{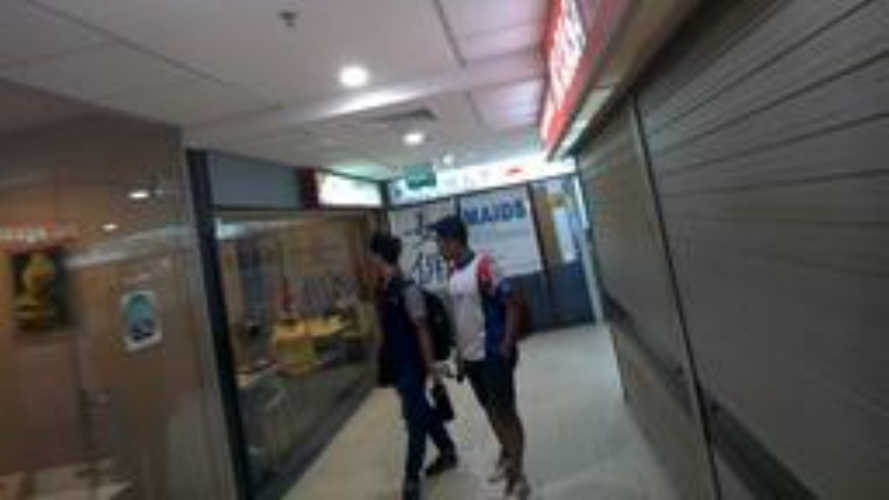}
    \includegraphics[width=.115\hsize]{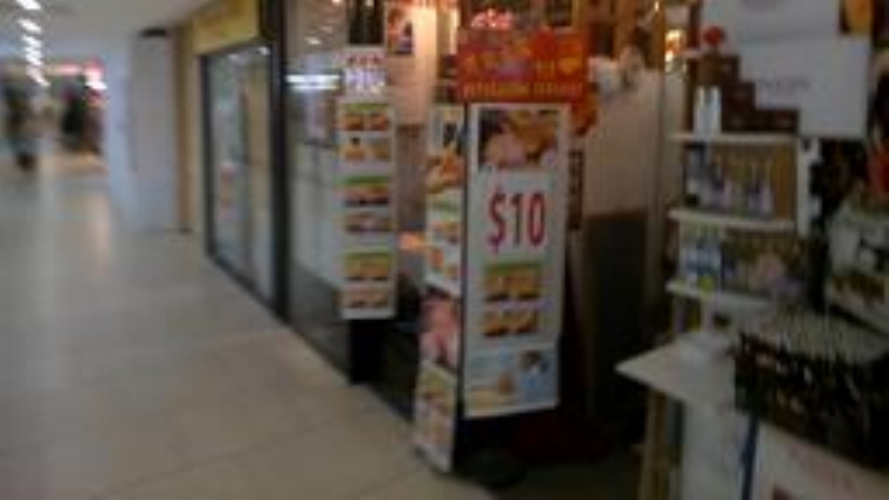}
    \includegraphics[width=.115\hsize]{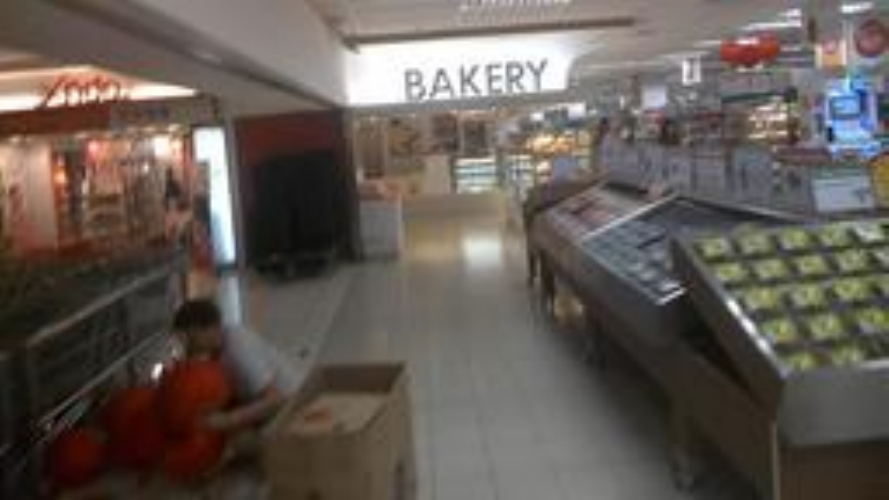}
    \includegraphics[width=.115\hsize]{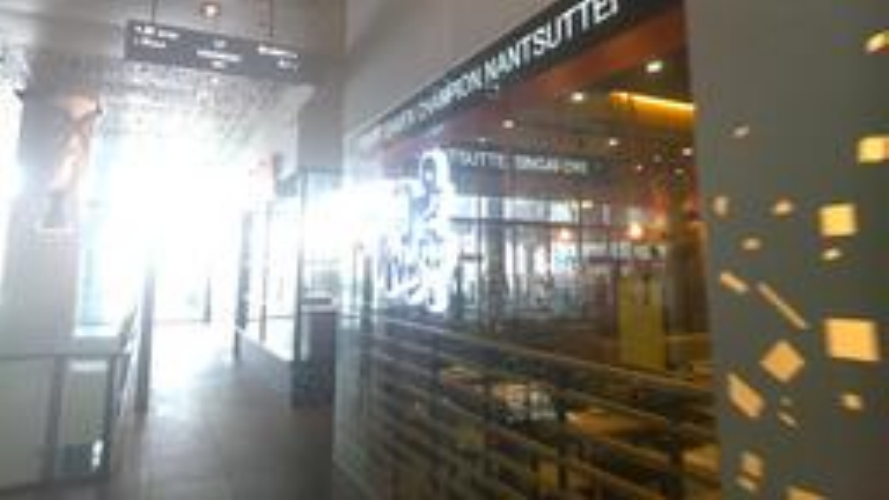}
    \subcaption{ICDAR RRC ``Incidental Scene Text'' (Challenge 4) Dataset in 2015~\cite{ICDAR2015_RRC}}
    \label{fig:ICDAR_RRC_C4}
  \end{minipage}
  \begin{minipage}[b]{\hsize}
    \centering    
    \includegraphics[width=.115\hsize]{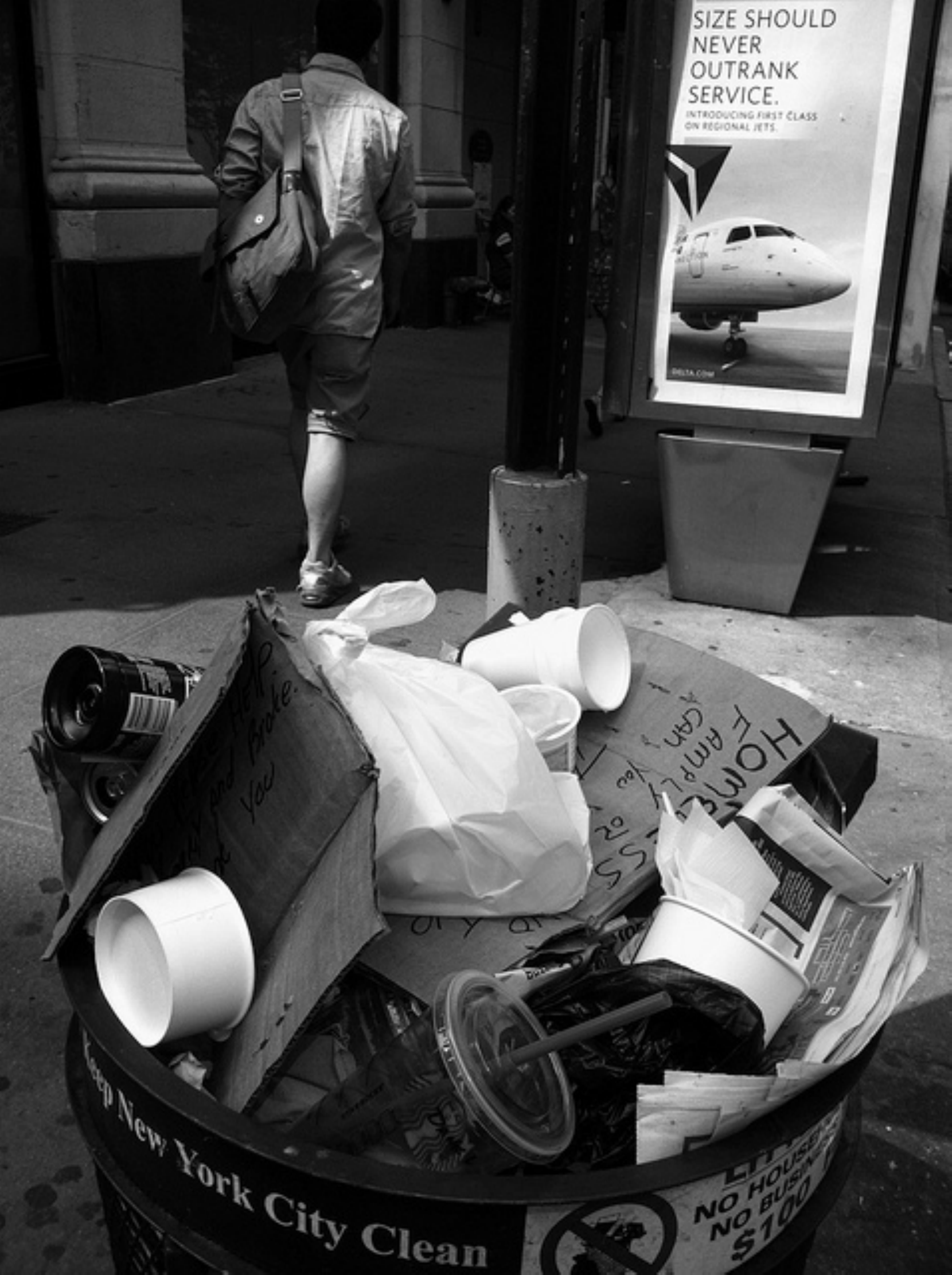}
    \includegraphics[width=.115\hsize]{db_image/COCO/COCO_train2014_000000016837.pdf}
    \includegraphics[width=.115\hsize]{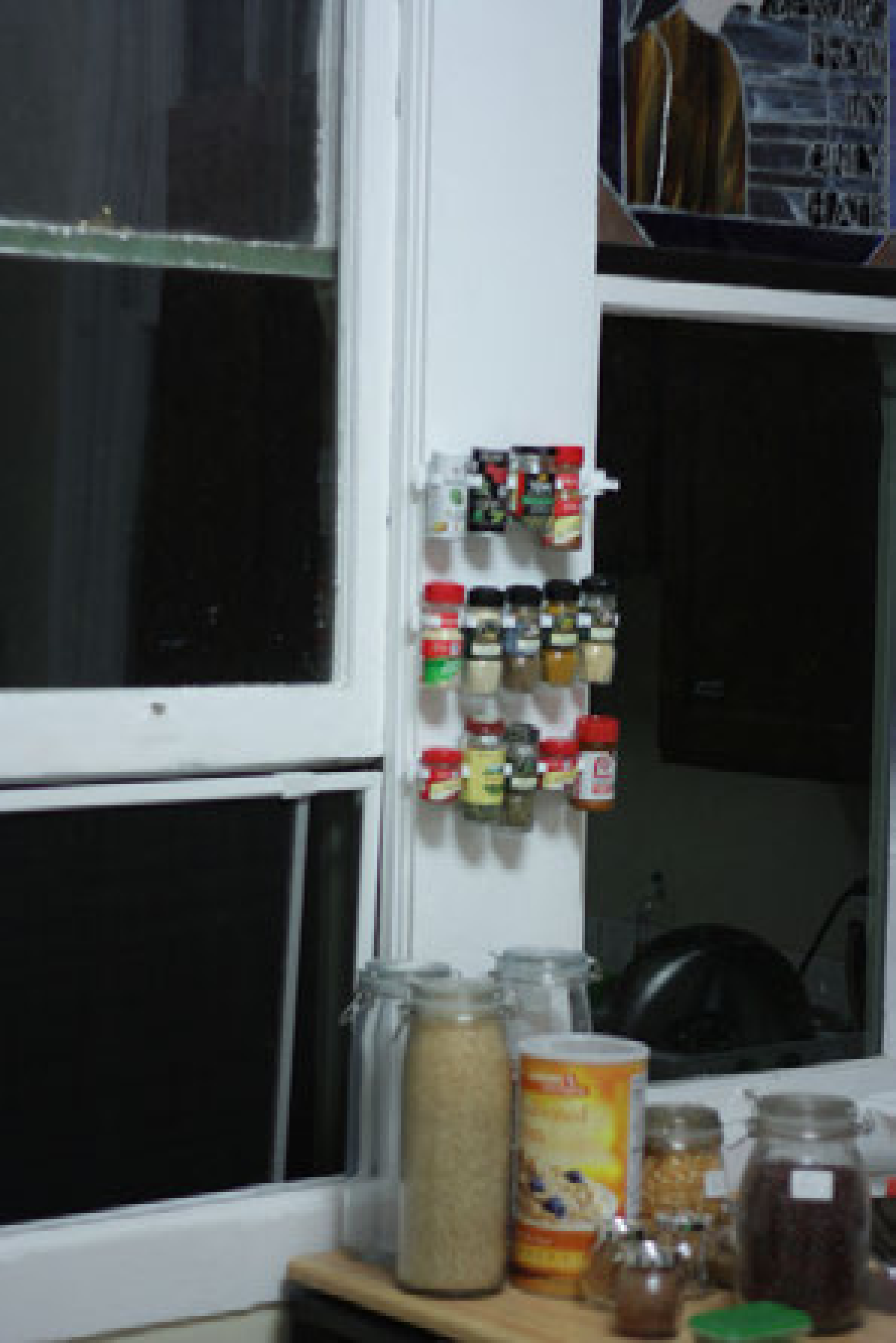}
    \includegraphics[width=.115\hsize]{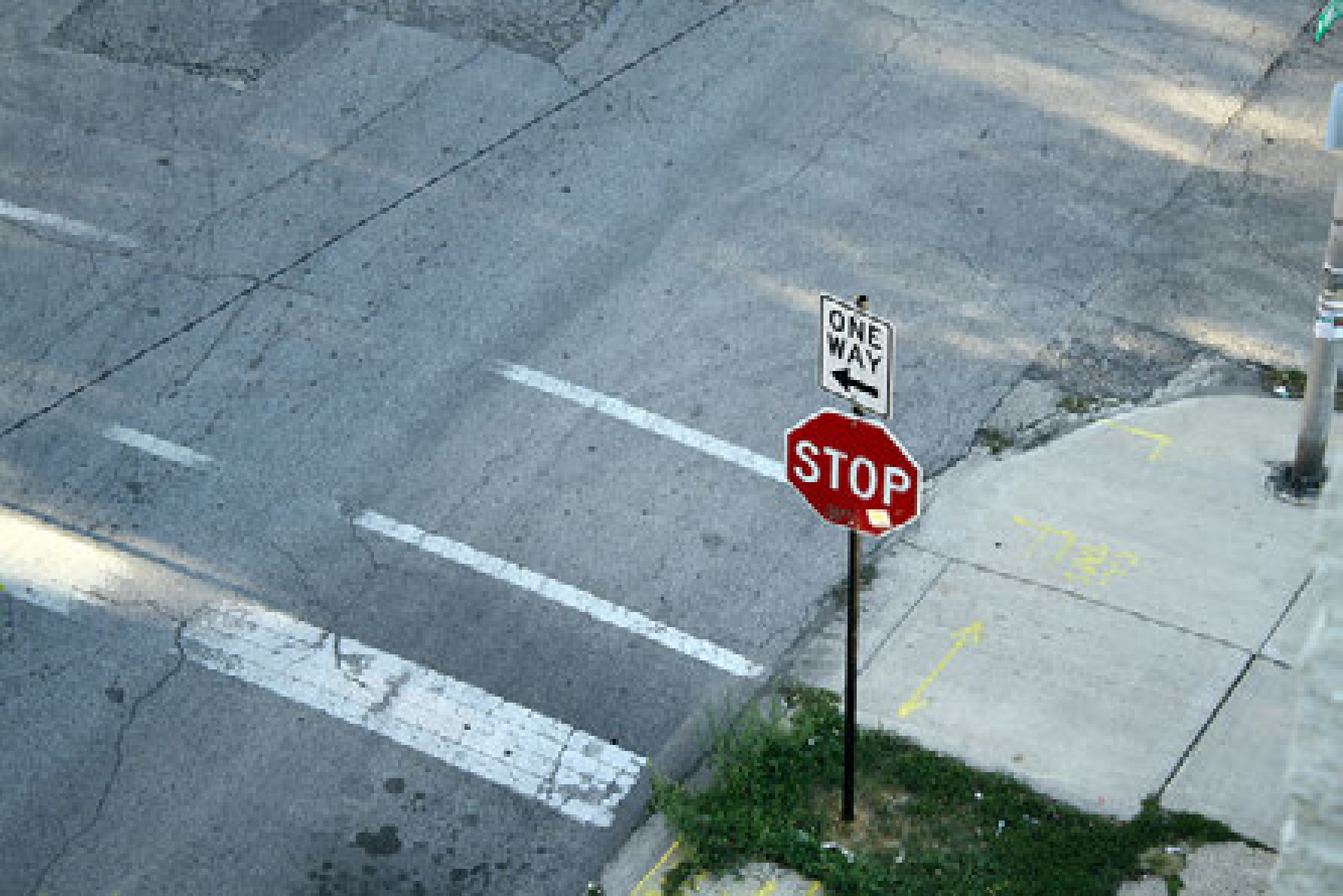}
    \includegraphics[width=.115\hsize]{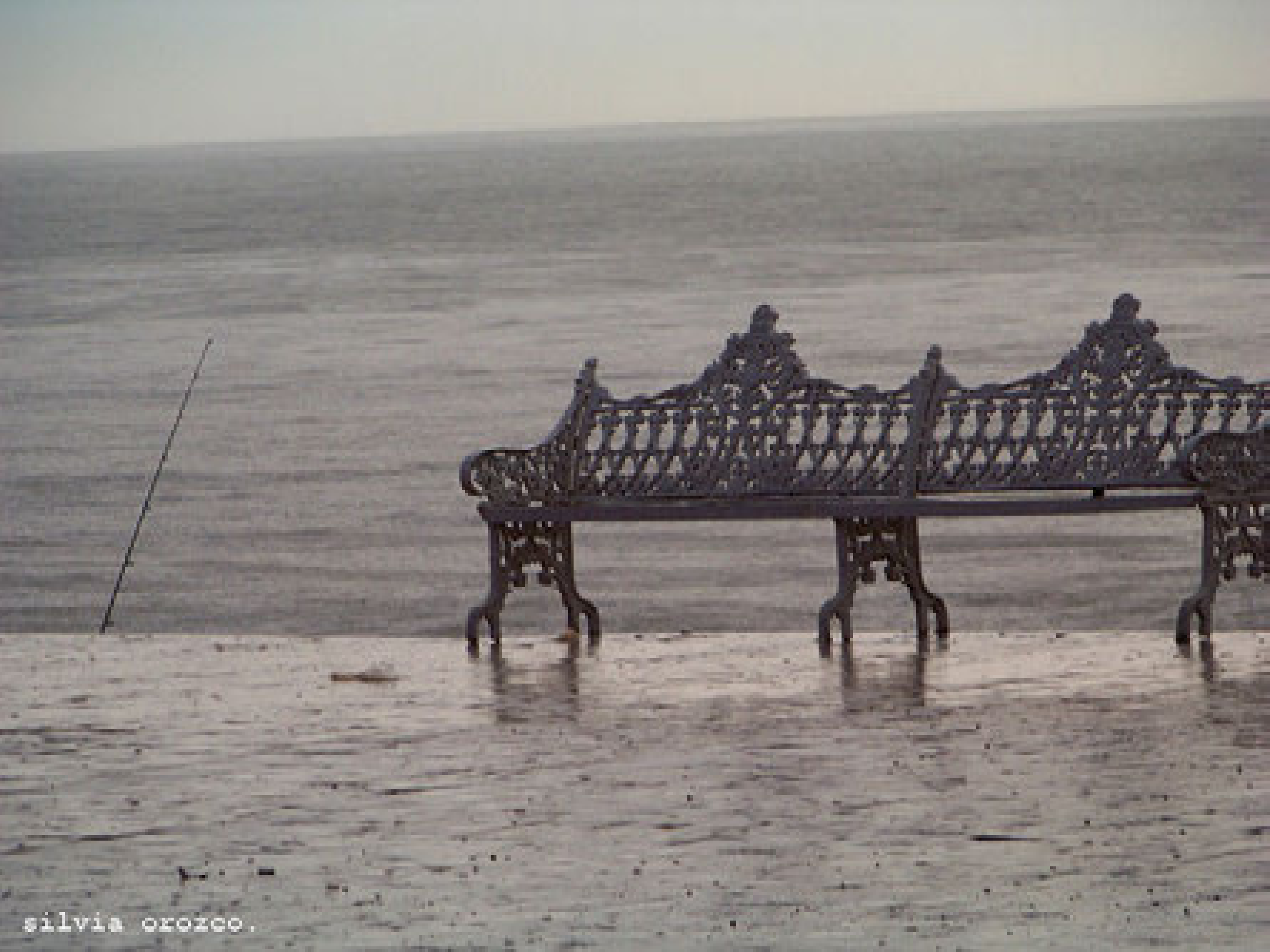}
    \includegraphics[width=.115\hsize]{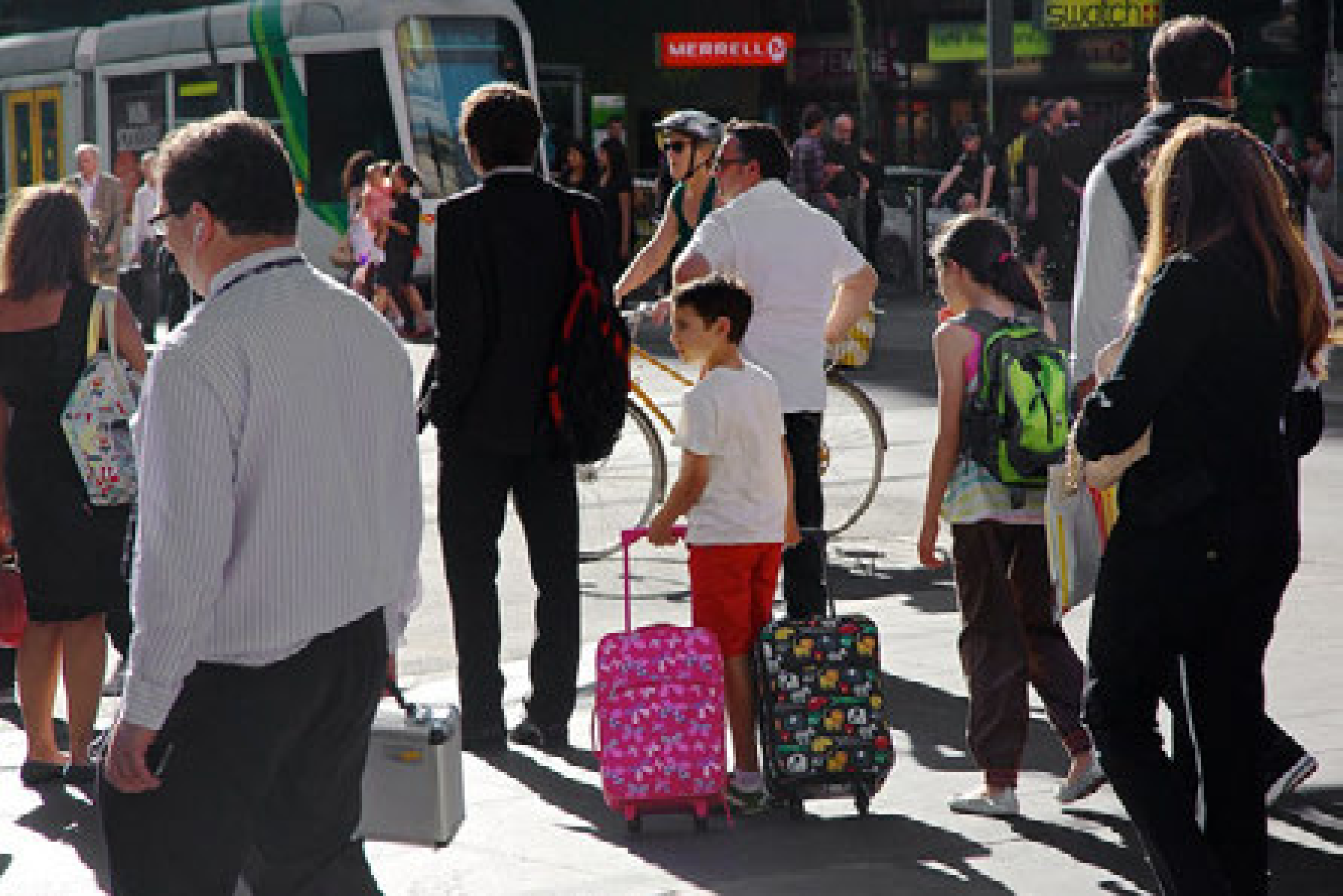}
    \includegraphics[width=.115\hsize]{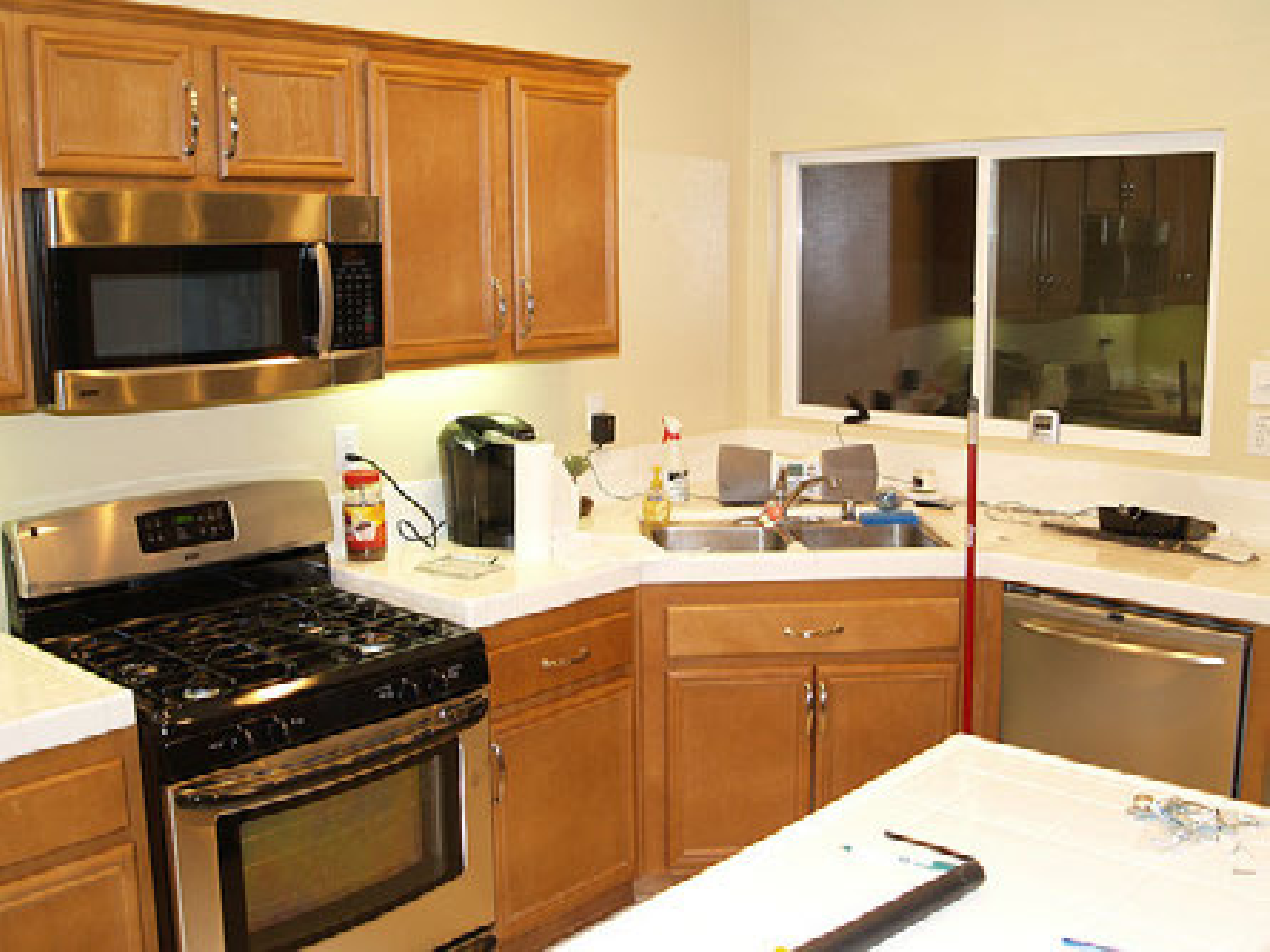}
    \includegraphics[width=.115\hsize]{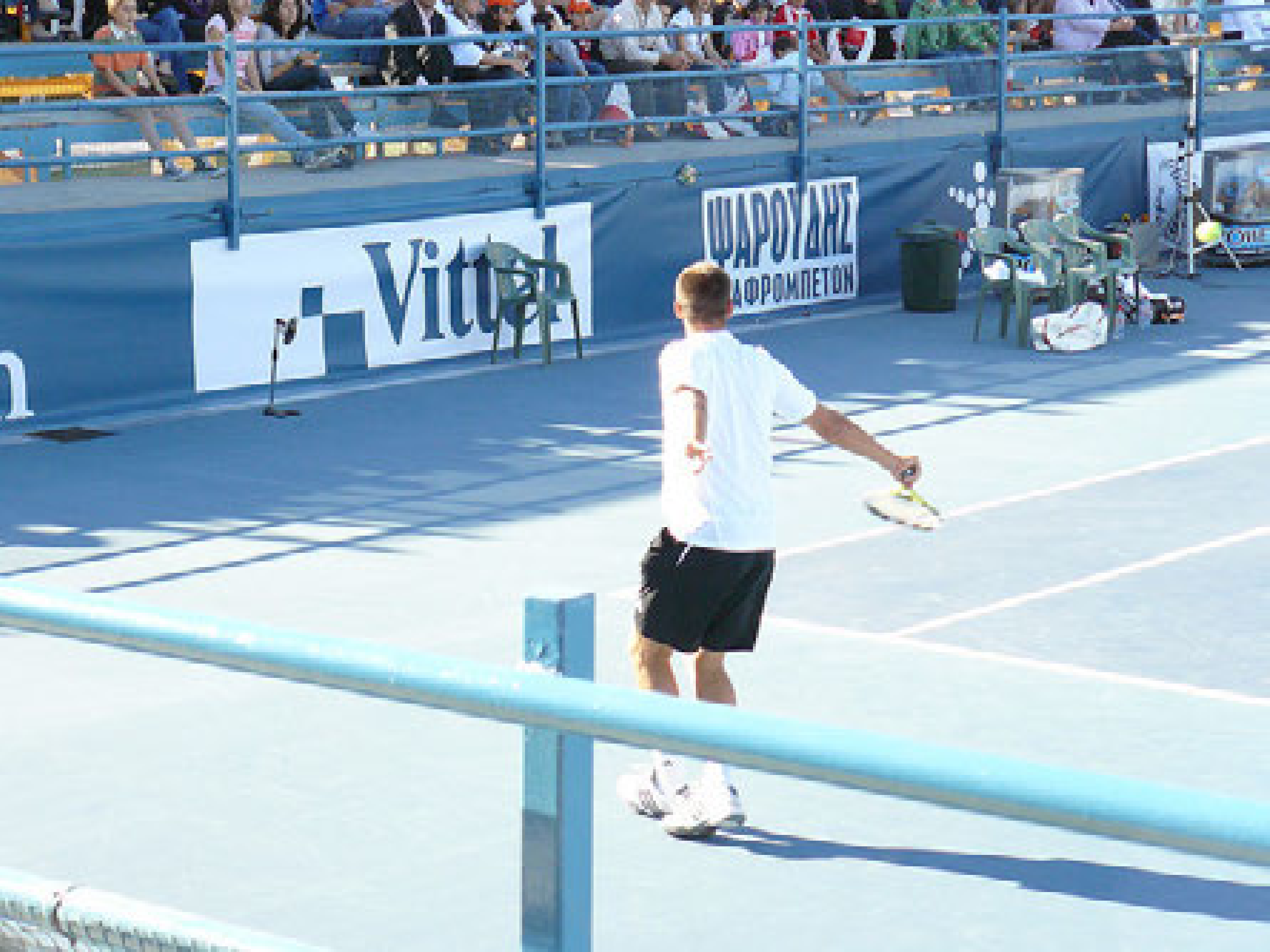}
    \subcaption{COCO-Text Dataset~\cite{Veit_arXiv2016} / ICDAR2017 Robust Reading Challenge (RRC) on COCO-Text~\cite{ICDAR_COCO2017}}
    \label{fig:COCO-Text}
  \end{minipage}
  \begin{minipage}[b]{\hsize}
    \centering    
    \includegraphics[width=.32\hsize]{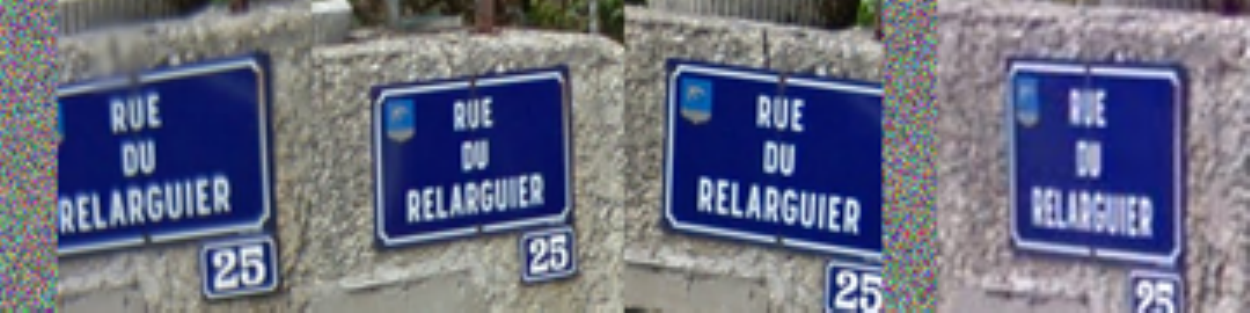}
    \includegraphics[width=.32\hsize]{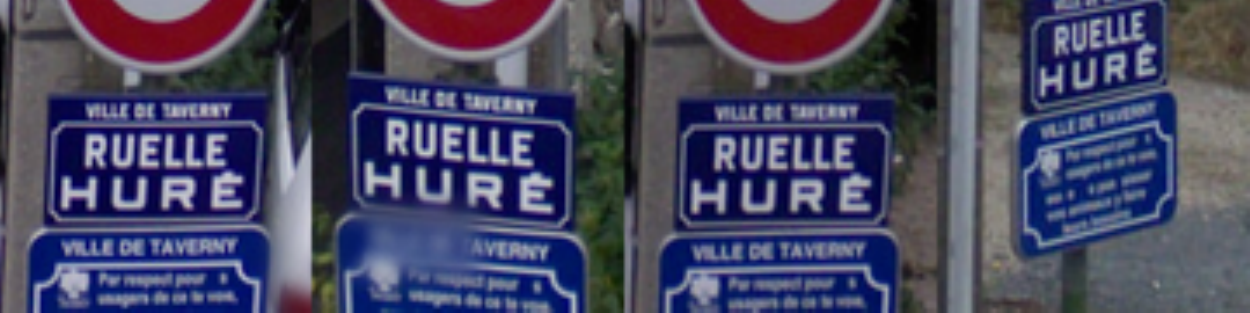}
    \includegraphics[width=.32\hsize]{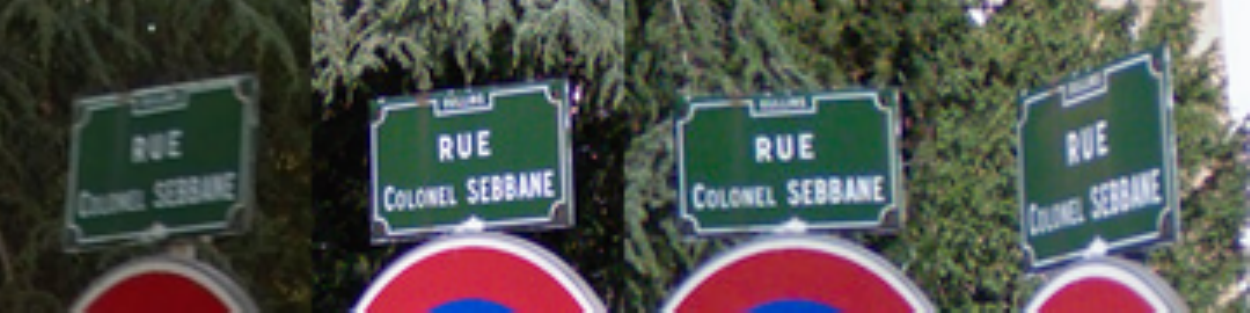}
    \subcaption{French Street Name Signs (FSNS) Dataset~\cite{Smith_IWRR2016} / ICDAR2017 Robust Reading Challenge (RRC) on End-to-End Recognition on the Google FSNS Dataset}
    \label{fig:FSNS}
  \end{minipage}
  \begin{minipage}[b]{\hsize}
    \centering    
    \includegraphics[width=.115\hsize]{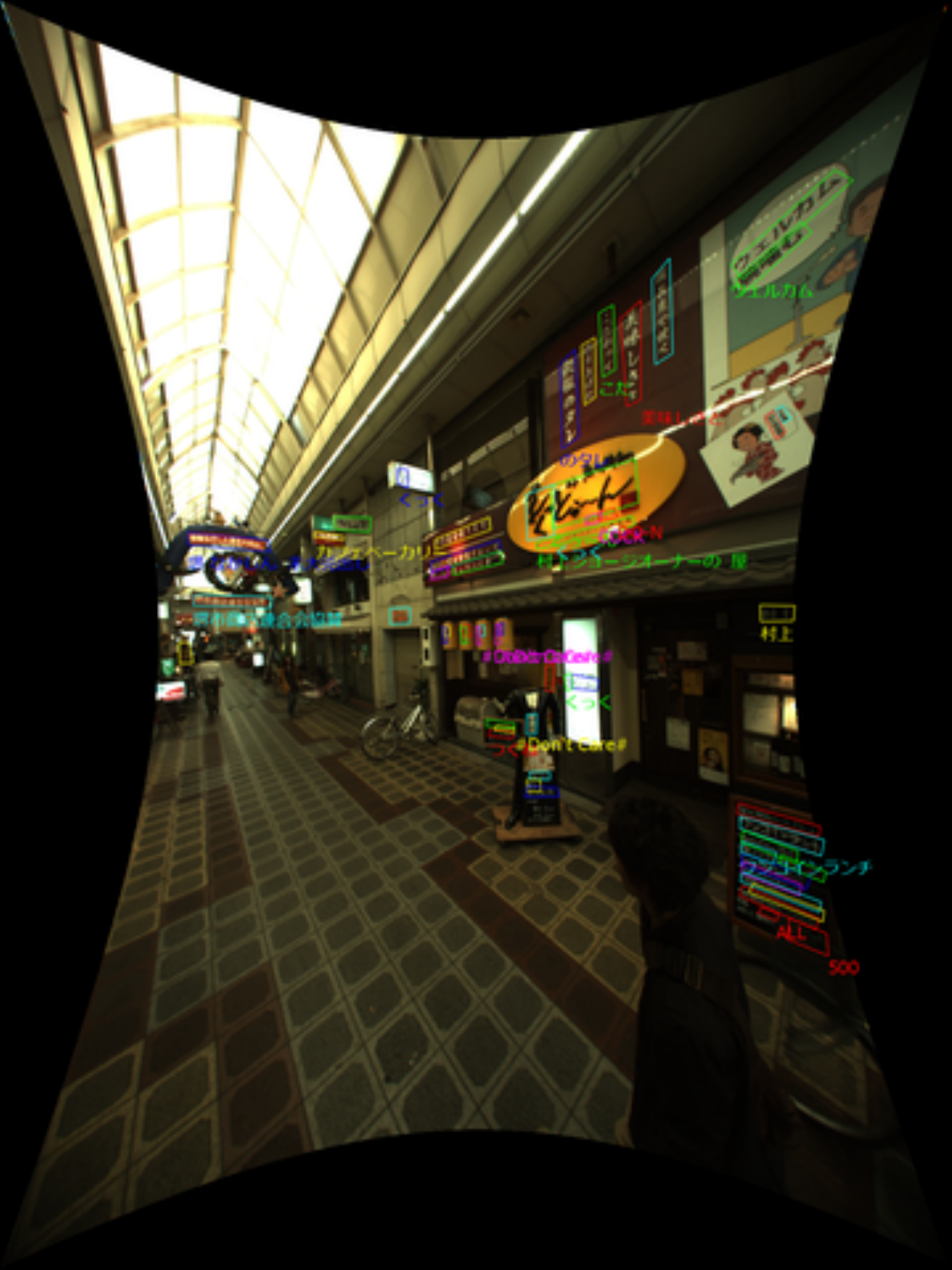}
    \includegraphics[width=.115\hsize]{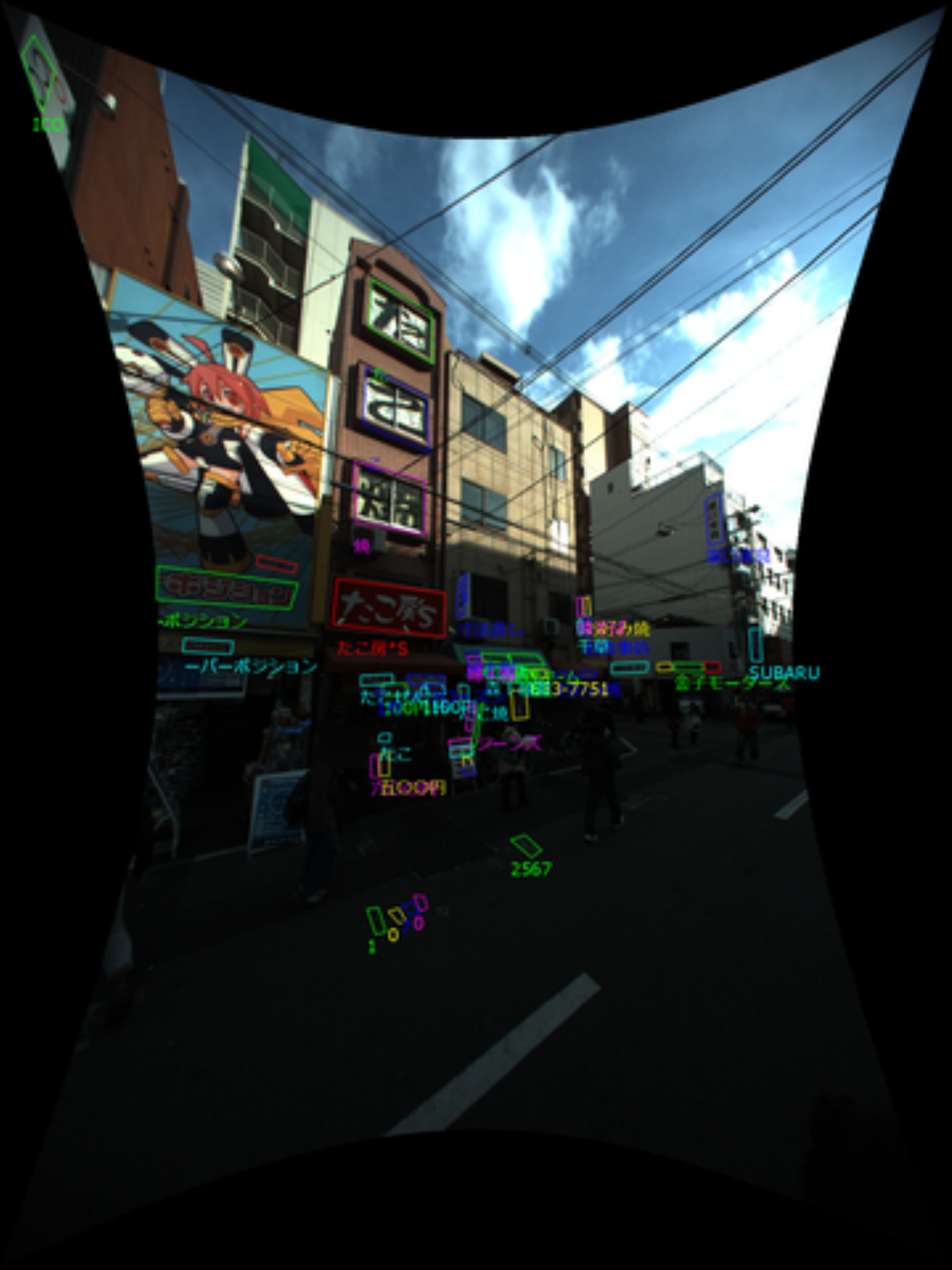}
    \includegraphics[width=.115\hsize]{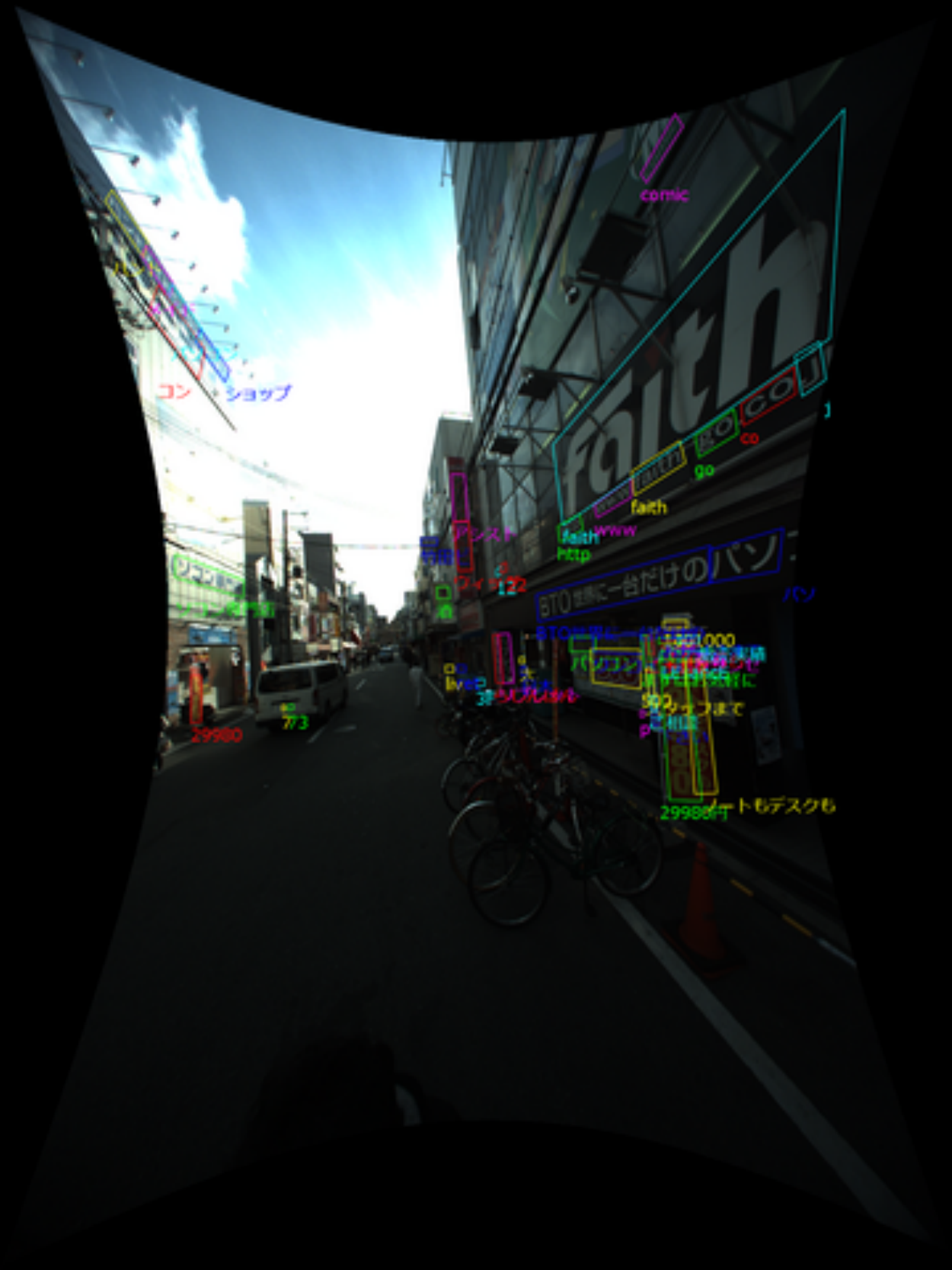}
    \includegraphics[width=.115\hsize]{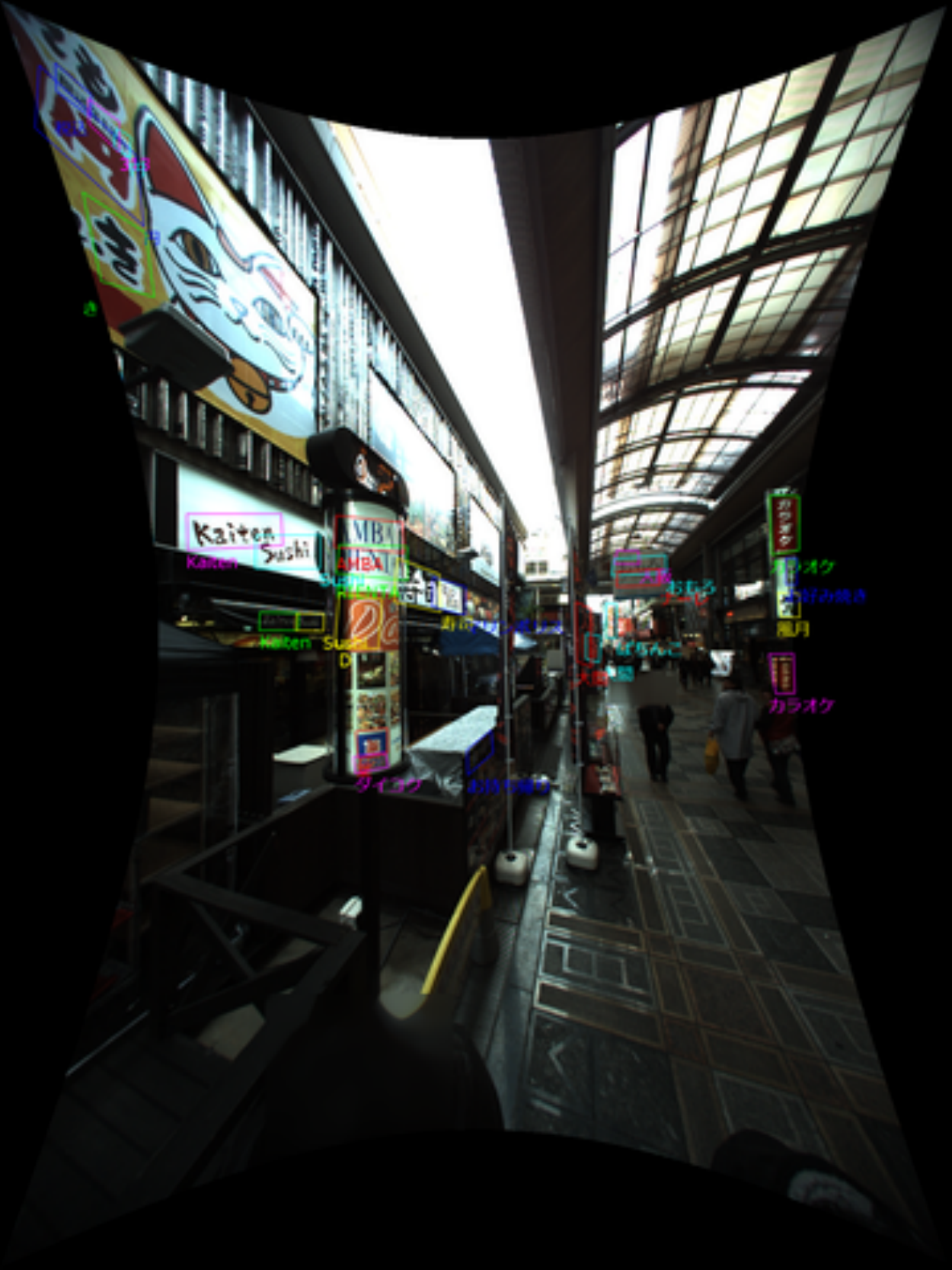}
    \includegraphics[width=.115\hsize]{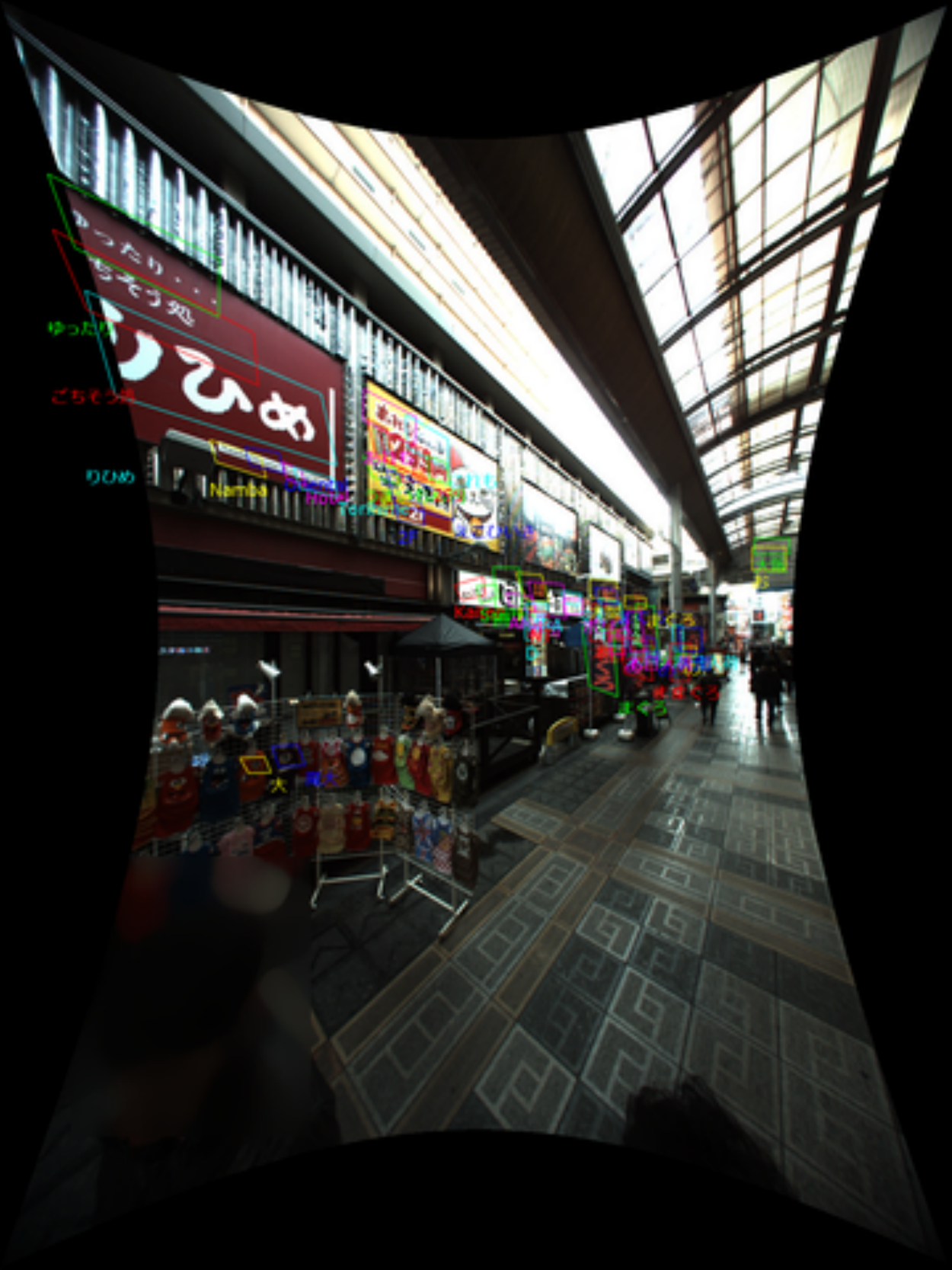}
    \includegraphics[width=.115\hsize]{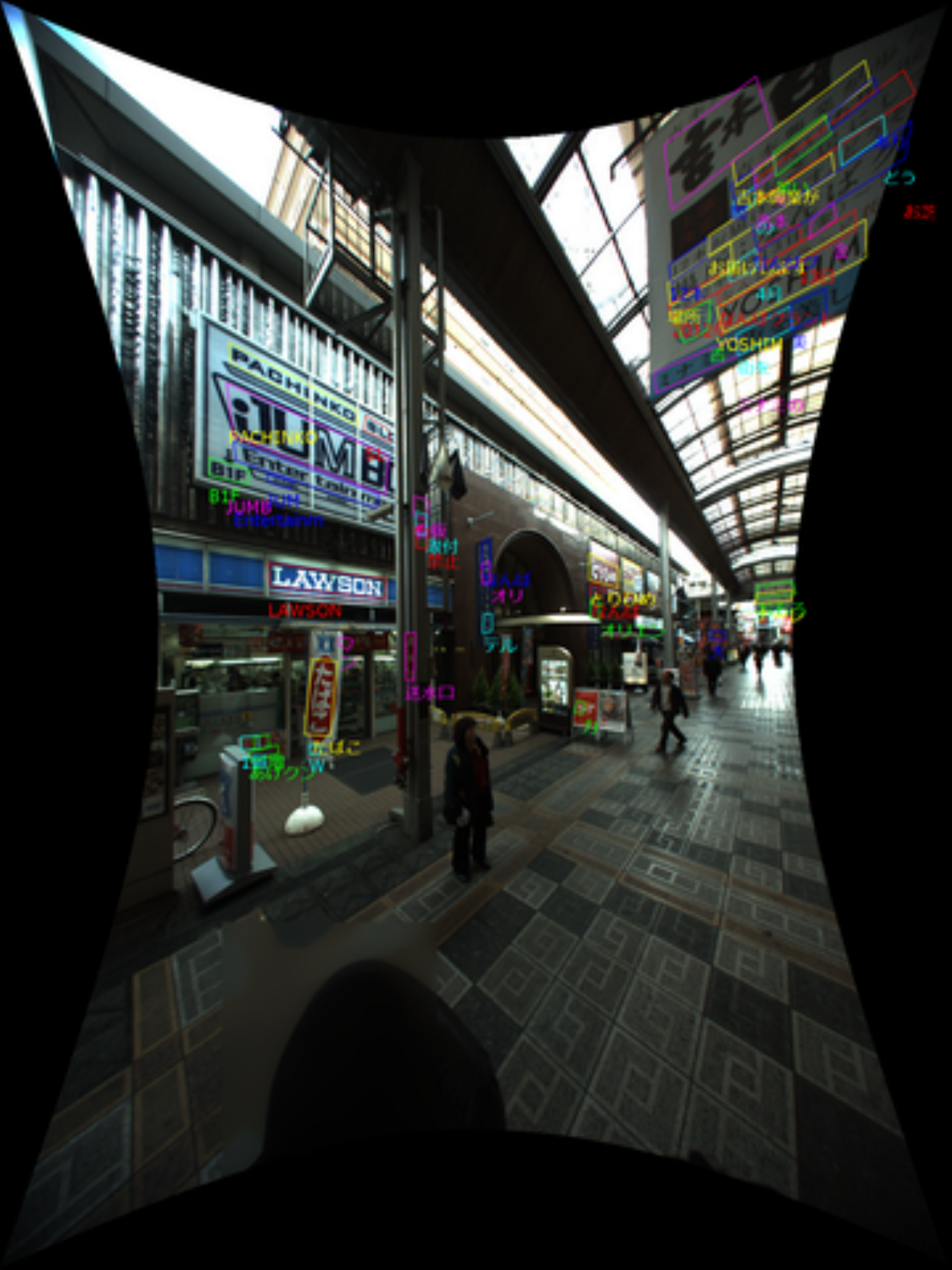}
    \includegraphics[width=.115\hsize]{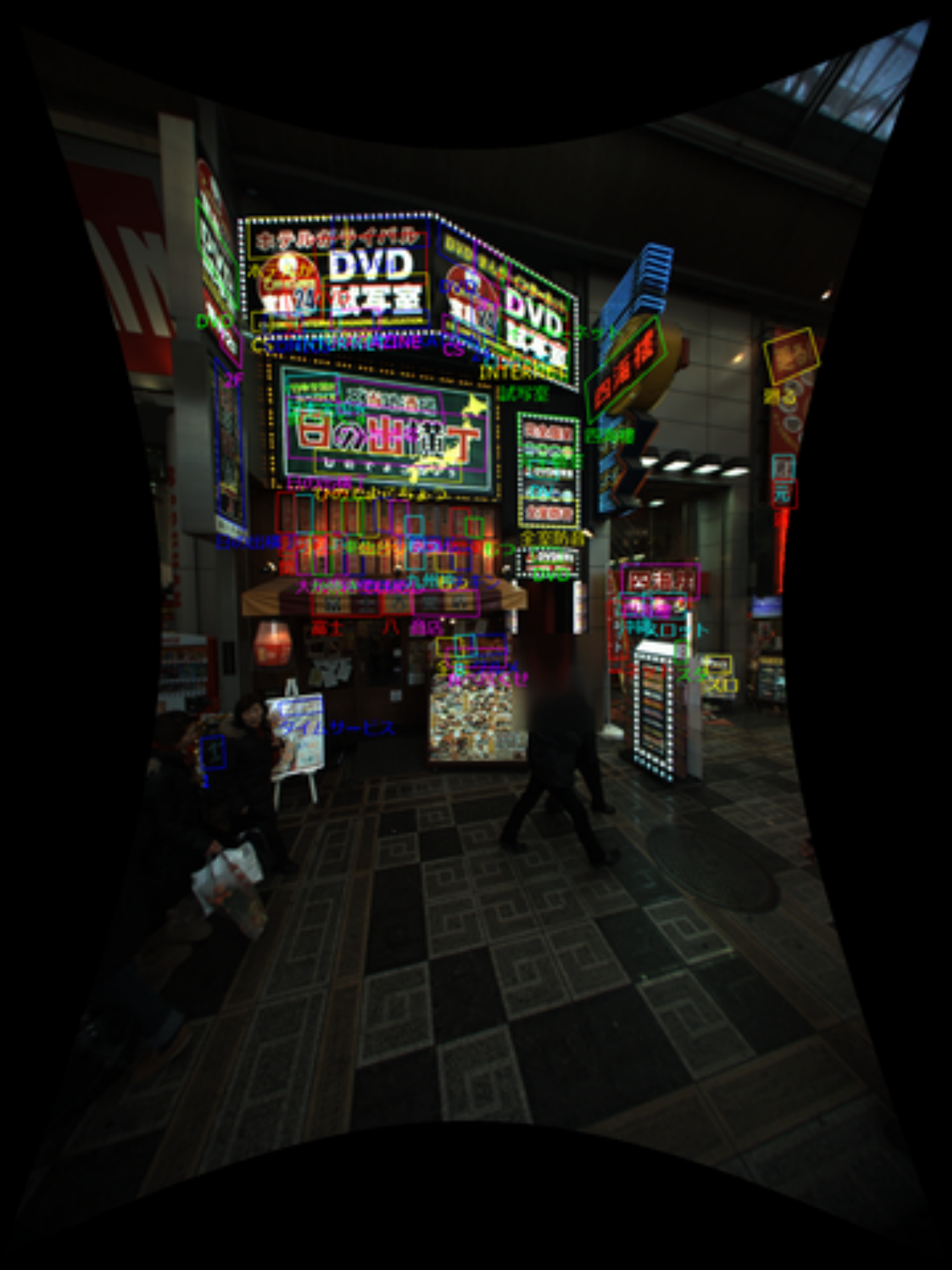}
    \includegraphics[width=.115\hsize]{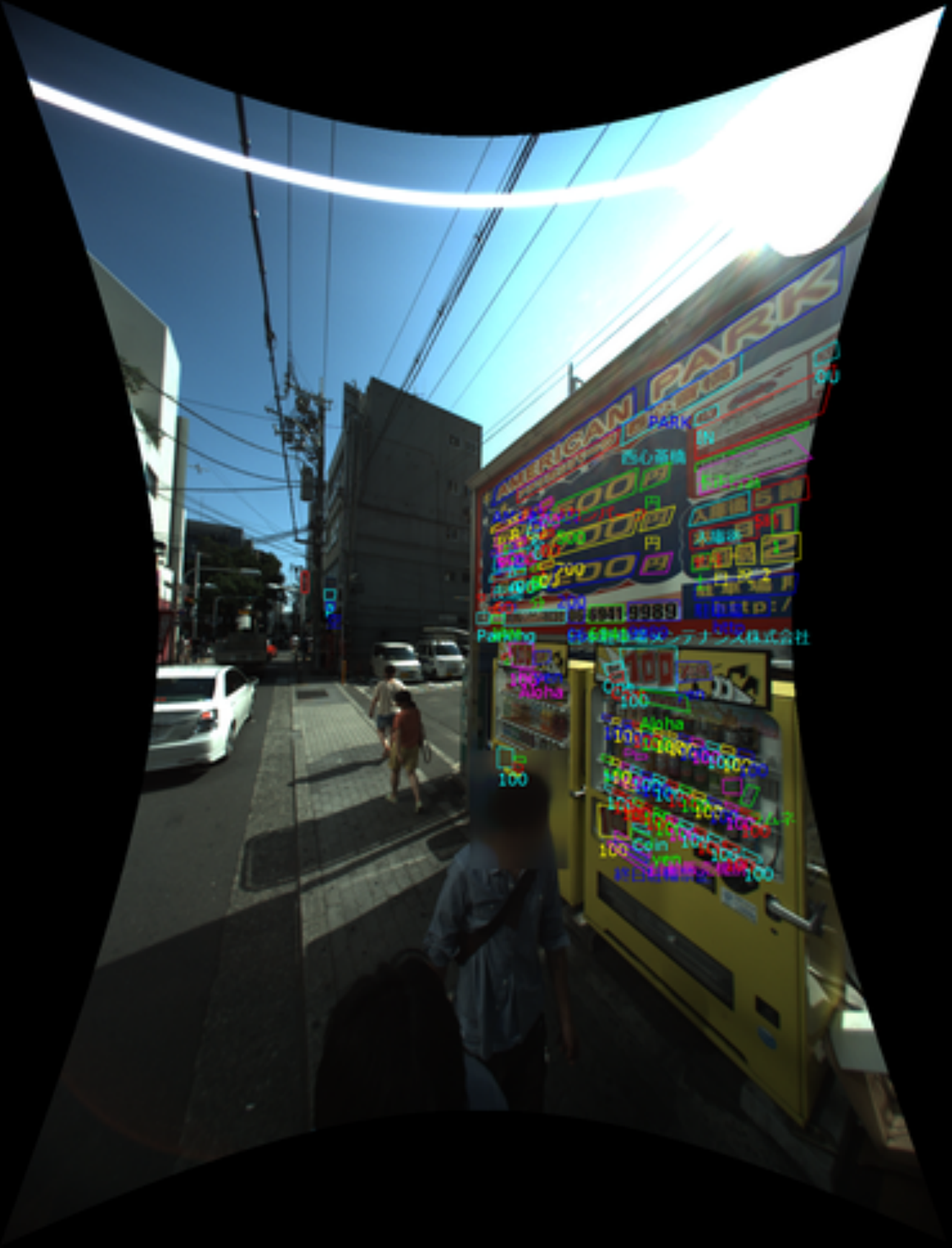}
    \subcaption{Downtown Osaka Scene Text (DOST) Dataset~\cite{Iwamura_IWRR2016} / ICDAR 2017 Robust Reading Challenge (RRC) on Omnidirectional Video (DOST)~\cite{ICDAR_DOST2017}}
    \label{fig:DOST}
  \end{minipage}

  \caption{Sample images of databases \#1.} 
  \label{fig:db_sample1}
\end{figure}


\begin{figure}[tbp]
  \centering
  \begin{minipage}[b]{\hsize}
    \centering    
    \includegraphics[width=.115\hsize]{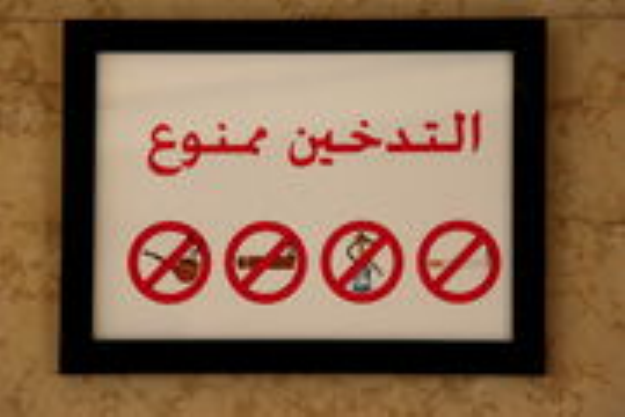}
    \includegraphics[width=.115\hsize]{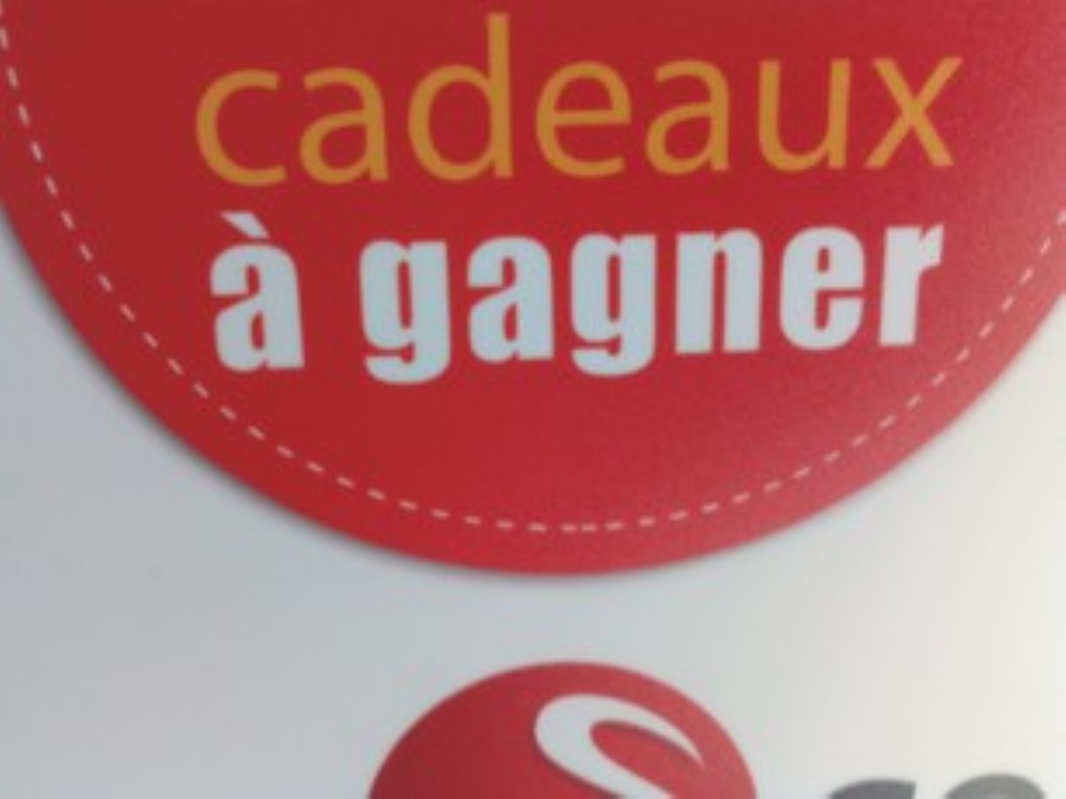}
    \includegraphics[width=.115\hsize]{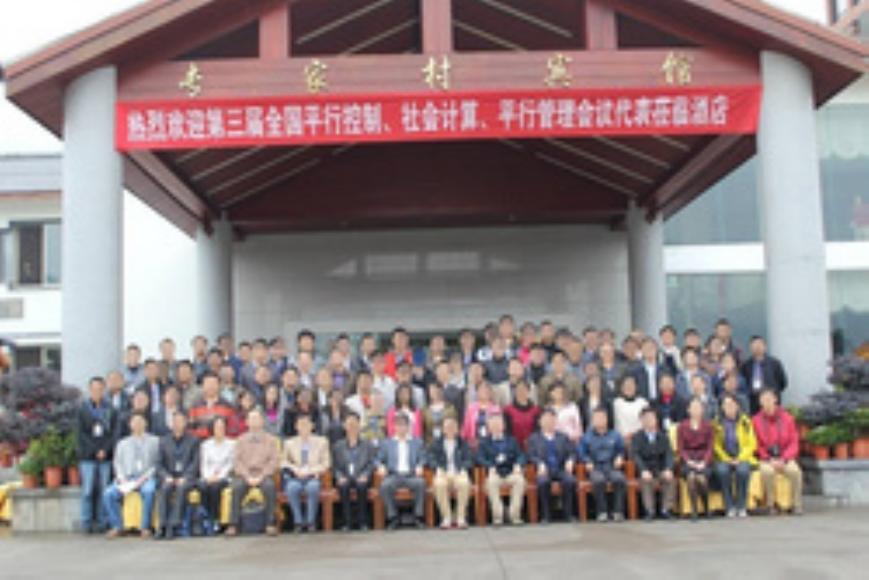}
    \includegraphics[width=.115\hsize]{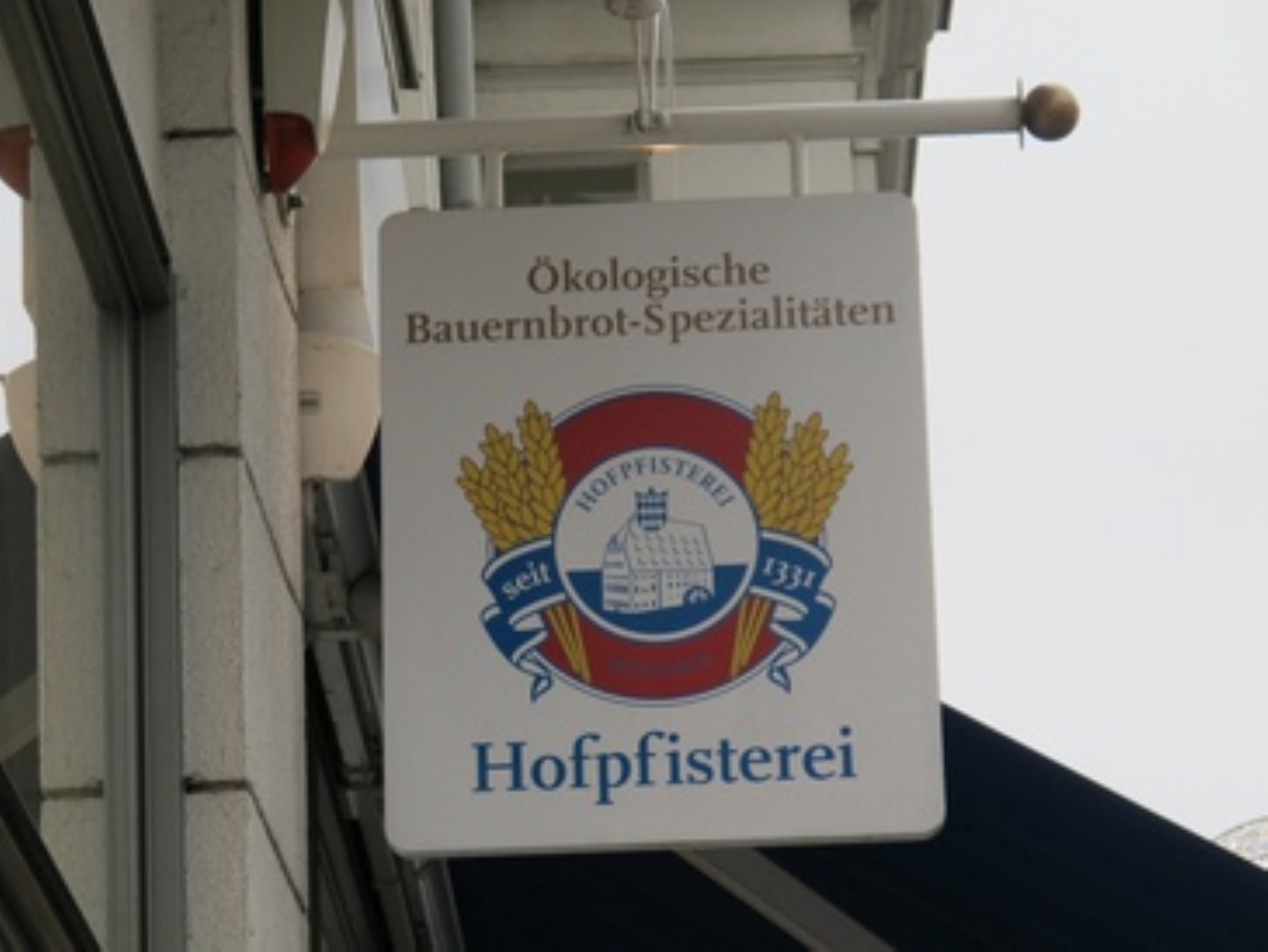}
    \includegraphics[width=.115\hsize]{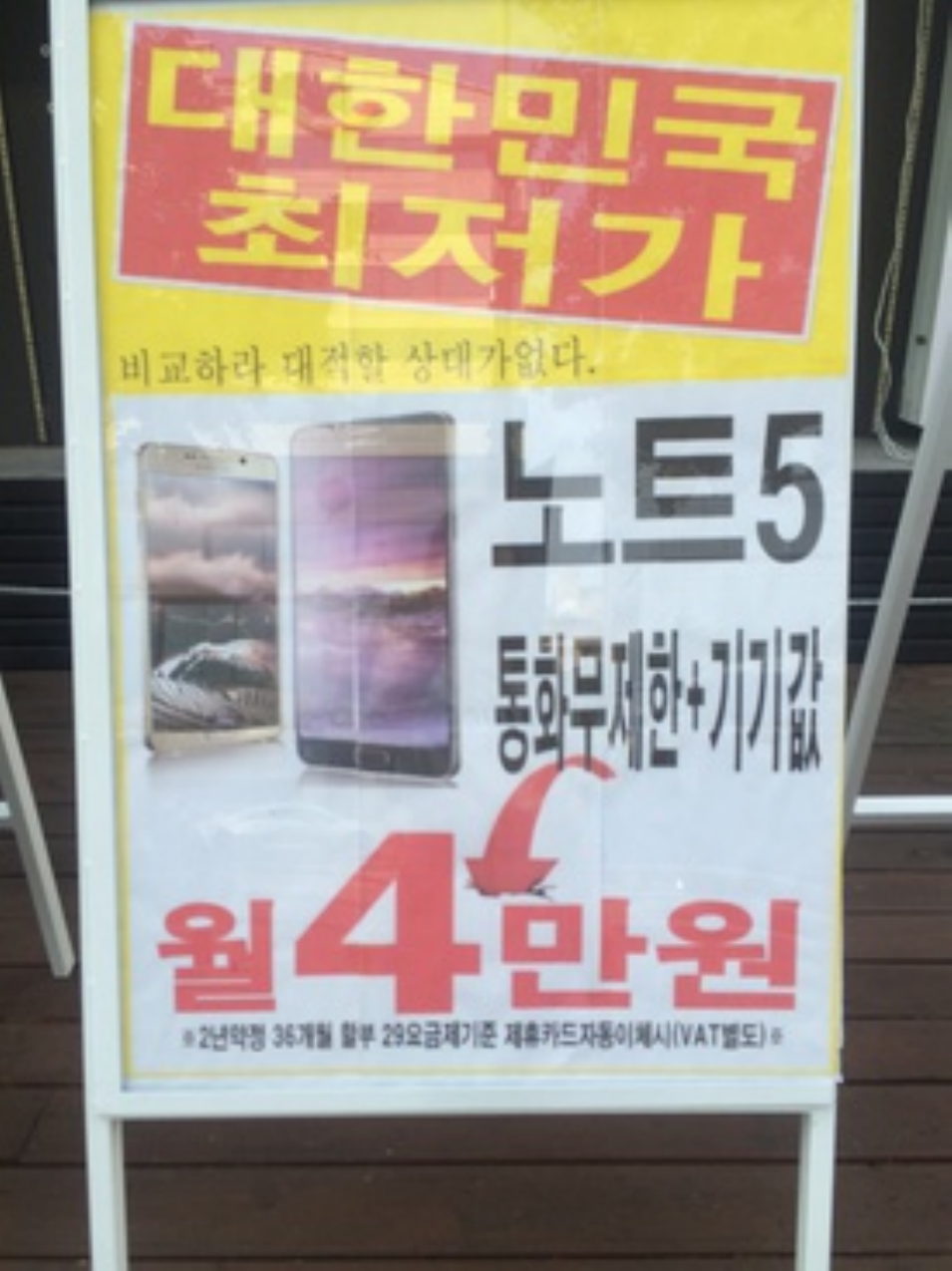}
    \includegraphics[width=.115\hsize]{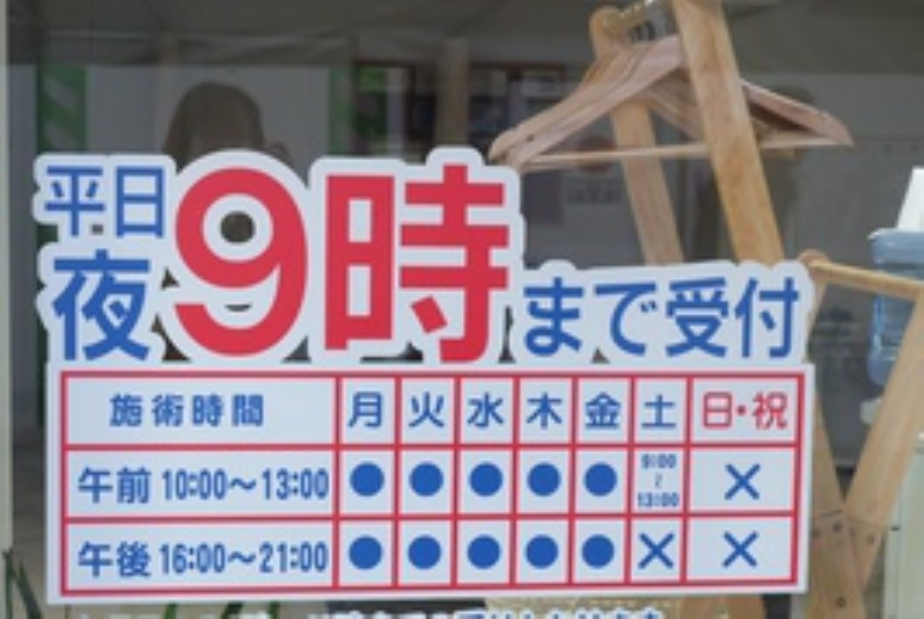}
    \includegraphics[width=.115\hsize]{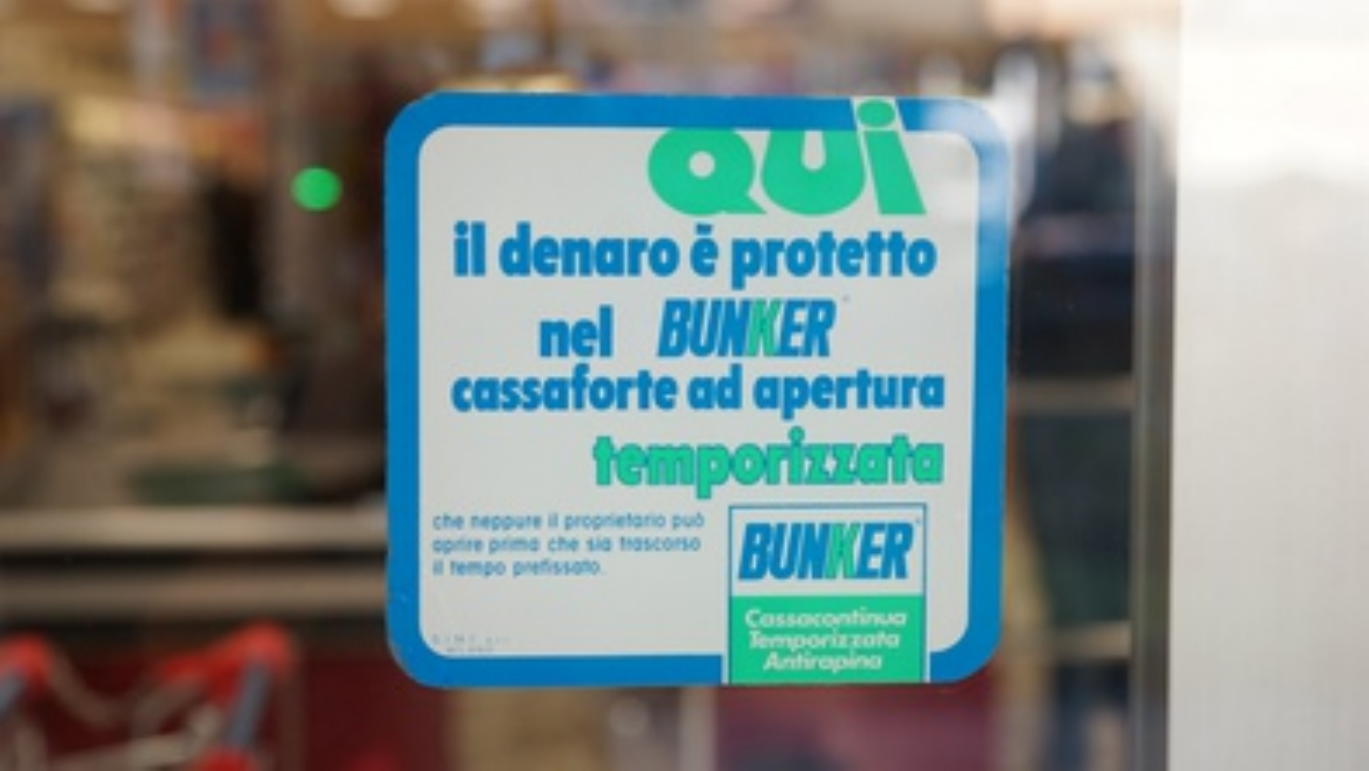}
    \includegraphics[width=.115\hsize]{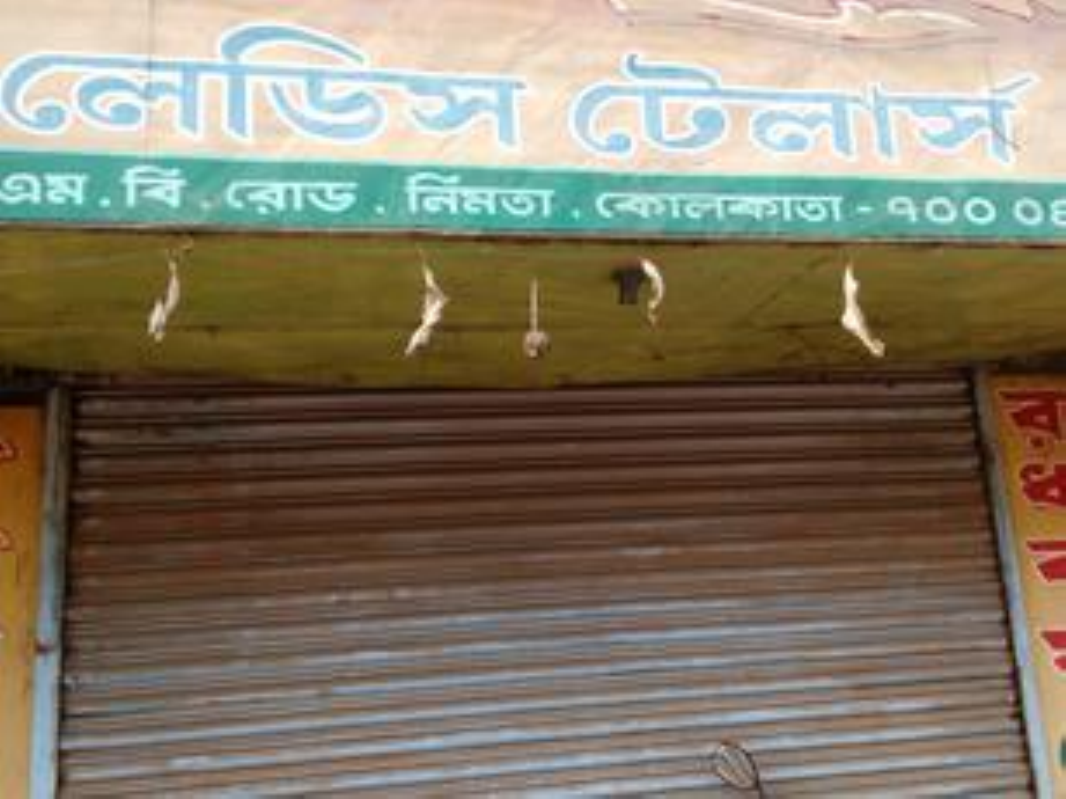}
    \subcaption{ICDAR2017 Competition on Multi-lingual Scene Text Detection and Script Identification (MLT) dataset~\cite{ICDAR_MLT2017}}
    \label{fig:ICDAR2017_MLT}
  \end{minipage}
  \begin{minipage}[b]{\hsize}
    \centering    
    \includegraphics[width=.115\hsize]{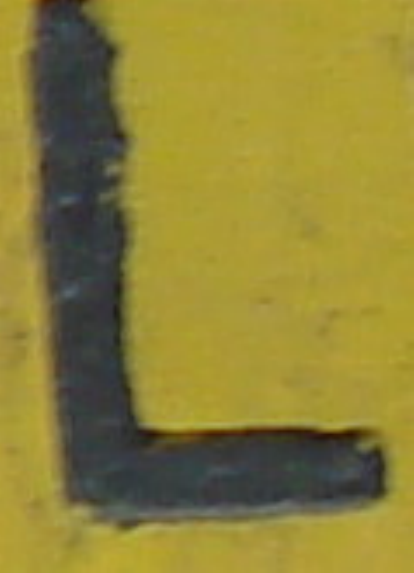}
    \includegraphics[width=.115\hsize]{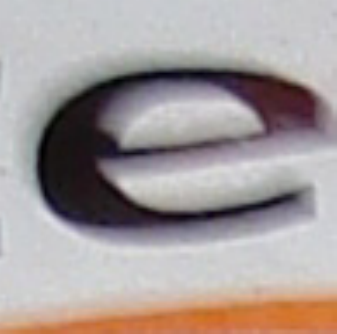}
    \includegraphics[width=.115\hsize]{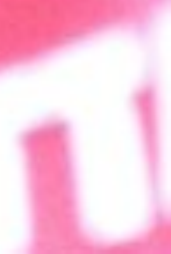}
    \includegraphics[width=.115\hsize]{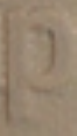}
    \includegraphics[width=.115\hsize]{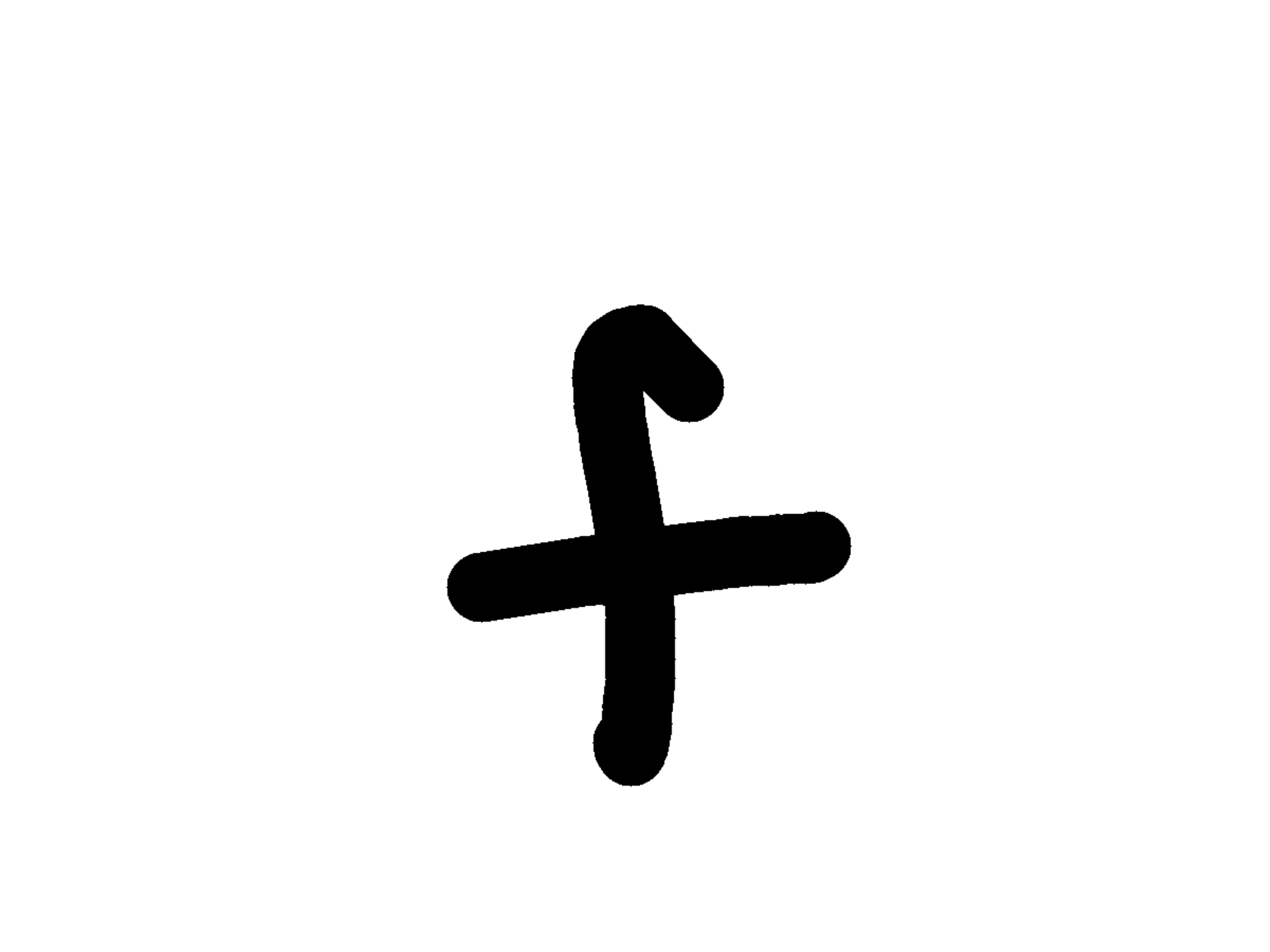}
    \includegraphics[width=.115\hsize]{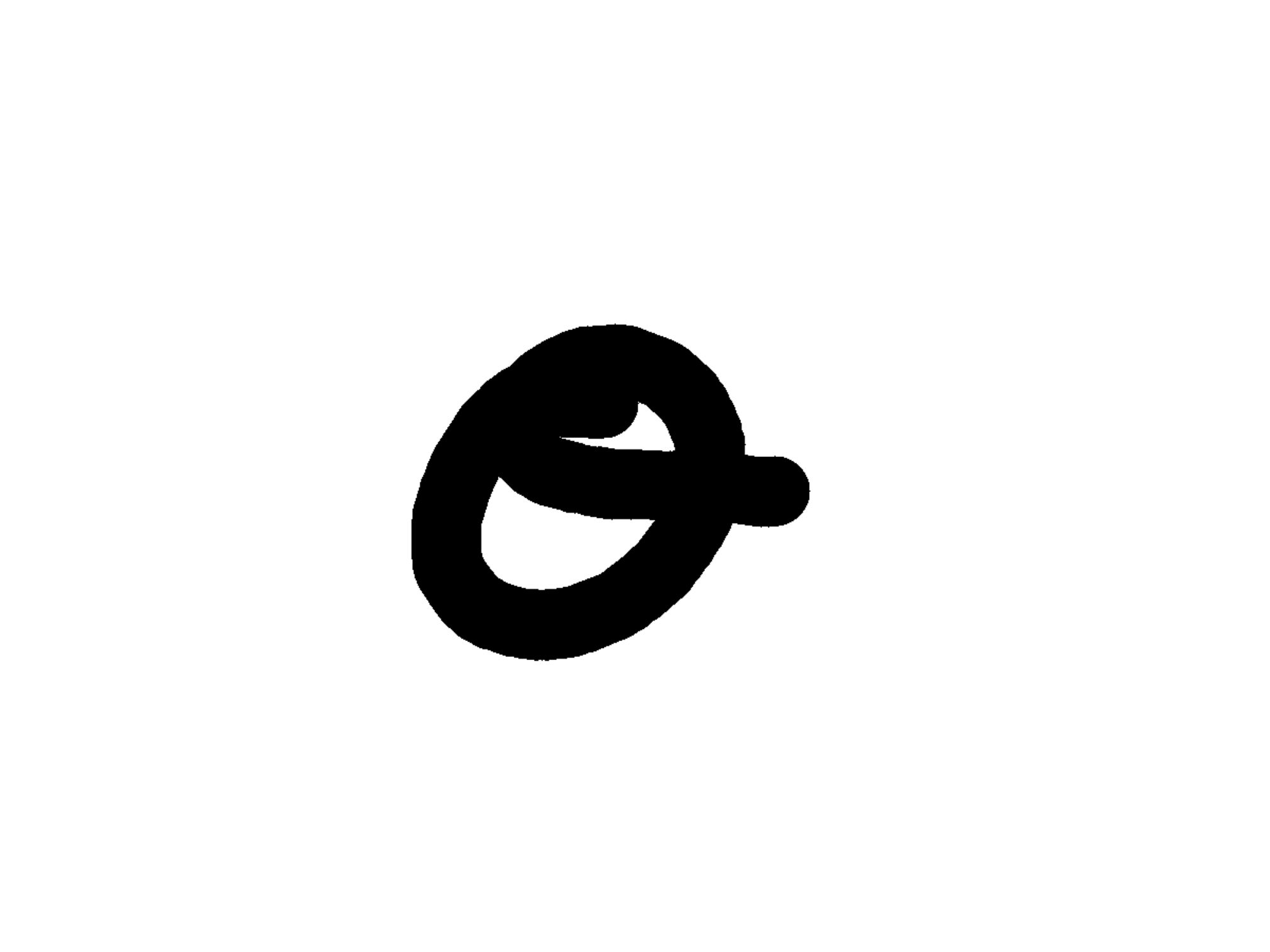}
    \includegraphics[width=.115\hsize]{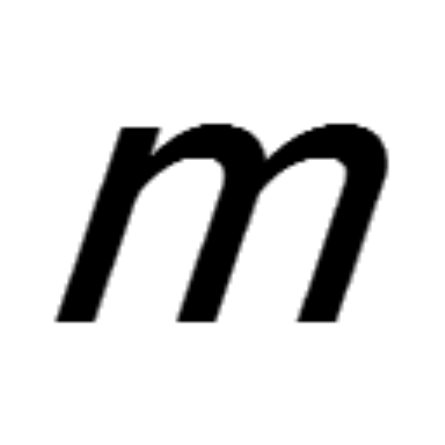}
    \includegraphics[width=.115\hsize]{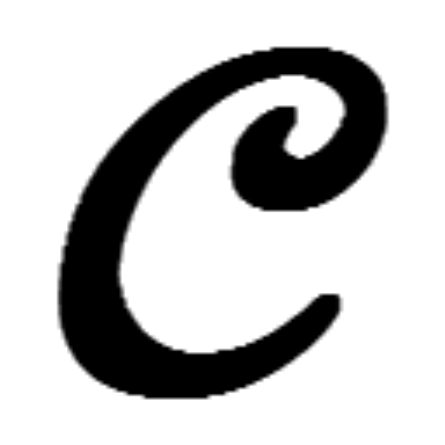}
    \subcaption{Chars74k Dataset~\cite{deCampos_VISAPP2009}}
    \label{fig:Chars74k}
  \end{minipage}
  \begin{minipage}[b]{\hsize}
    \centering    
    \includegraphics[width=.115\hsize]{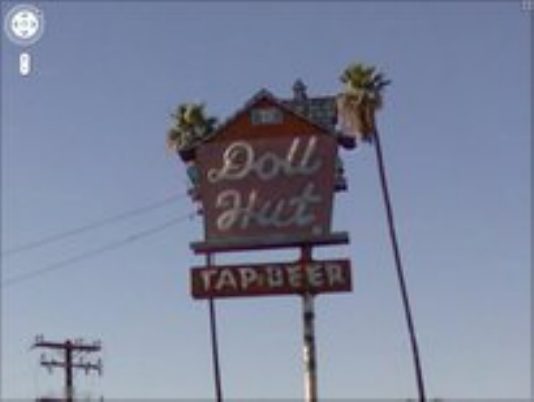}
    \includegraphics[width=.115\hsize]{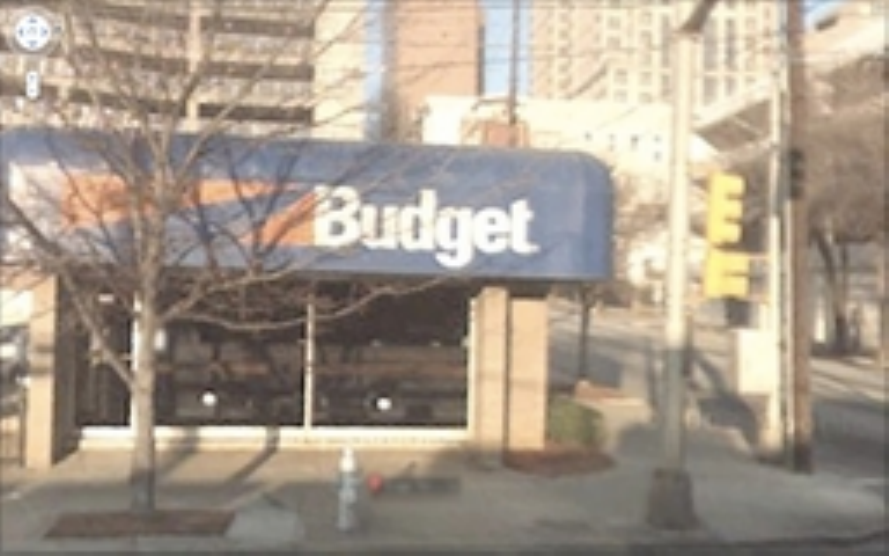}
    \includegraphics[width=.115\hsize]{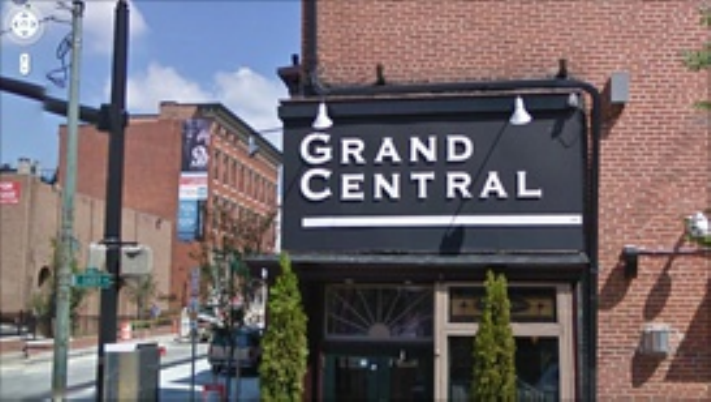}
    \includegraphics[width=.115\hsize]{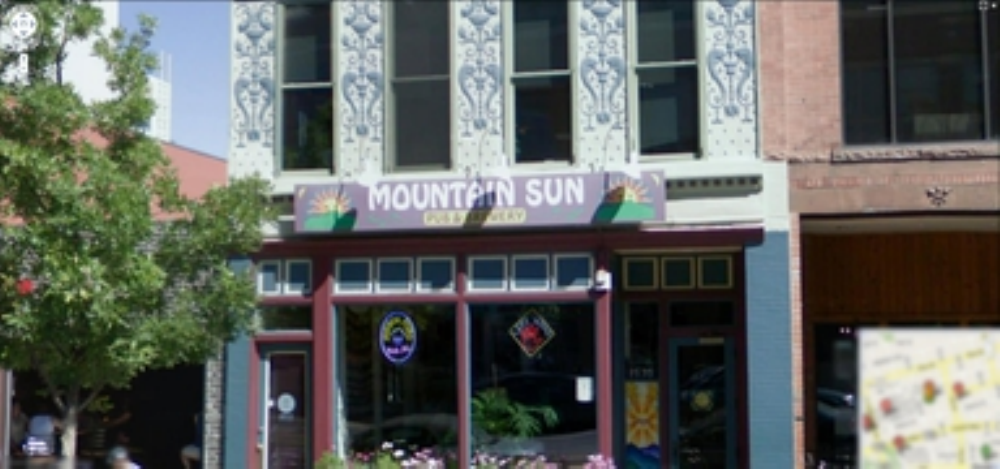}
    \includegraphics[width=.115\hsize]{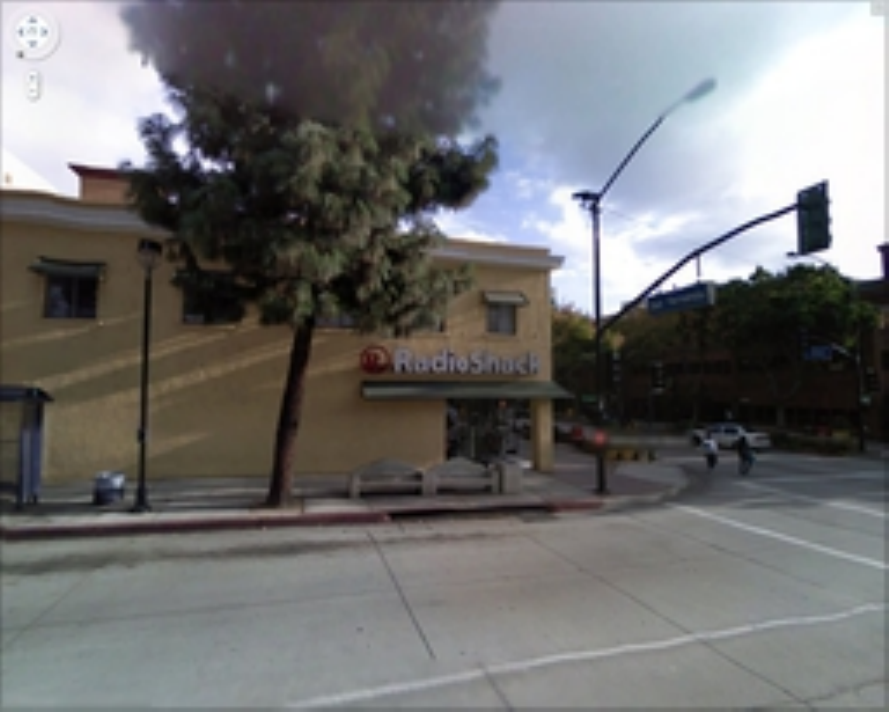}
    \includegraphics[width=.115\hsize]{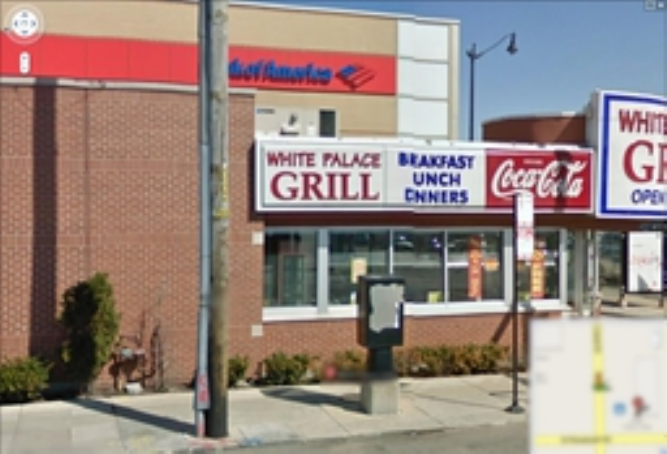}
    \includegraphics[width=.115\hsize]{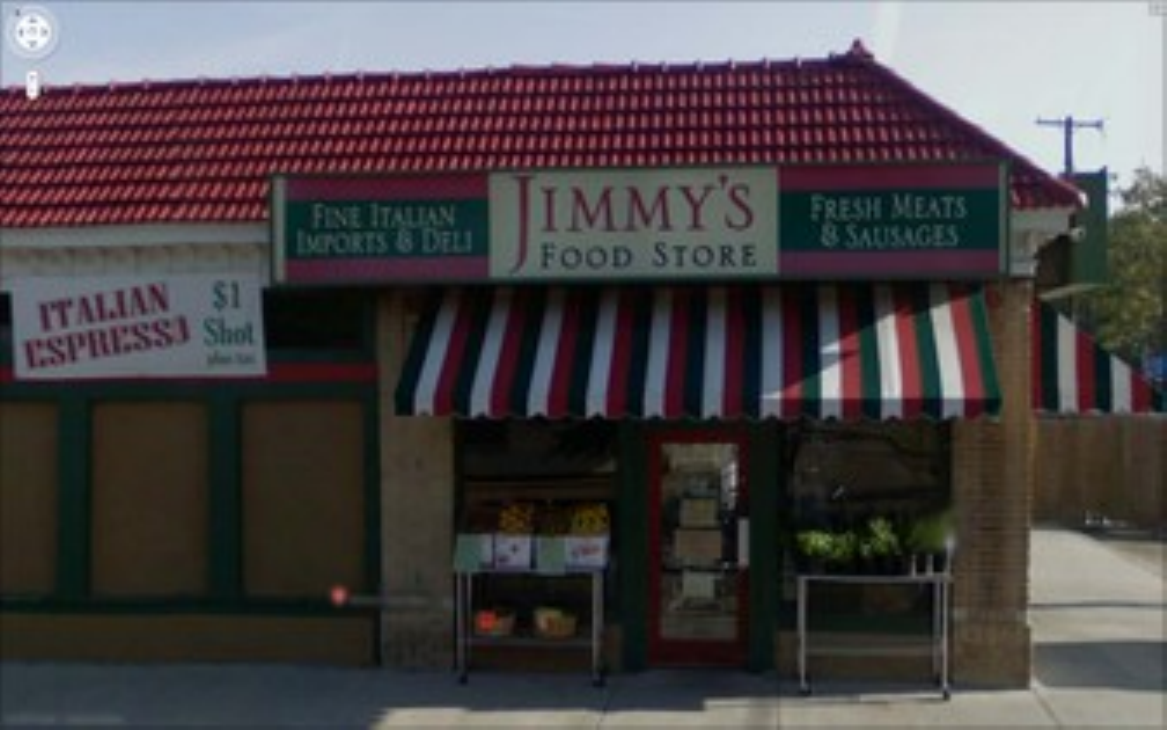}
    \includegraphics[width=.115\hsize]{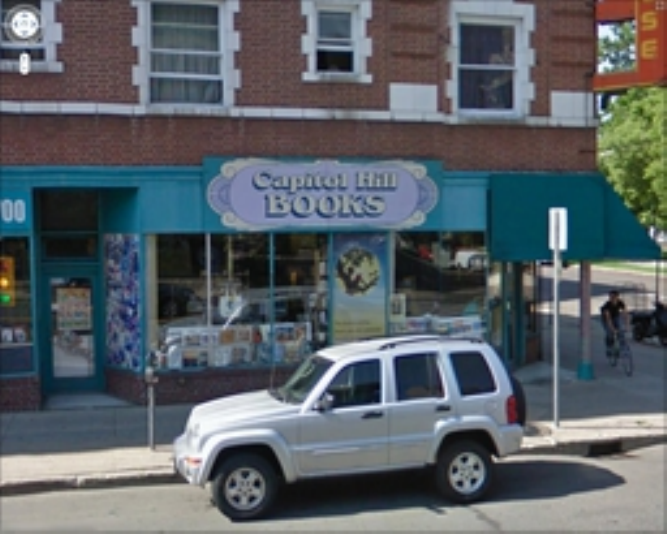}
    \subcaption{Street View Text (SVT) Dataset~\cite{Wang_ECCV2010,Wang_ICCV2011}}
    \label{fig:SVT}
  \end{minipage}
  \begin{minipage}[b]{\hsize}
    \centering    
    \includegraphics[width=.115\hsize]{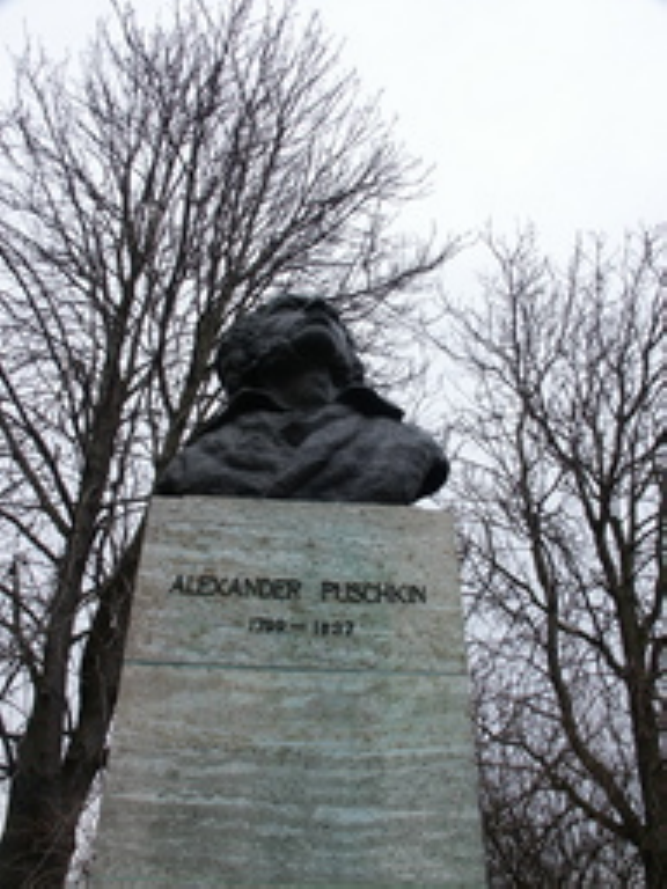}
    \includegraphics[width=.115\hsize]{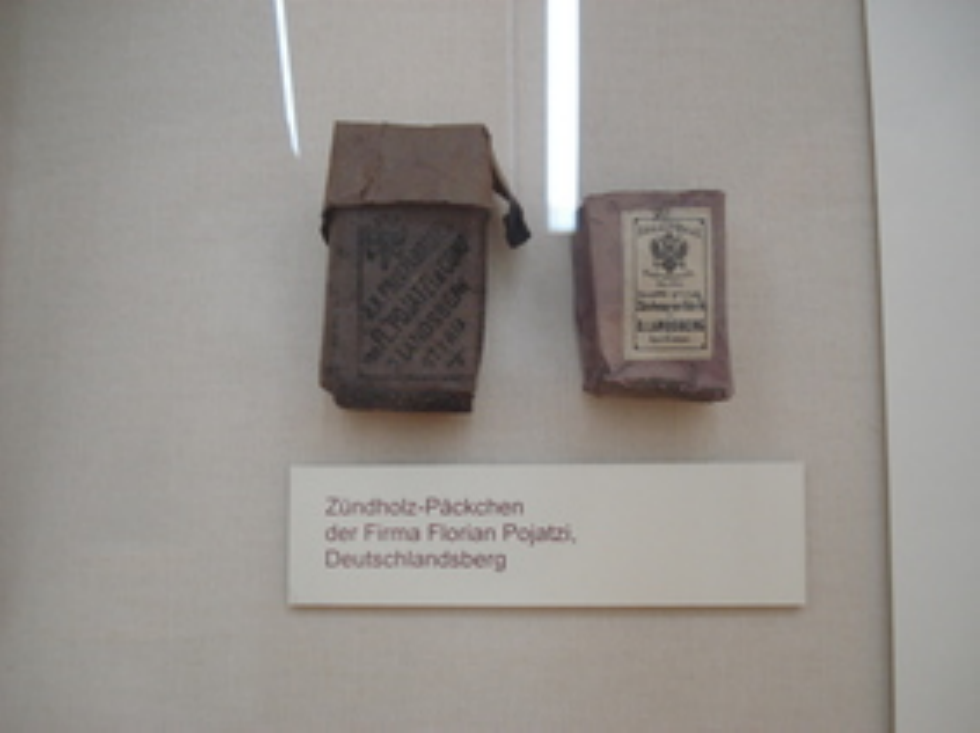}
    \includegraphics[width=.115\hsize]{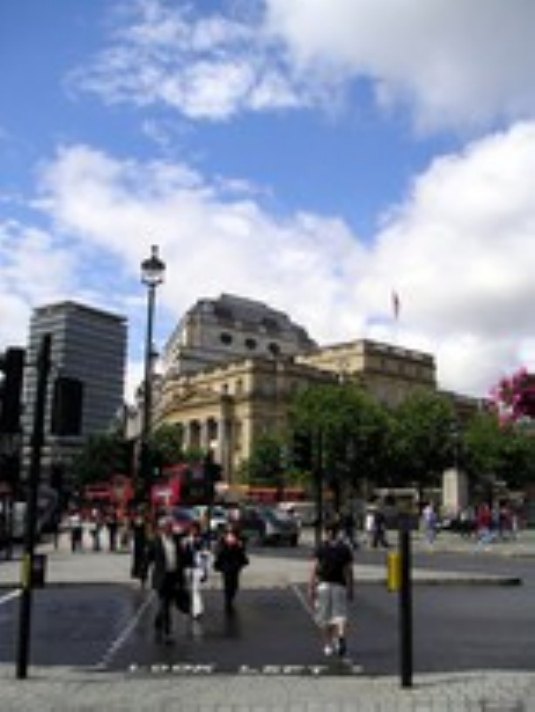}
    \includegraphics[width=.115\hsize]{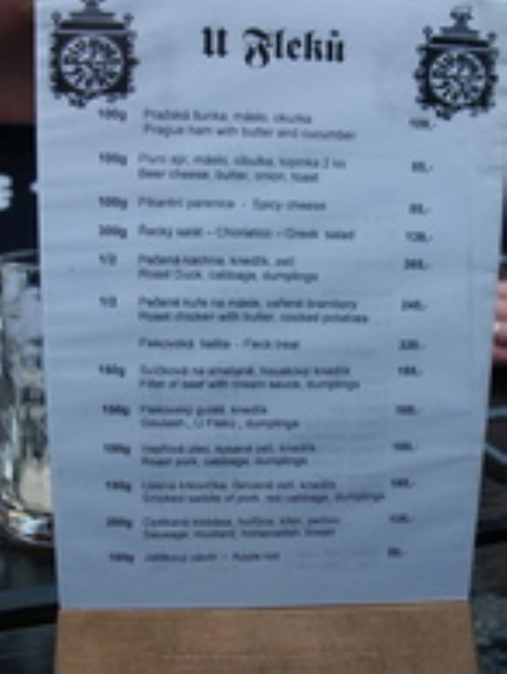}
    \includegraphics[width=.115\hsize]{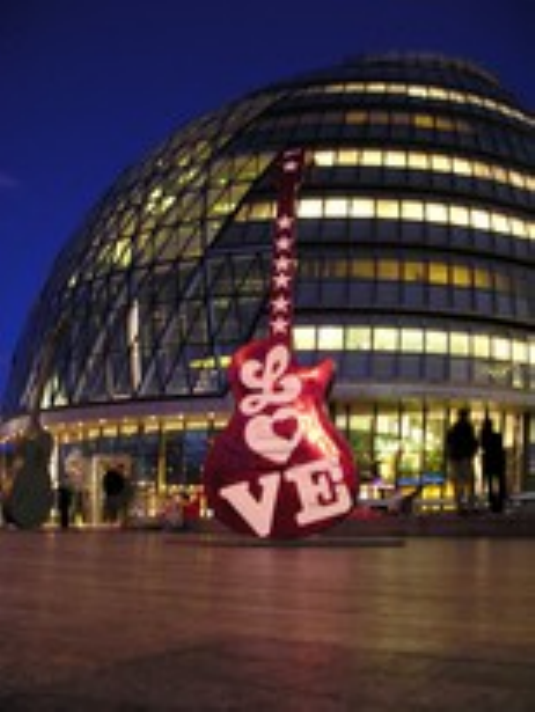}
    \includegraphics[width=.115\hsize]{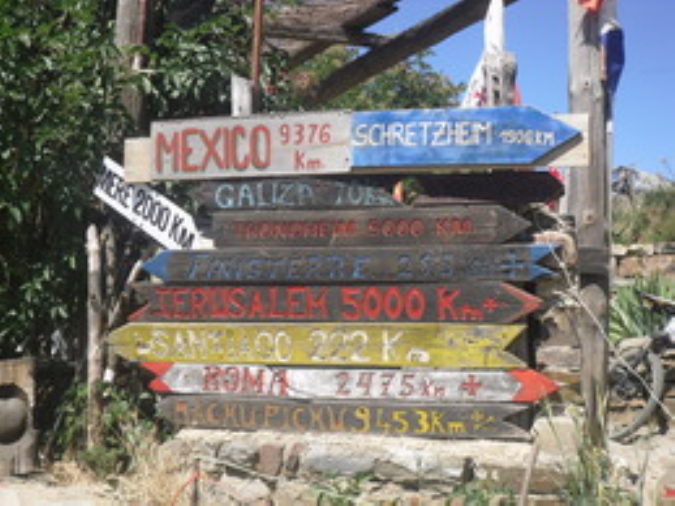}
    \includegraphics[width=.115\hsize]{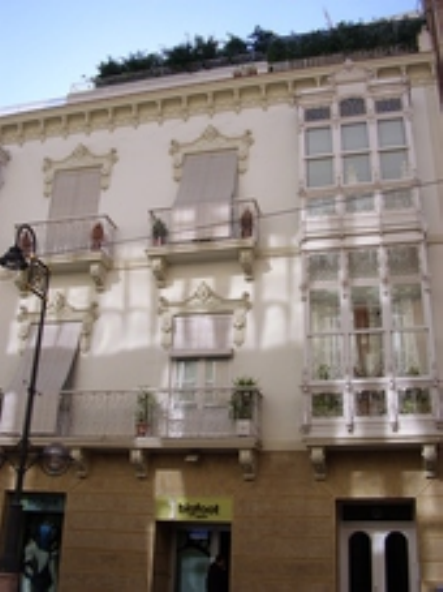}
    \includegraphics[width=.115\hsize]{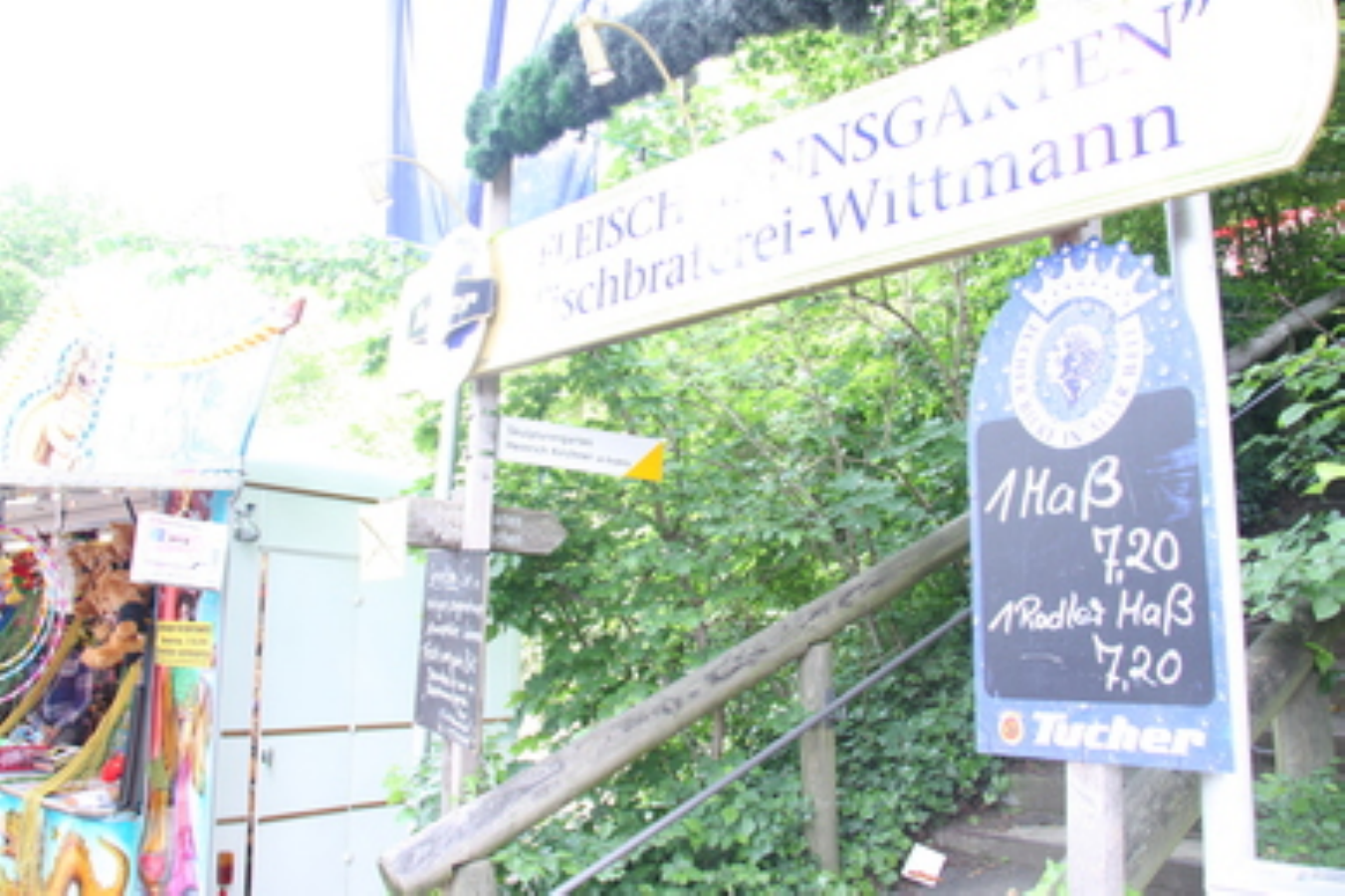}
    \subcaption{Natural Environment OCR (NEOCR) Dataset~\cite{Nagy_CBDAR2012}}
    \label{fig:NEOCR}
  \end{minipage}
  \begin{minipage}[b]{\hsize}
    \centering    
    \includegraphics[width=.115\hsize]{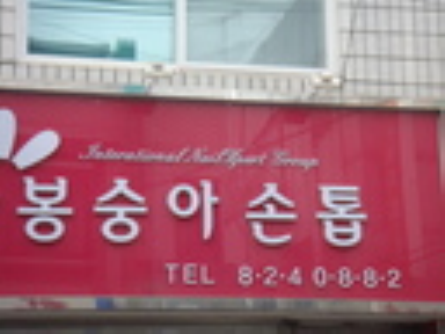}
    \includegraphics[width=.115\hsize]{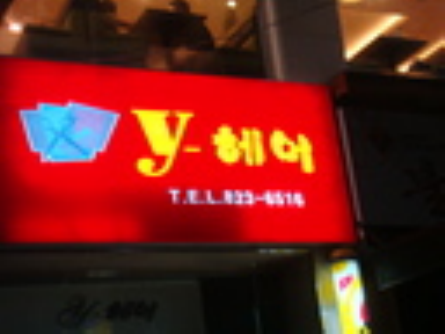}
    \includegraphics[width=.115\hsize]{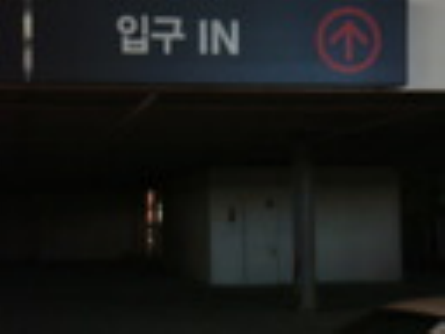}
    \includegraphics[width=.115\hsize]{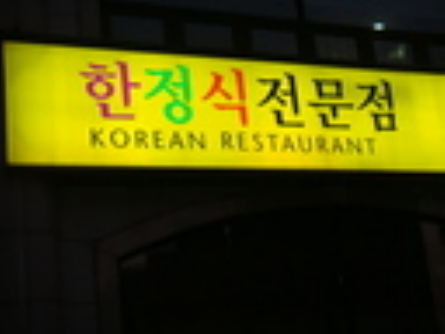}
    \includegraphics[width=.115\hsize]{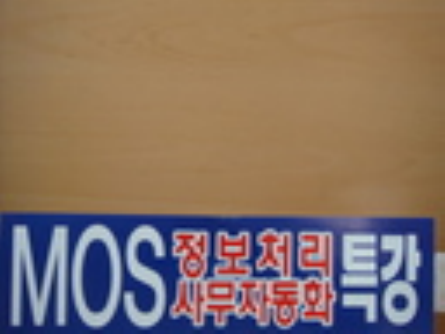}
    \includegraphics[width=.115\hsize]{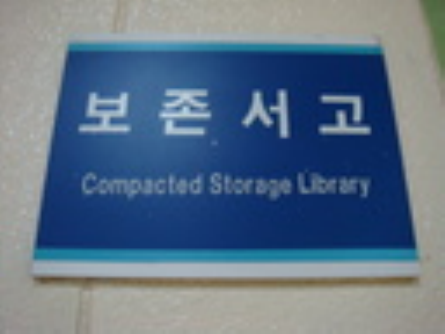}
    \includegraphics[width=.115\hsize]{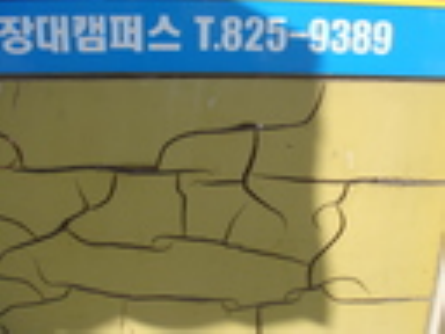}
    \includegraphics[width=.115\hsize]{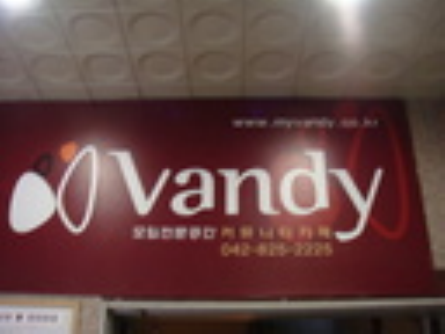}
    \subcaption{KAIST Scene Text Database~\cite{Jung_ETRI2011}}
    \label{fig:KAIST}
  \end{minipage}
  \begin{minipage}[b]{\hsize}
    \centering    
    \includegraphics[width=.115\hsize]{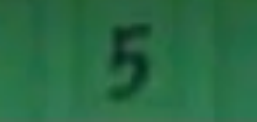}
    \includegraphics[width=.115\hsize]{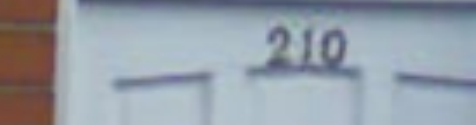}
    \includegraphics[width=.115\hsize]{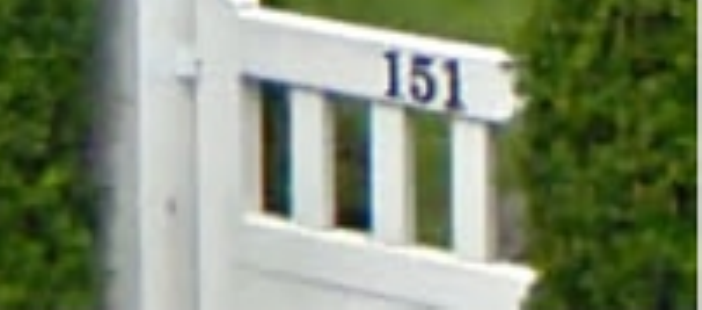}
    \includegraphics[width=.115\hsize]{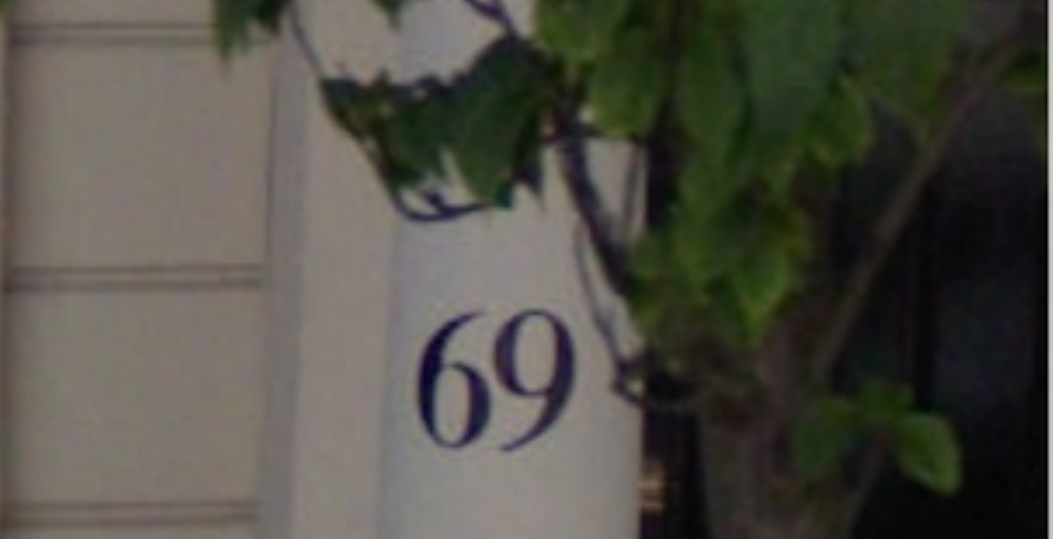}
    \includegraphics[width=.115\hsize]{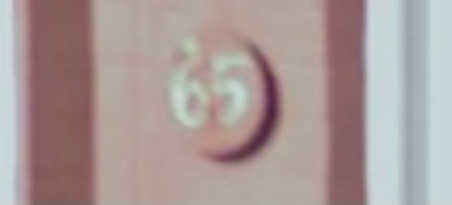}
    \includegraphics[width=.115\hsize]{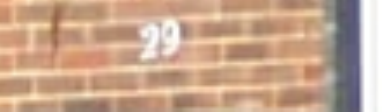}
    \includegraphics[width=.115\hsize]{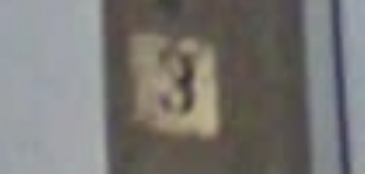}
    \includegraphics[width=.115\hsize]{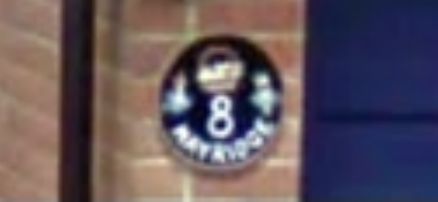}
    \subcaption{Street View House Numbers (SVHN) Dataset~\cite{svhn}}
    \label{fig:SVHN}
  \end{minipage}
  \begin{minipage}[b]{\hsize}
    \centering    
    \includegraphics[width=.115\hsize]{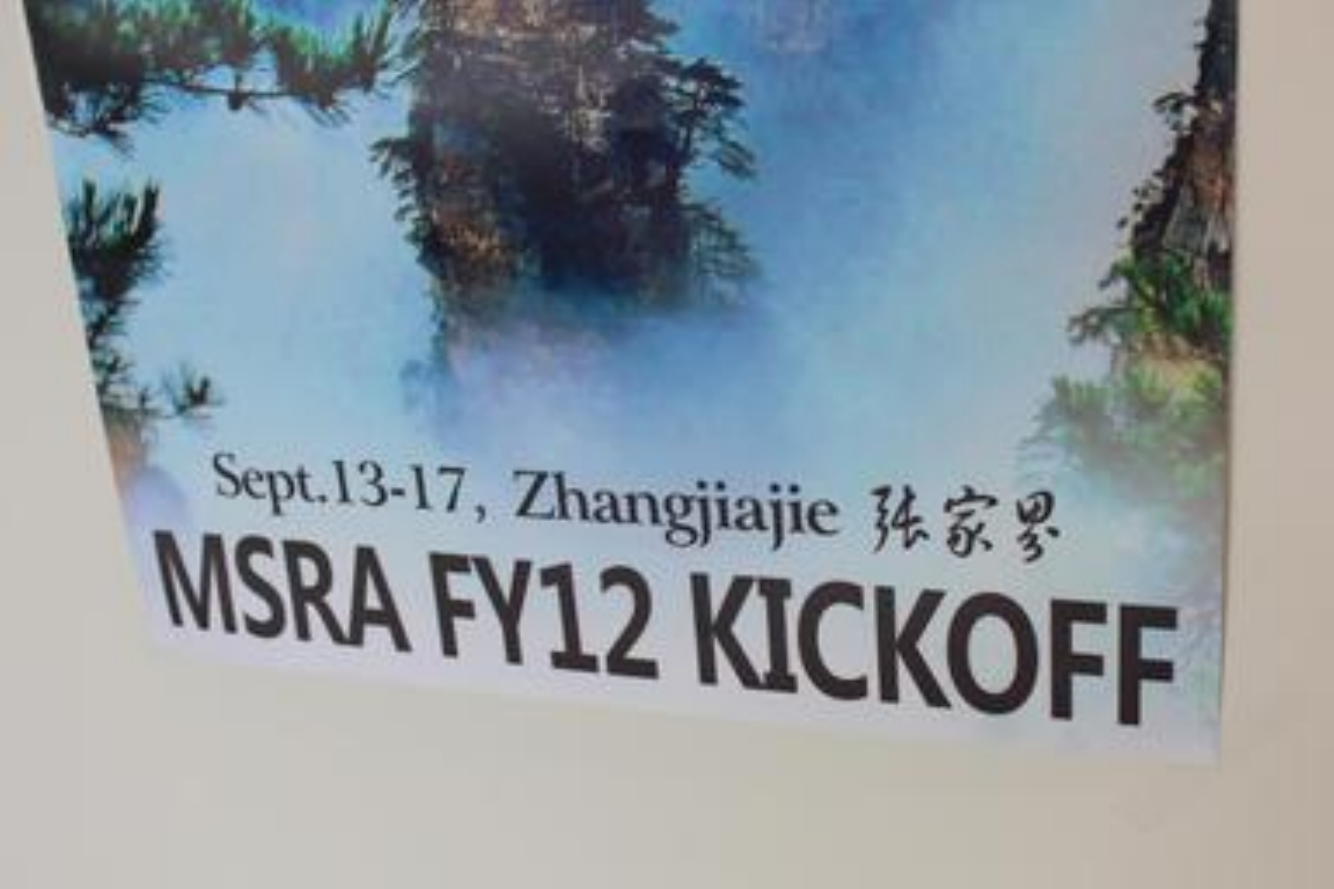}
    \includegraphics[width=.115\hsize]{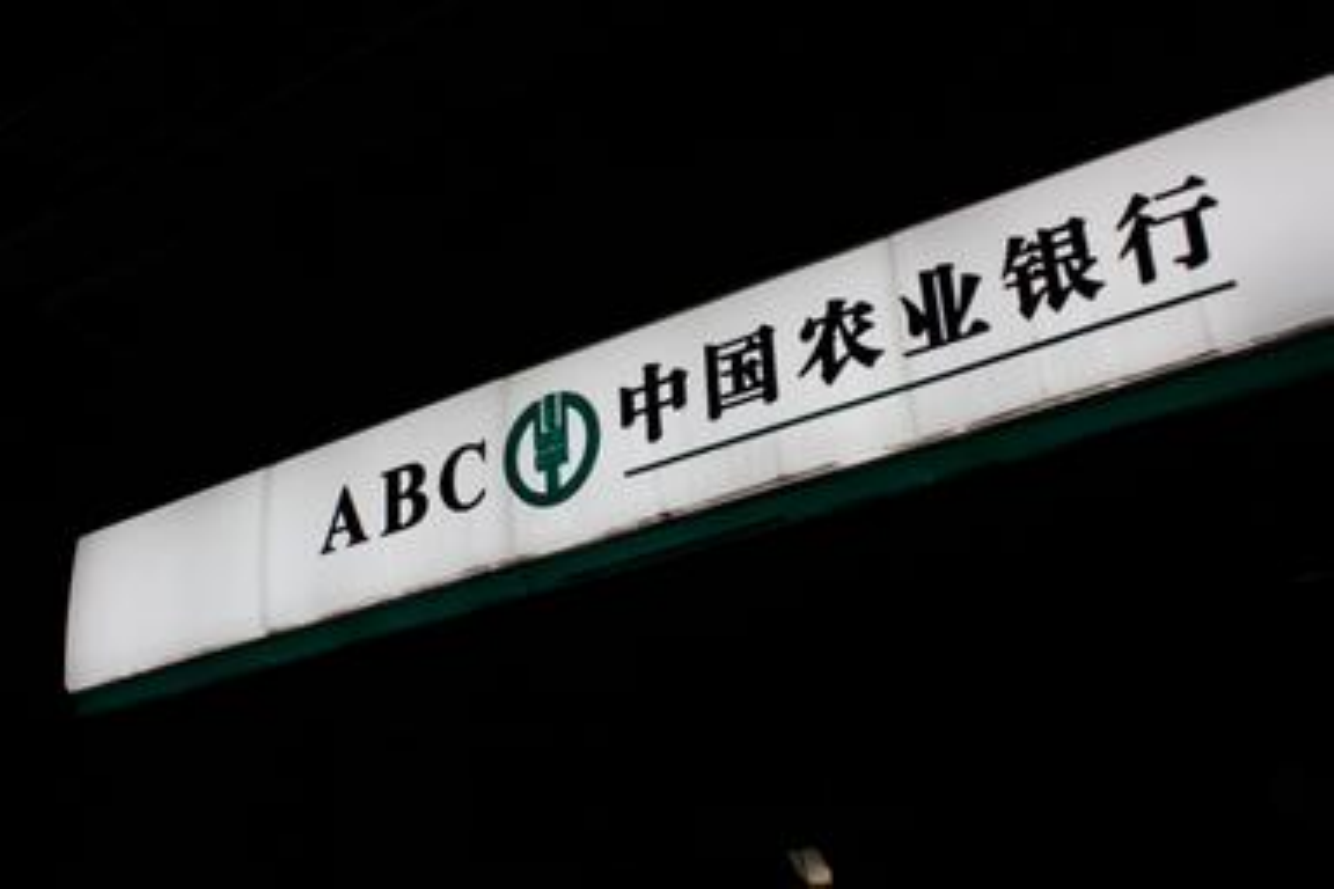}
    \includegraphics[width=.115\hsize]{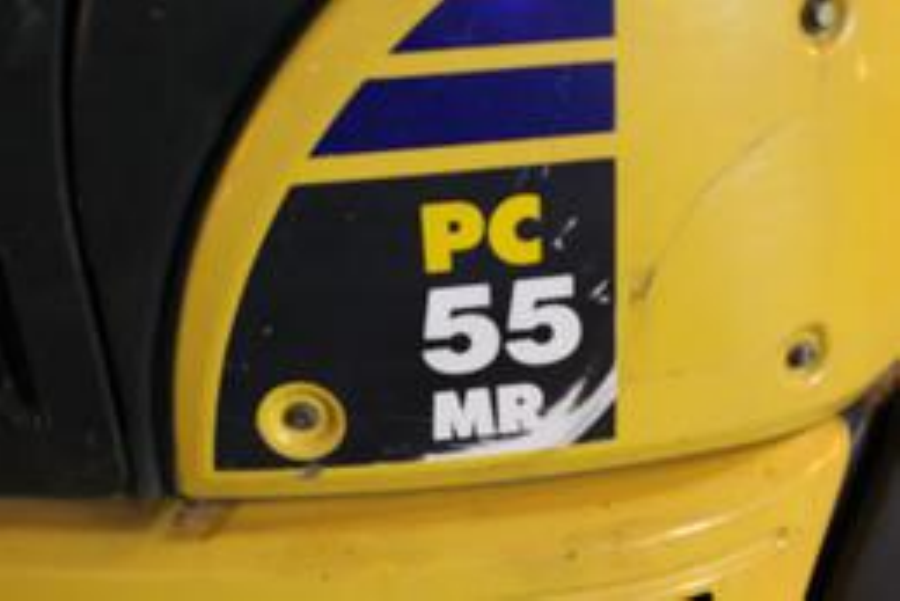}
    \includegraphics[width=.115\hsize]{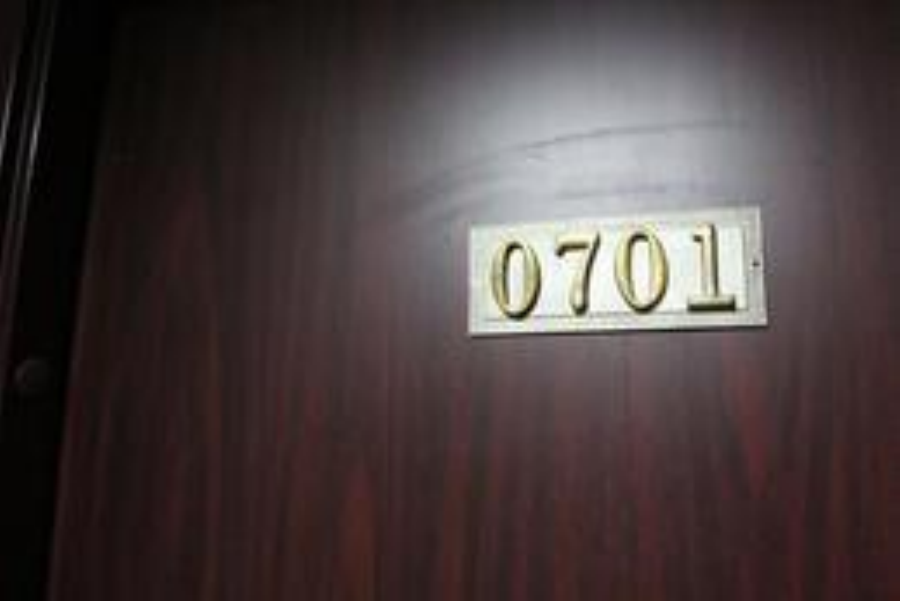}
    \includegraphics[width=.115\hsize]{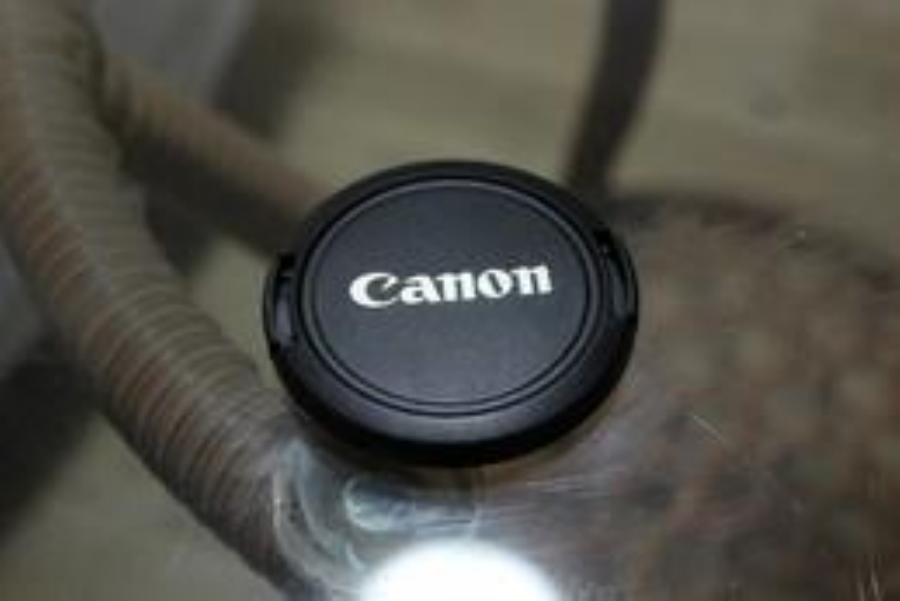}
    \includegraphics[width=.115\hsize]{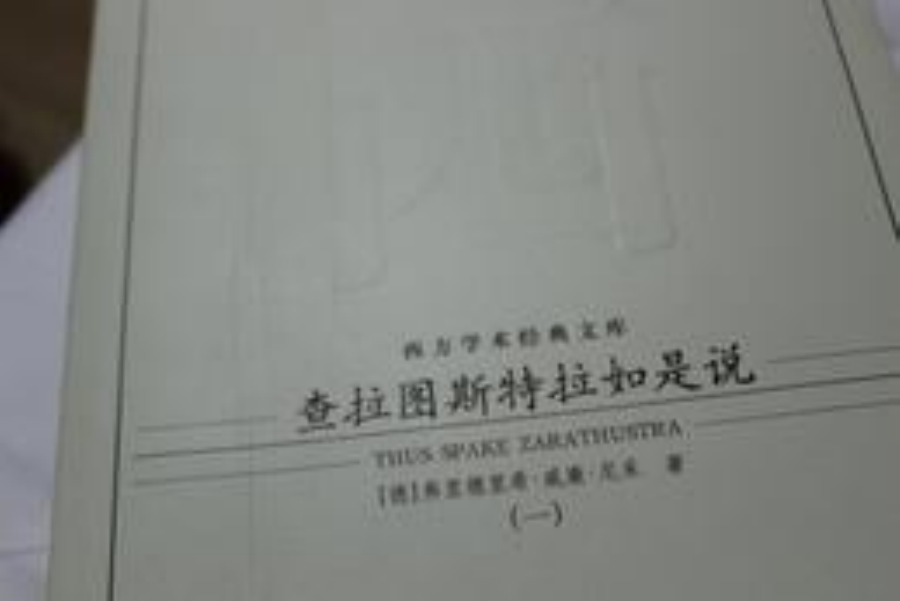}
    \includegraphics[width=.115\hsize]{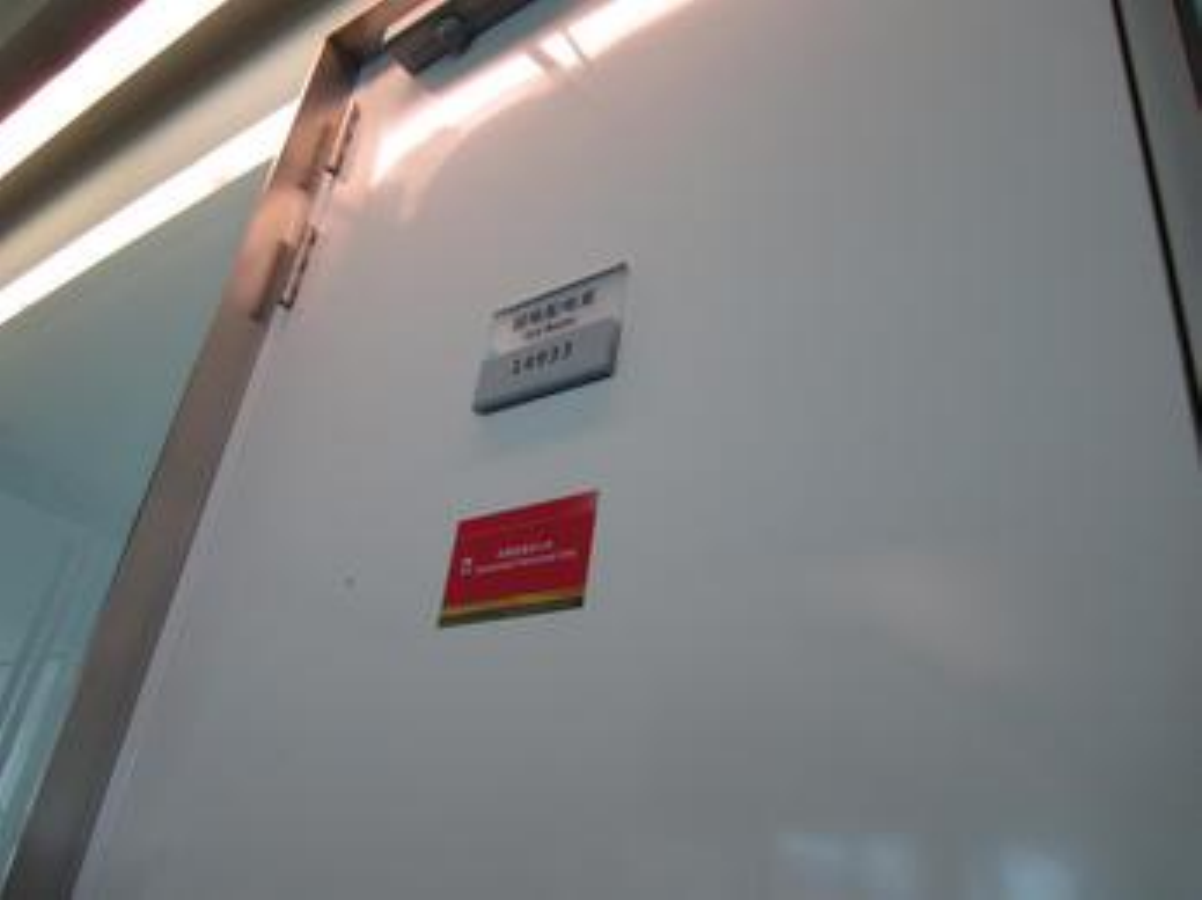}
    \includegraphics[width=.115\hsize]{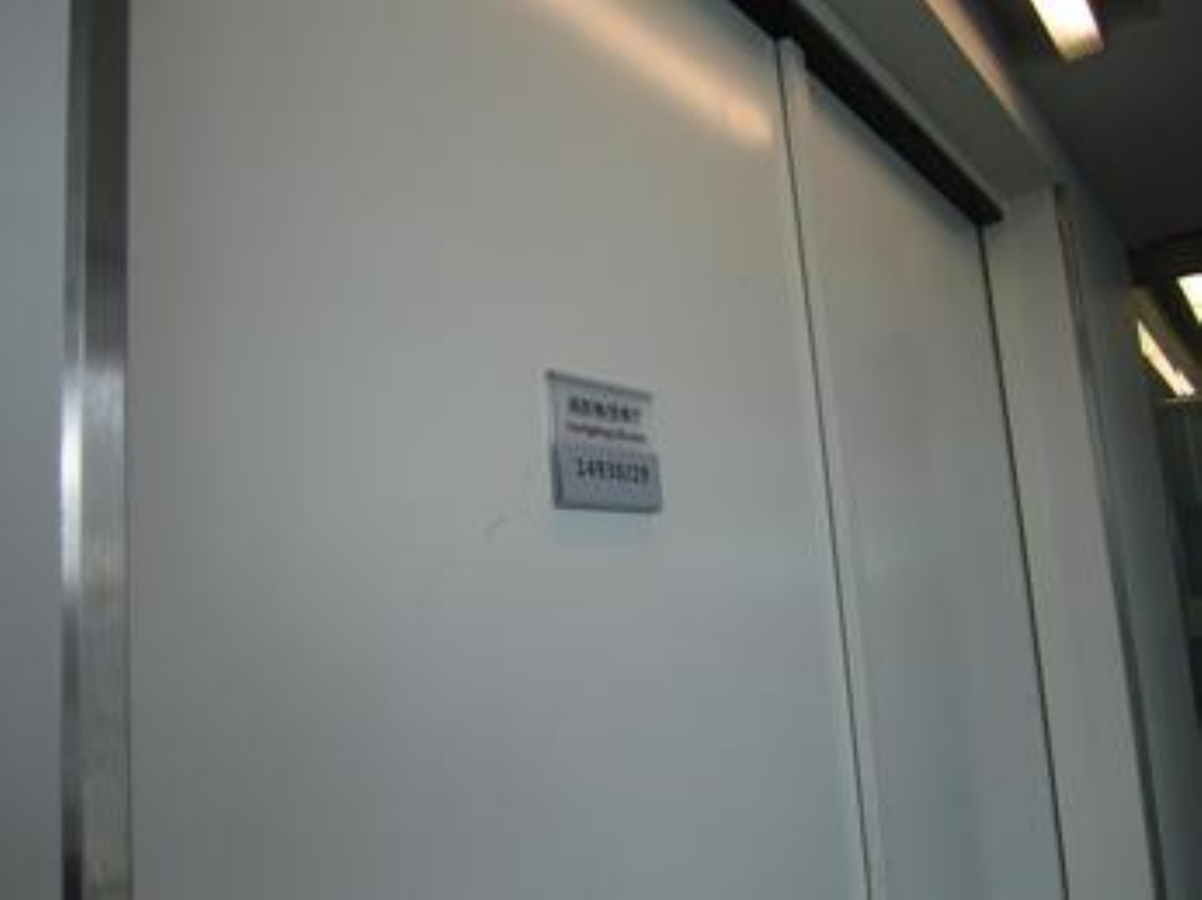}
    \subcaption{MSRA Text Detection 500 (MSRA-TD500) Database~\cite{Yao_CVPR2012}}
    \label{fig:MSRA-TD500}
  \end{minipage}
  \begin{minipage}[b]{\hsize}
    \centering    
    \includegraphics[width=.115\hsize]{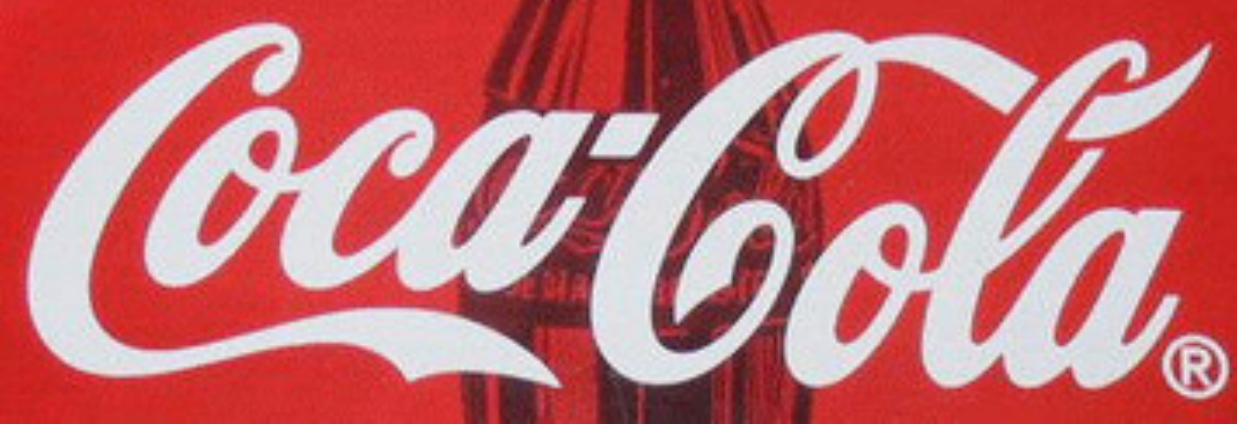}
    \includegraphics[width=.115\hsize]{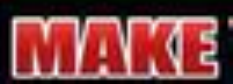}
    \includegraphics[width=.115\hsize]{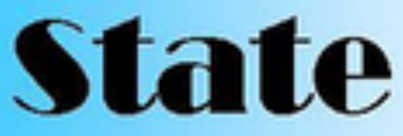}
    \includegraphics[width=.115\hsize]{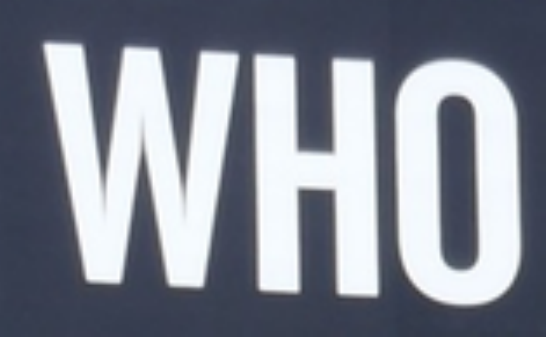}
    \includegraphics[width=.115\hsize]{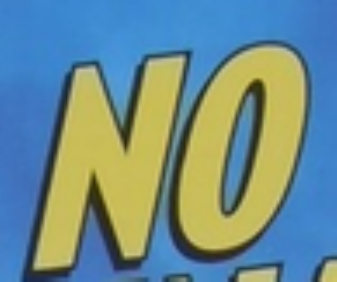}
    \includegraphics[width=.115\hsize]{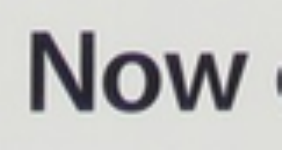}
    \includegraphics[width=.115\hsize]{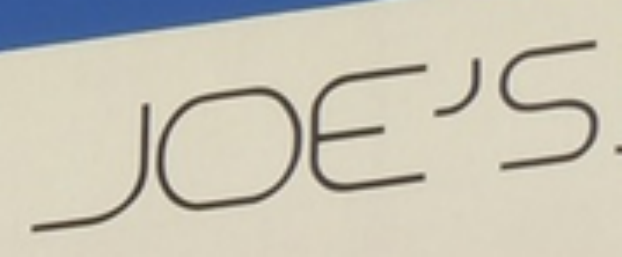}
    \includegraphics[width=.115\hsize]{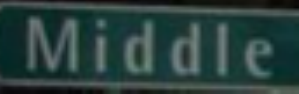}
    \subcaption{IIIT 5K-Word Dataset~\cite{Mishra_BMVC2012}}
    \label{fig:IIIT5K}
  \end{minipage}
  \begin{minipage}[b]{\hsize}
    \centering    
    \includegraphics[width=.115\hsize]{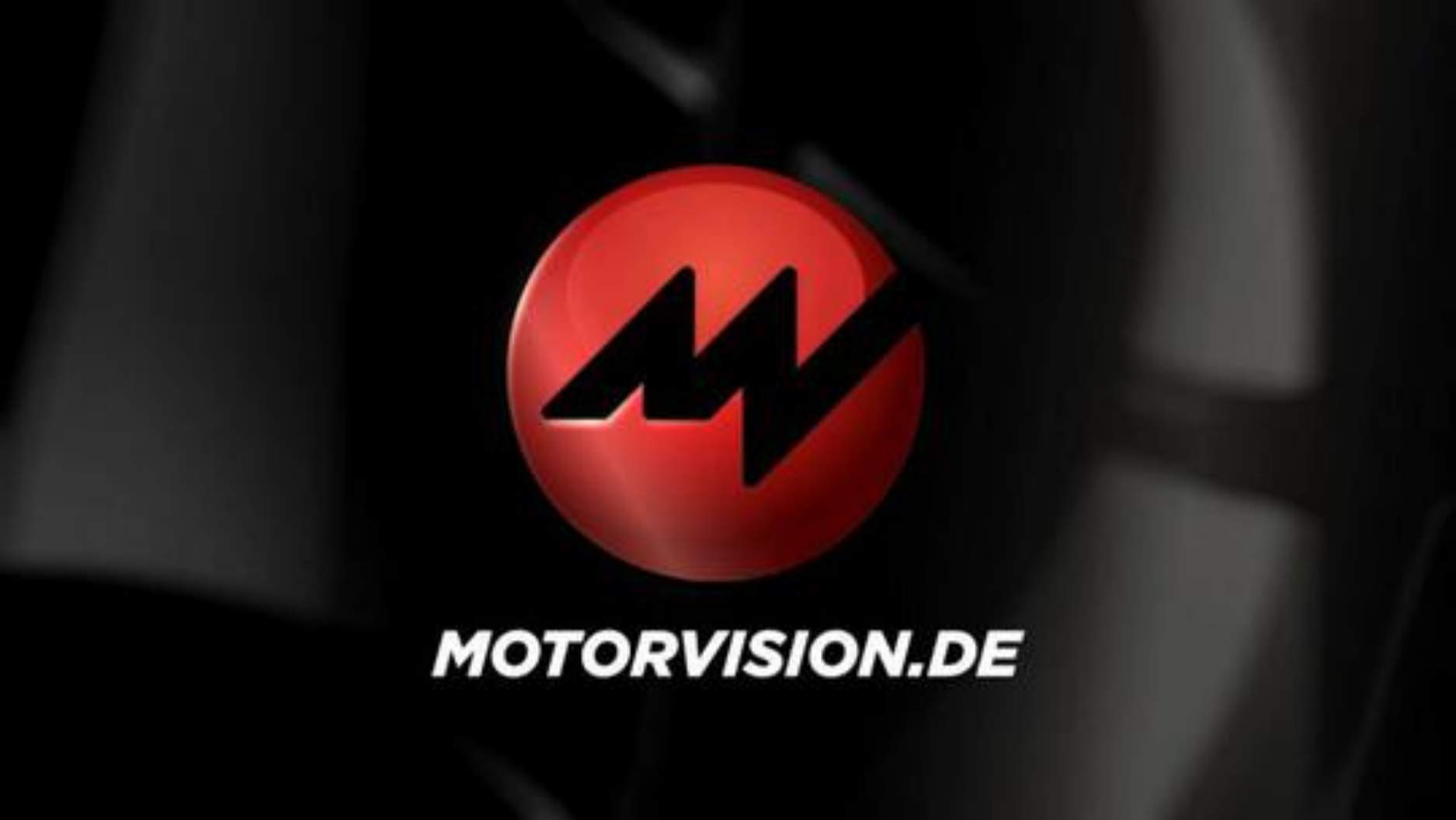}
    \includegraphics[width=.115\hsize]{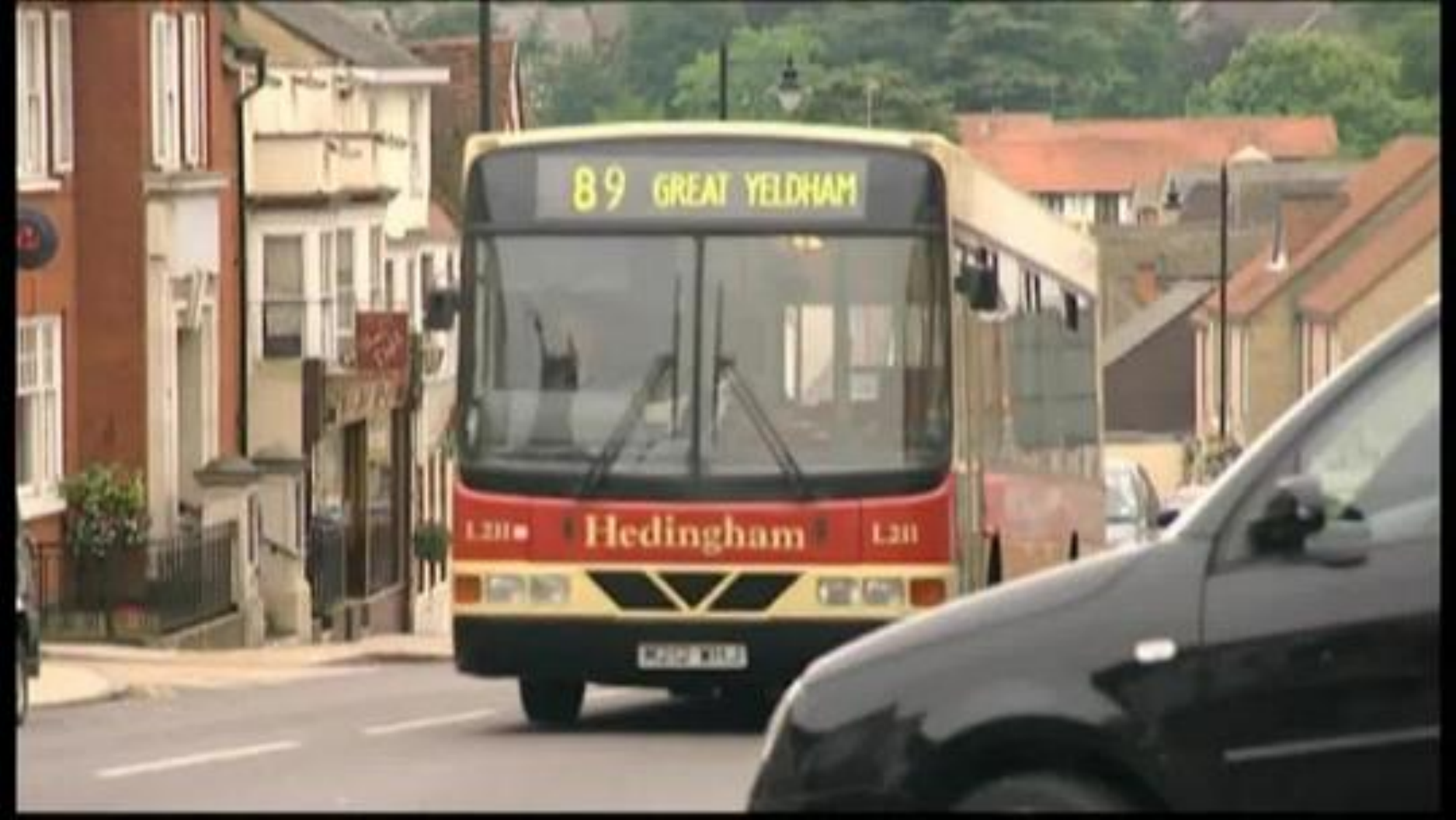}
    \includegraphics[width=.115\hsize]{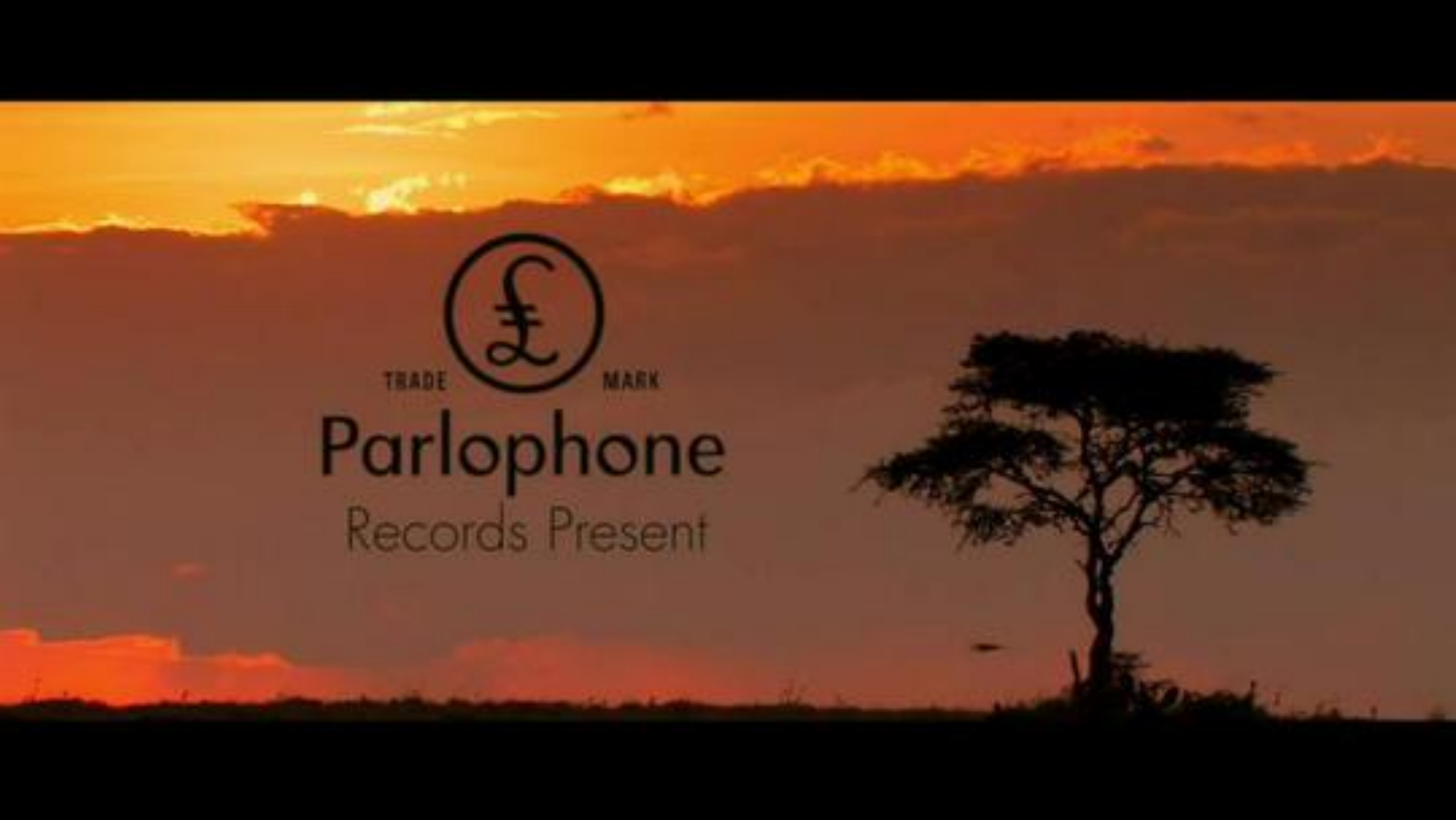}
    \includegraphics[width=.115\hsize]{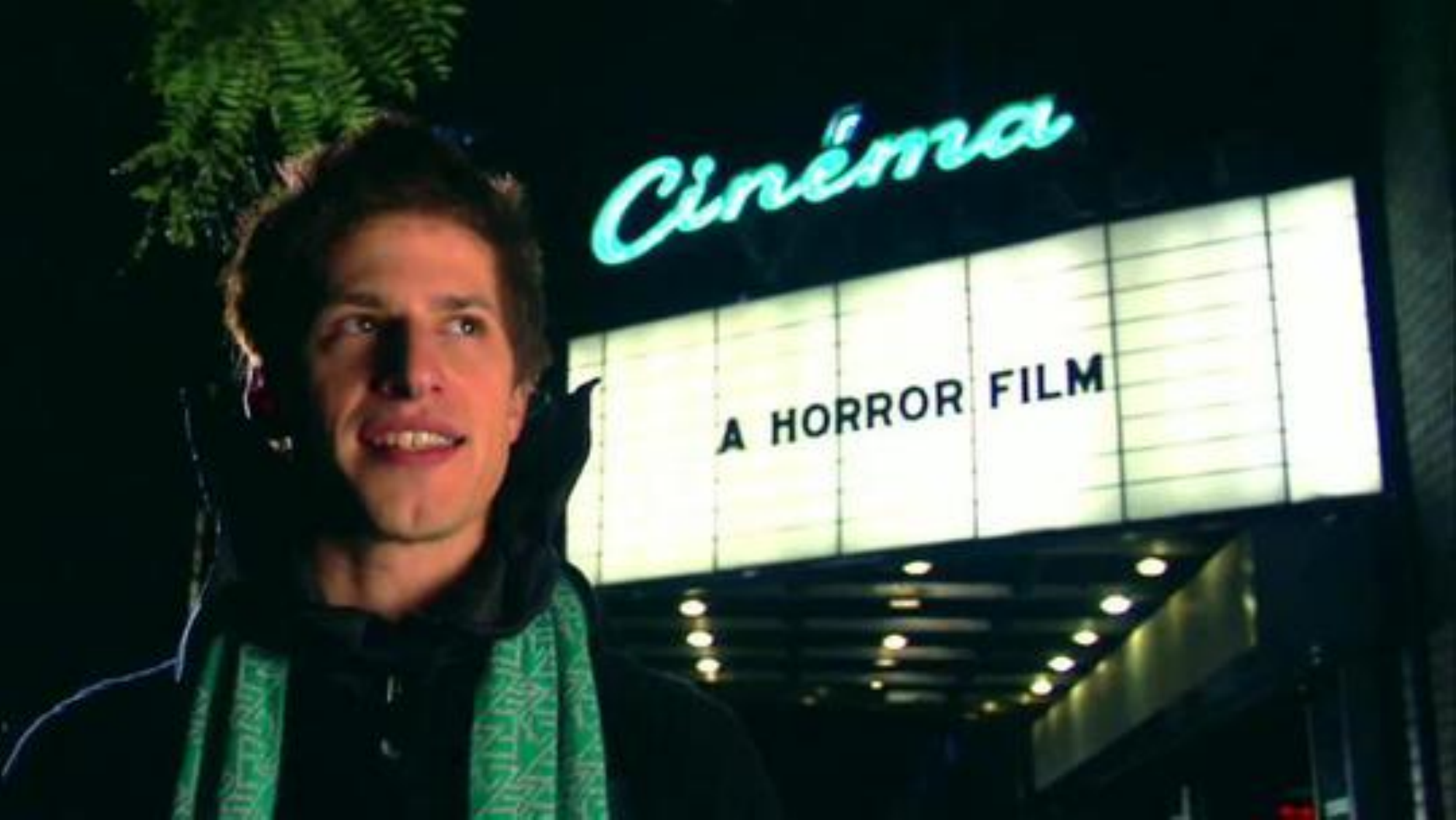}
    \includegraphics[width=.115\hsize]{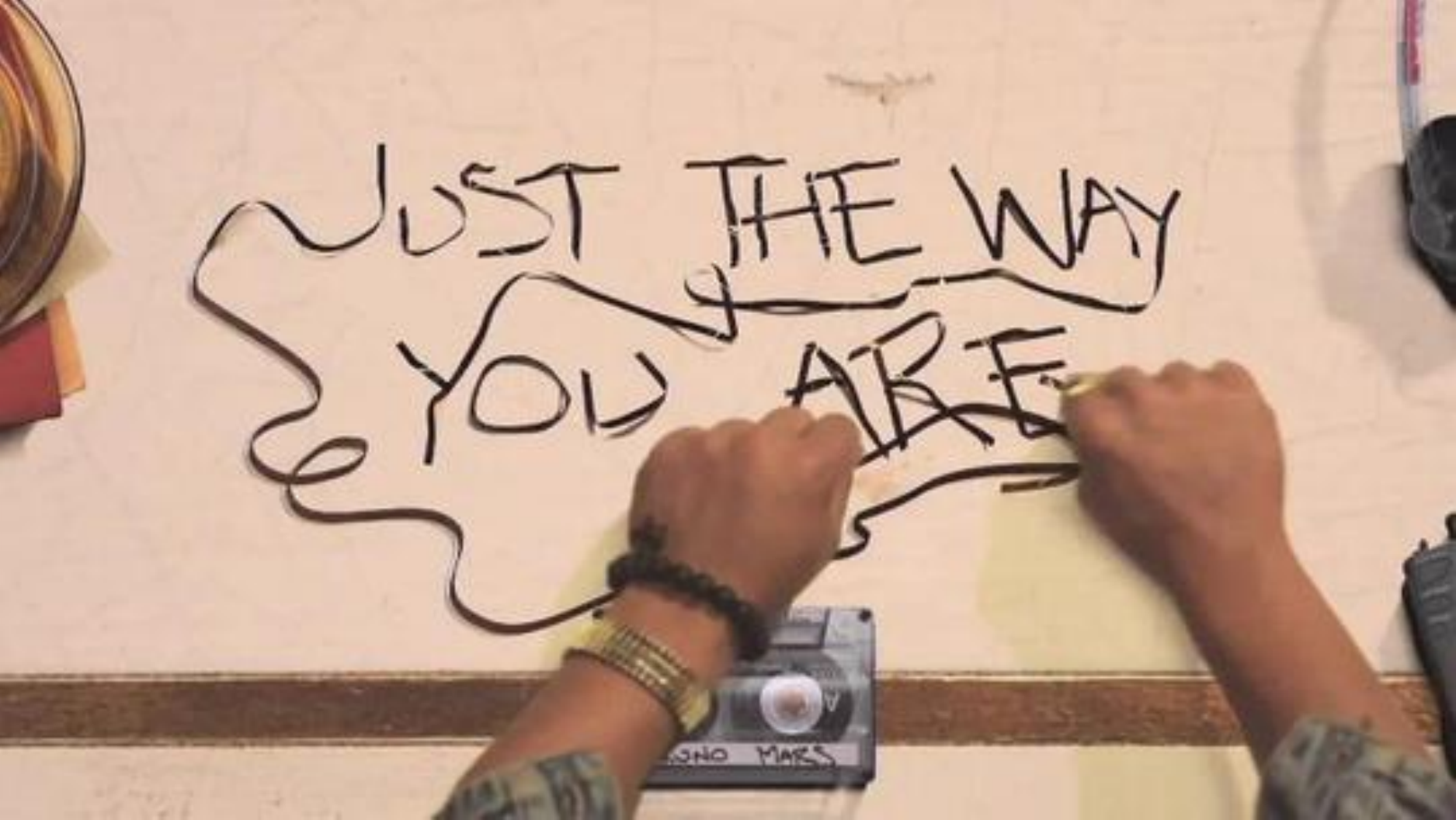}
    \includegraphics[width=.115\hsize]{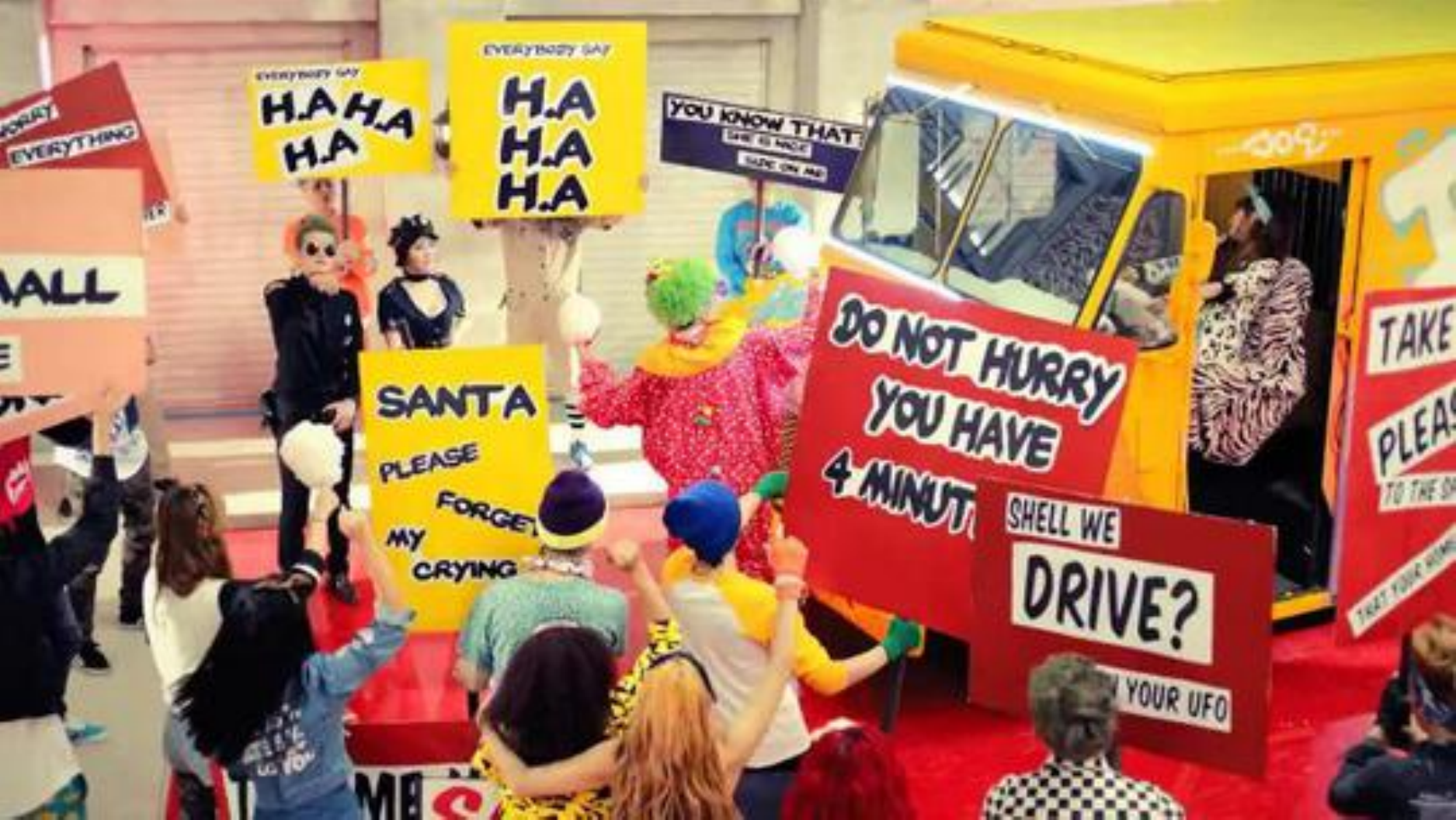}
    \includegraphics[width=.115\hsize]{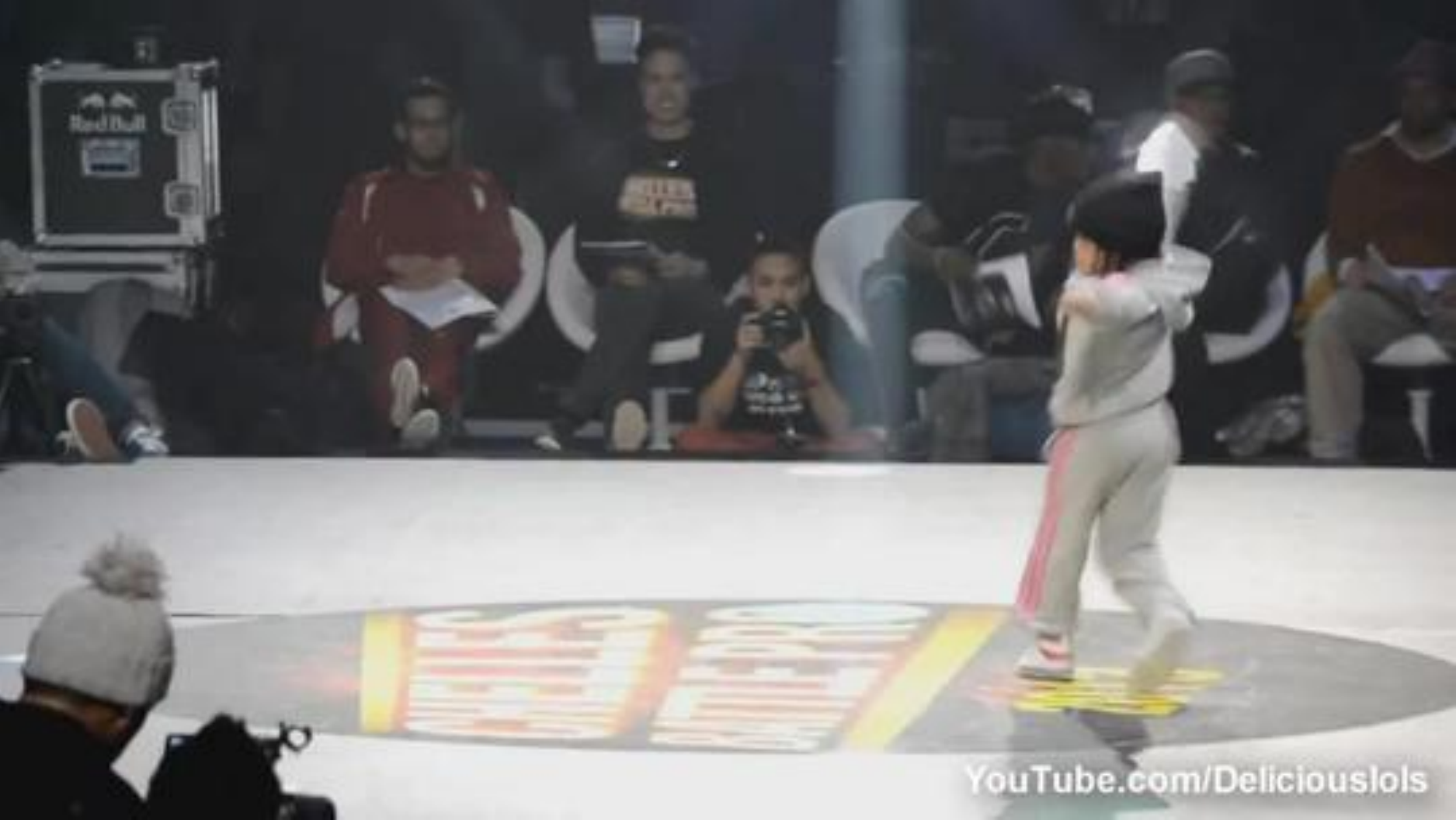}
    \includegraphics[width=.115\hsize]{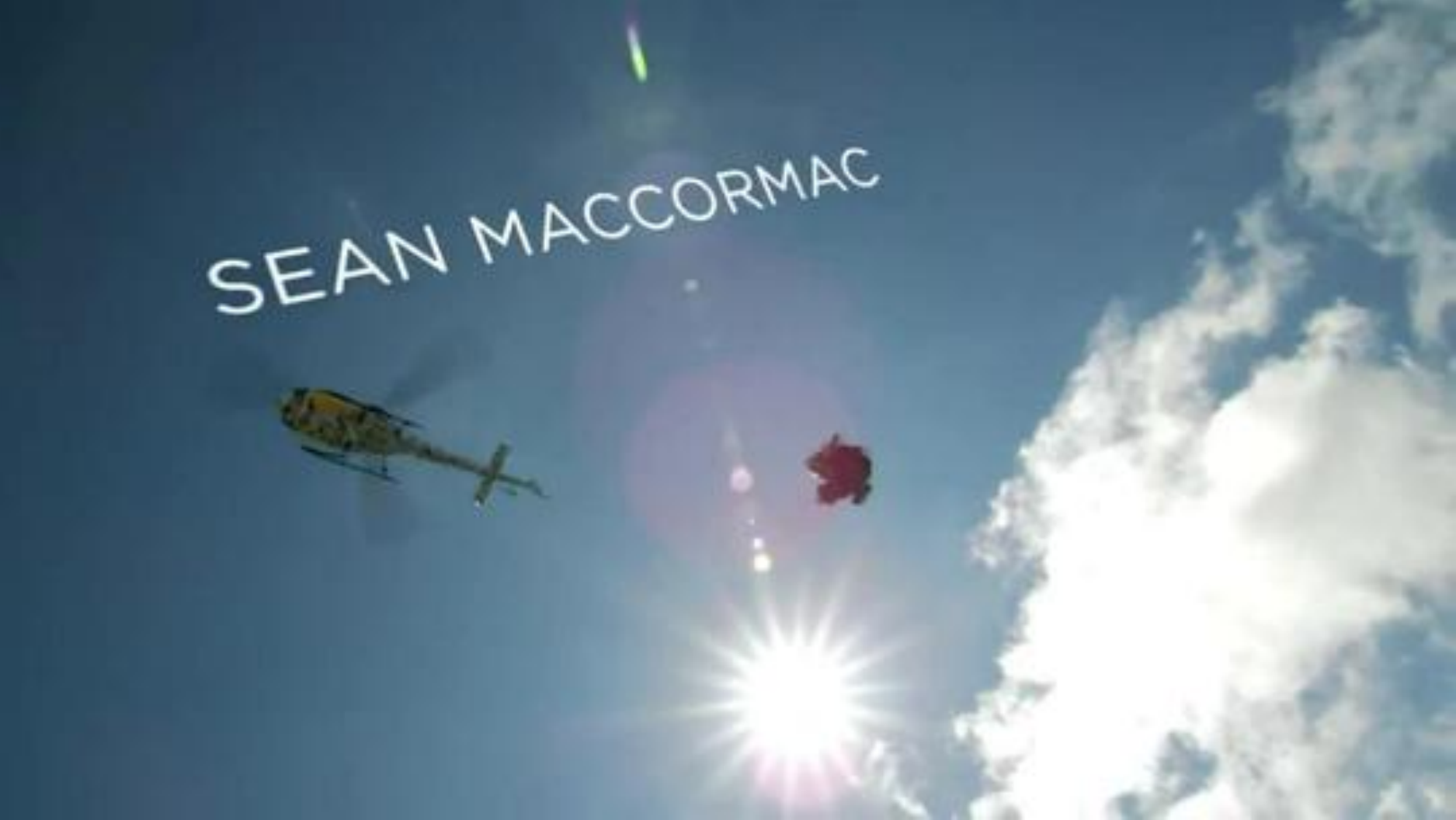}
    \subcaption{YouTube Video Text (YVT) Dataset~\cite{Nguyen_WACV2014}}
    \label{fig:YVT}
  \end{minipage}

  \caption{Sample images of databases \#2.} 
  \label{fig:db_sample2}
\end{figure}


\begin{figure}[tb]
  \centering

  \begin{minipage}[b]{\hsize}
    \centering    
    \includegraphics[width=.115\hsize]{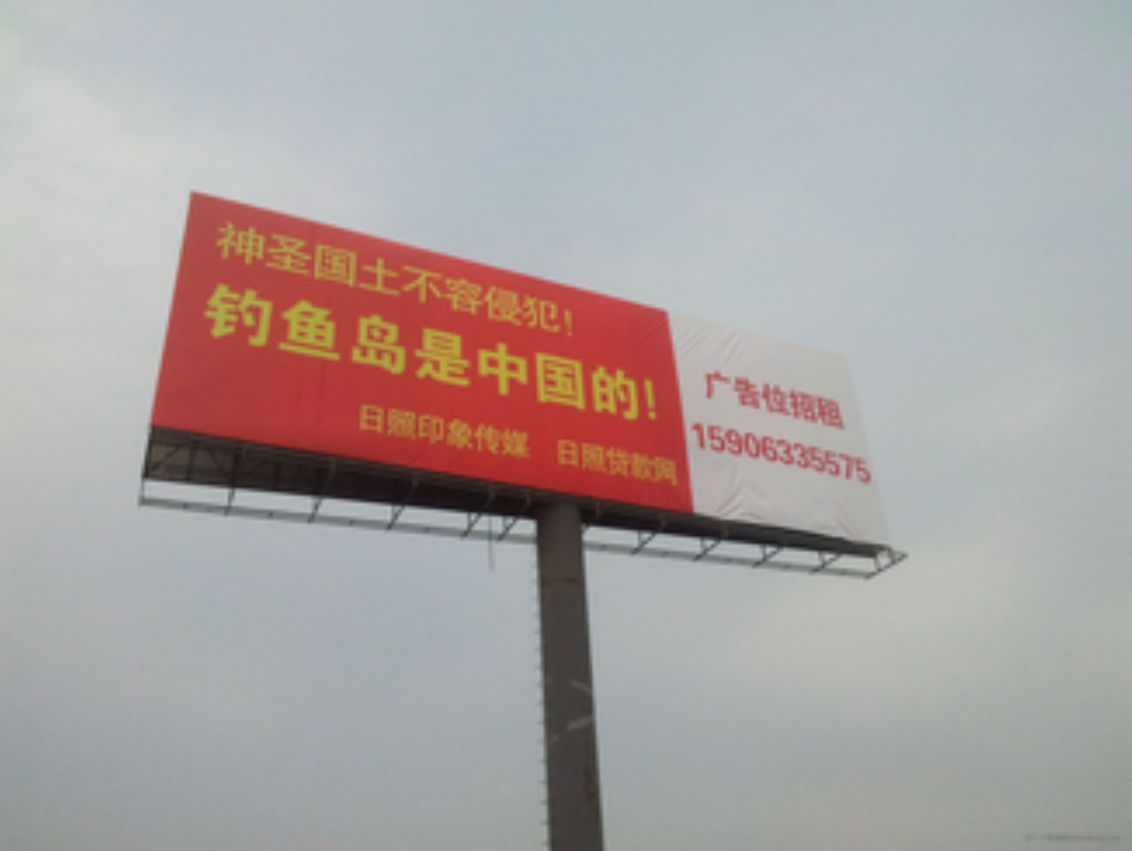}
    \includegraphics[width=.115\hsize]{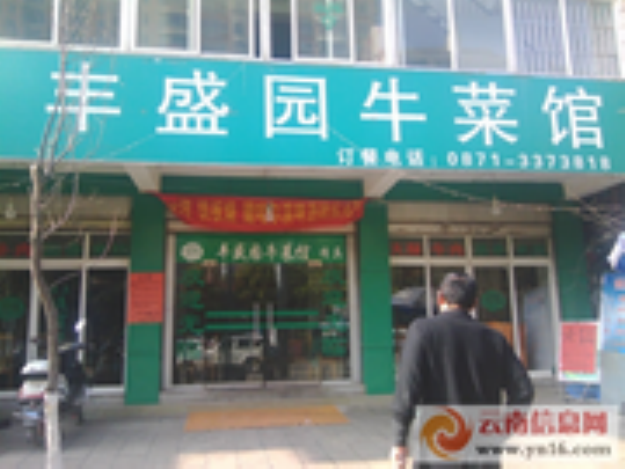}
    \includegraphics[width=.115\hsize]{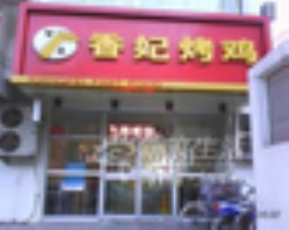}
    \includegraphics[width=.115\hsize]{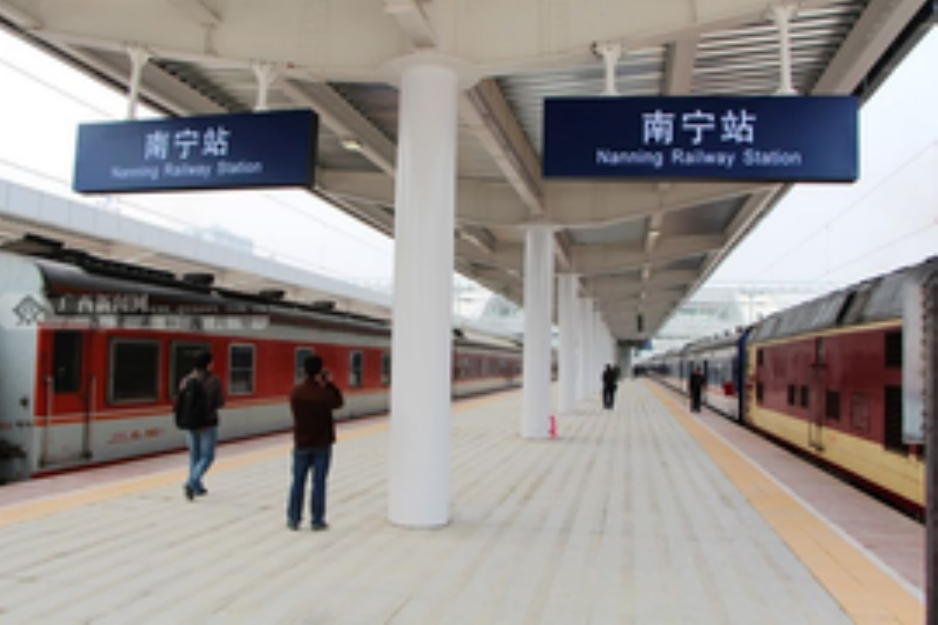}
    \includegraphics[width=.115\hsize]{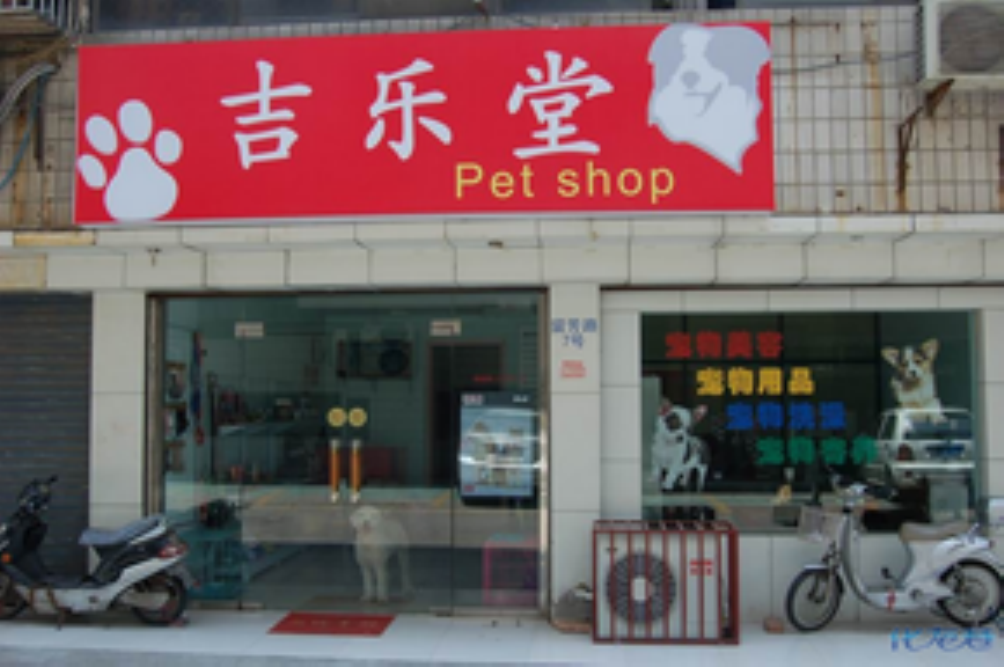}
    \includegraphics[width=.115\hsize]{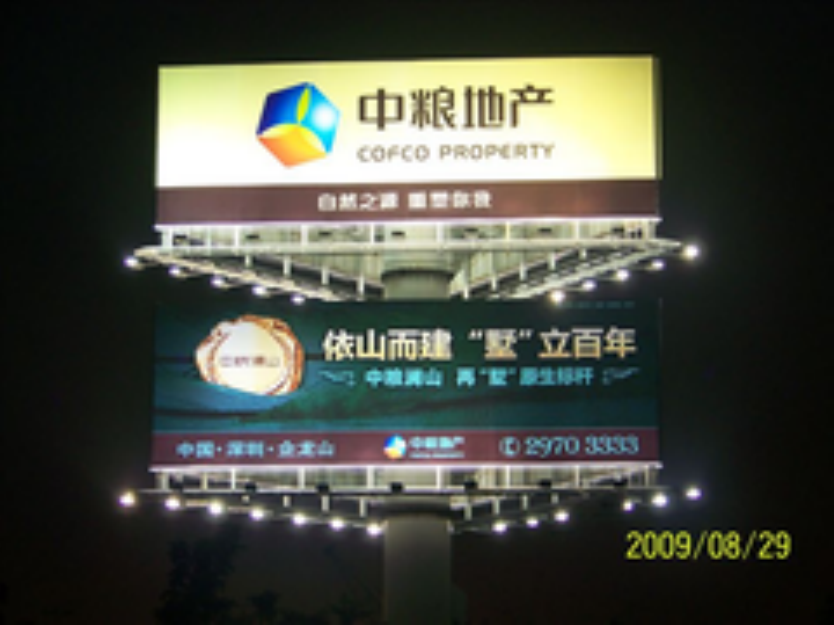}
    \includegraphics[width=.115\hsize]{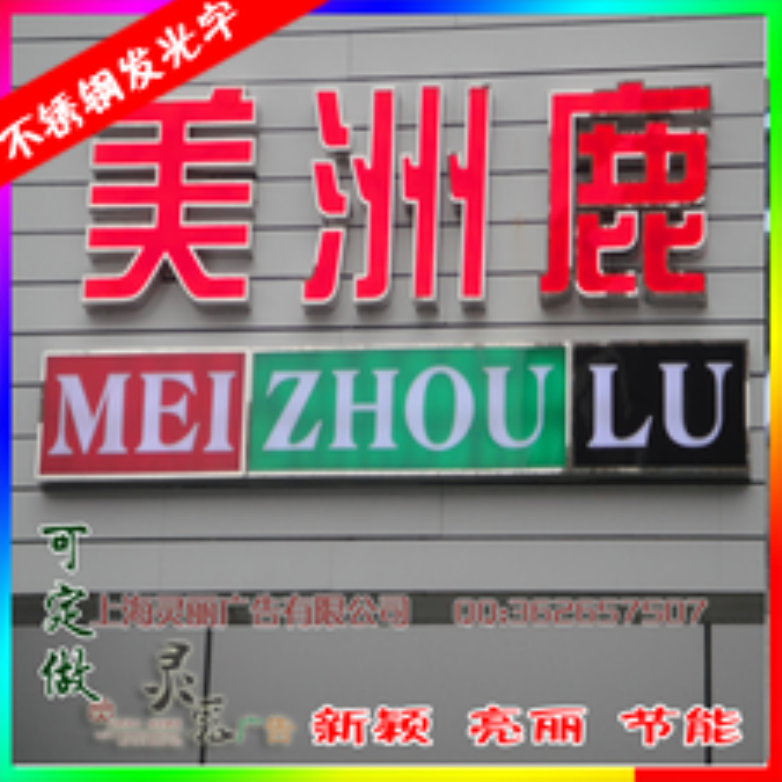}
    \includegraphics[width=.115\hsize]{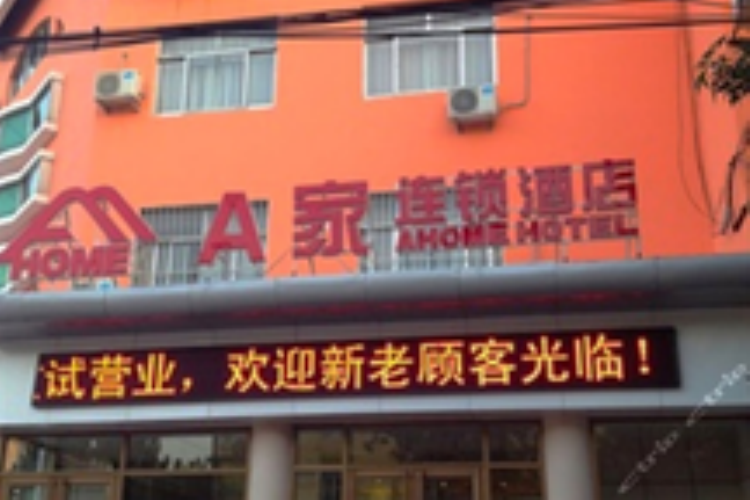}
    \subcaption{ICDAR2015 Competition on Text Reading in the Wild (TRW) Dataset~\cite{ICDAR_TRW2015}}
    \label{fig:ICDAR2015_TRW}
  \end{minipage}
  \begin{minipage}[b]{\hsize}
    \centering    
    \includegraphics[width=.115\hsize]{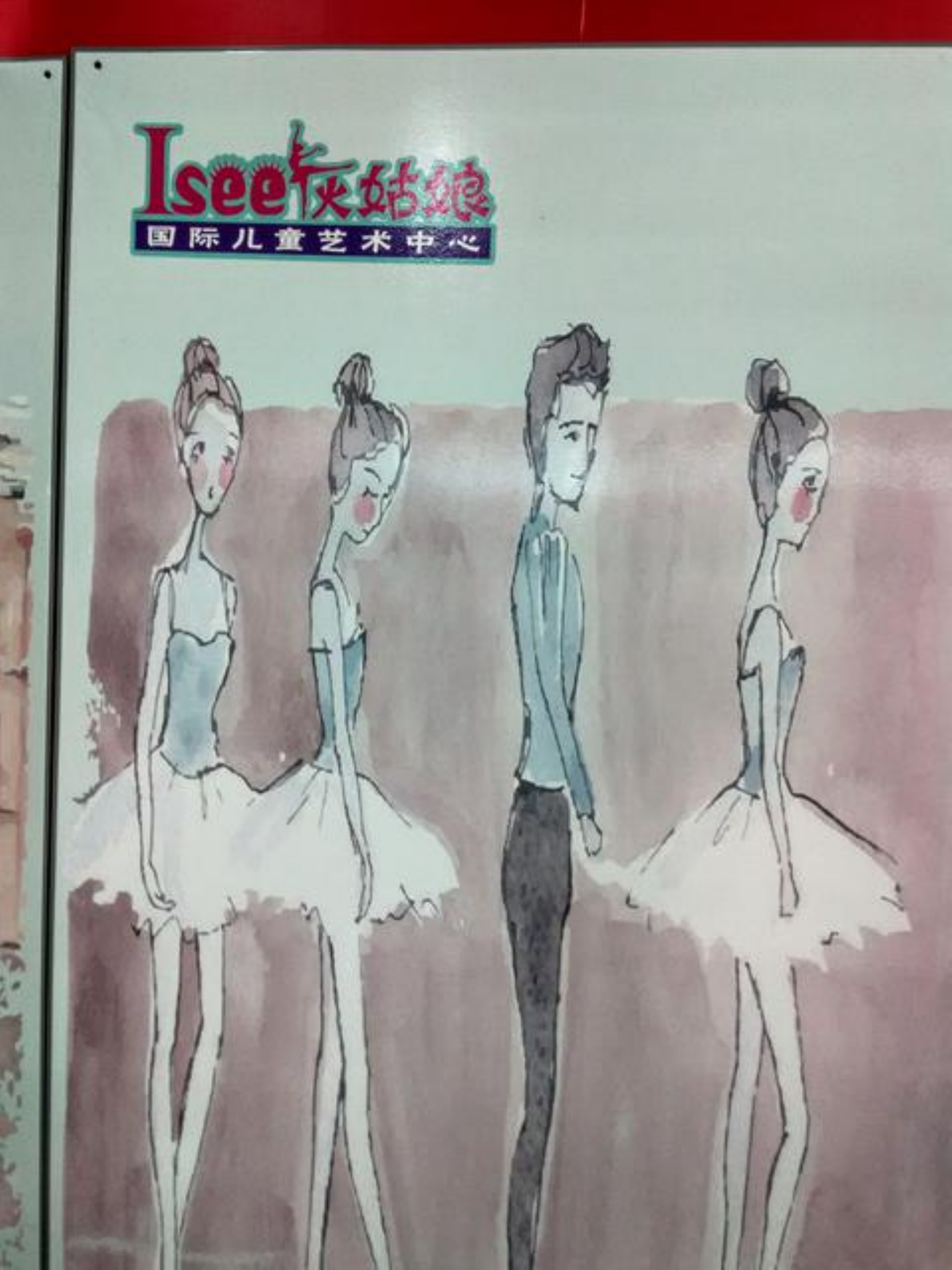}
    \includegraphics[width=.115\hsize]{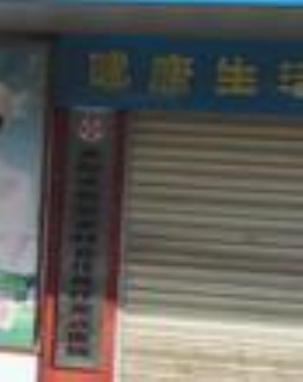}
    \includegraphics[width=.115\hsize]{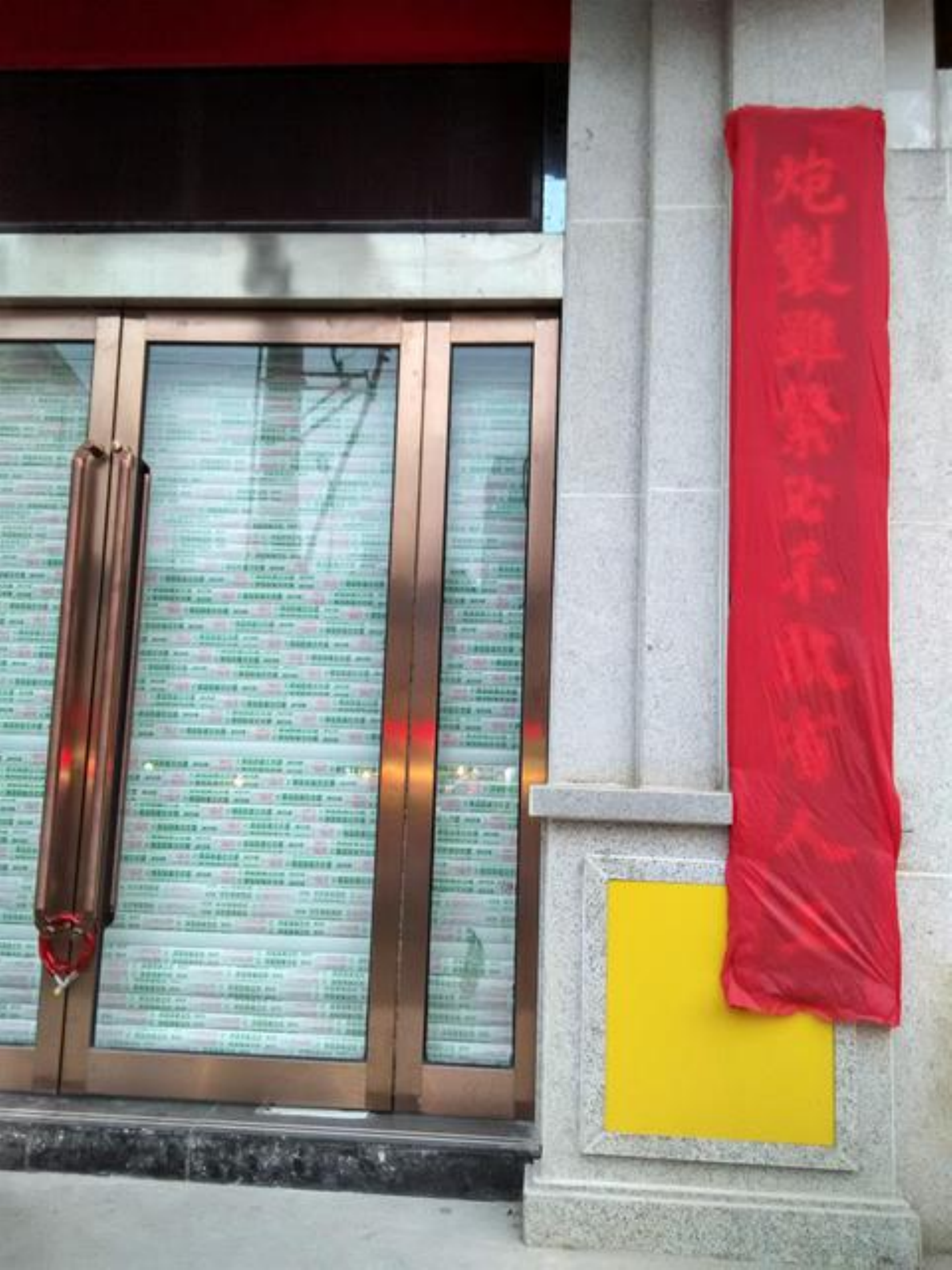}
    \includegraphics[width=.115\hsize]{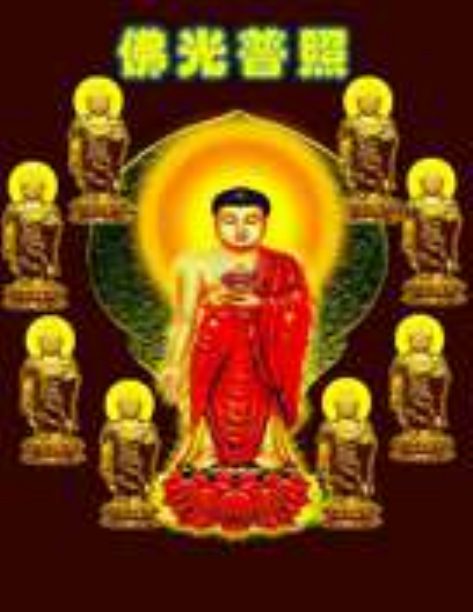}
    \includegraphics[width=.115\hsize]{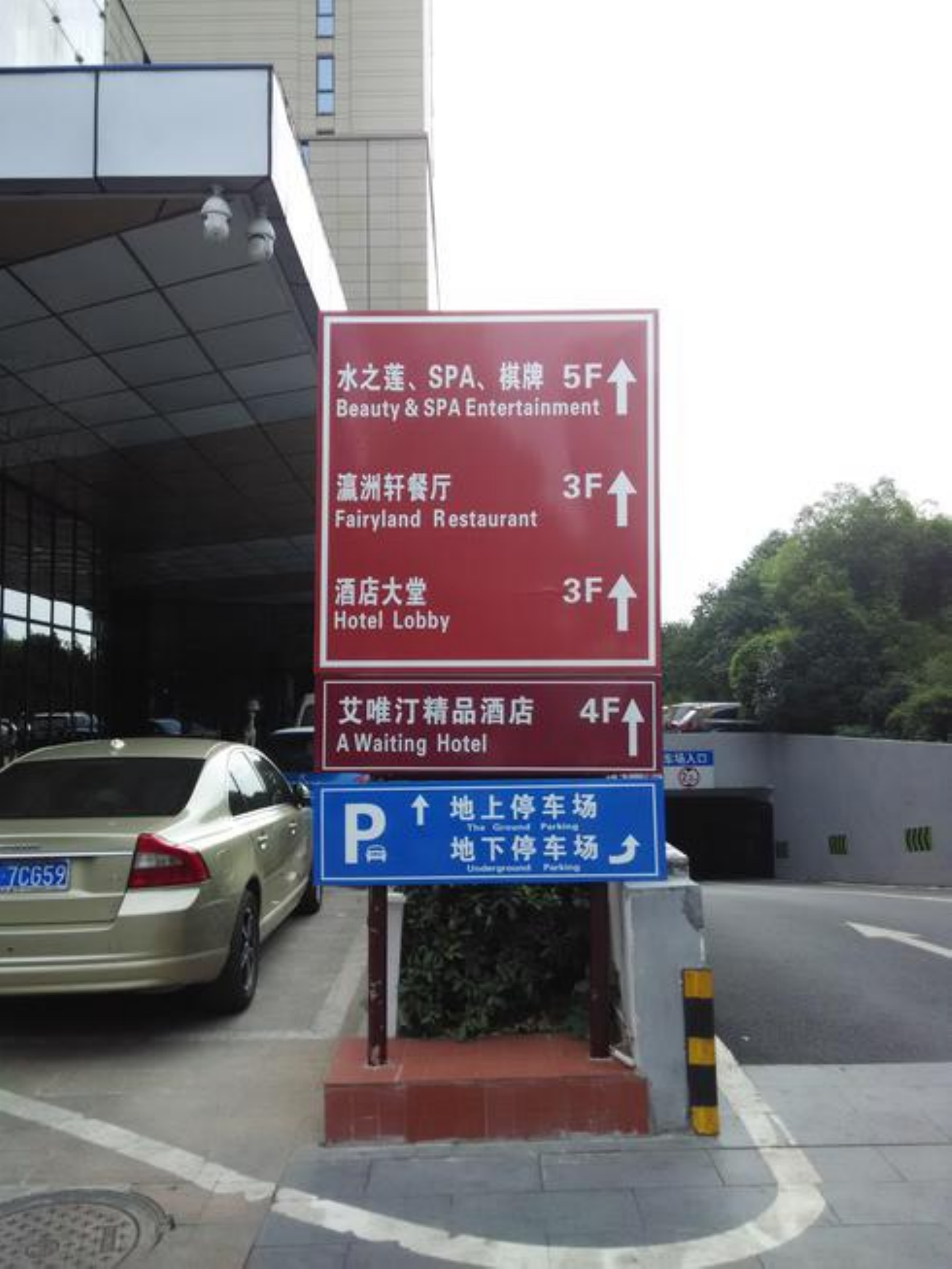}
    \includegraphics[width=.115\hsize]{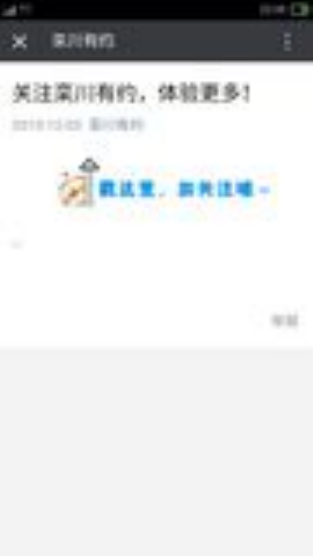}
    \includegraphics[width=.115\hsize]{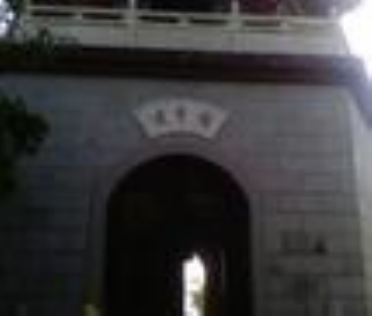}
    \includegraphics[width=.115\hsize]{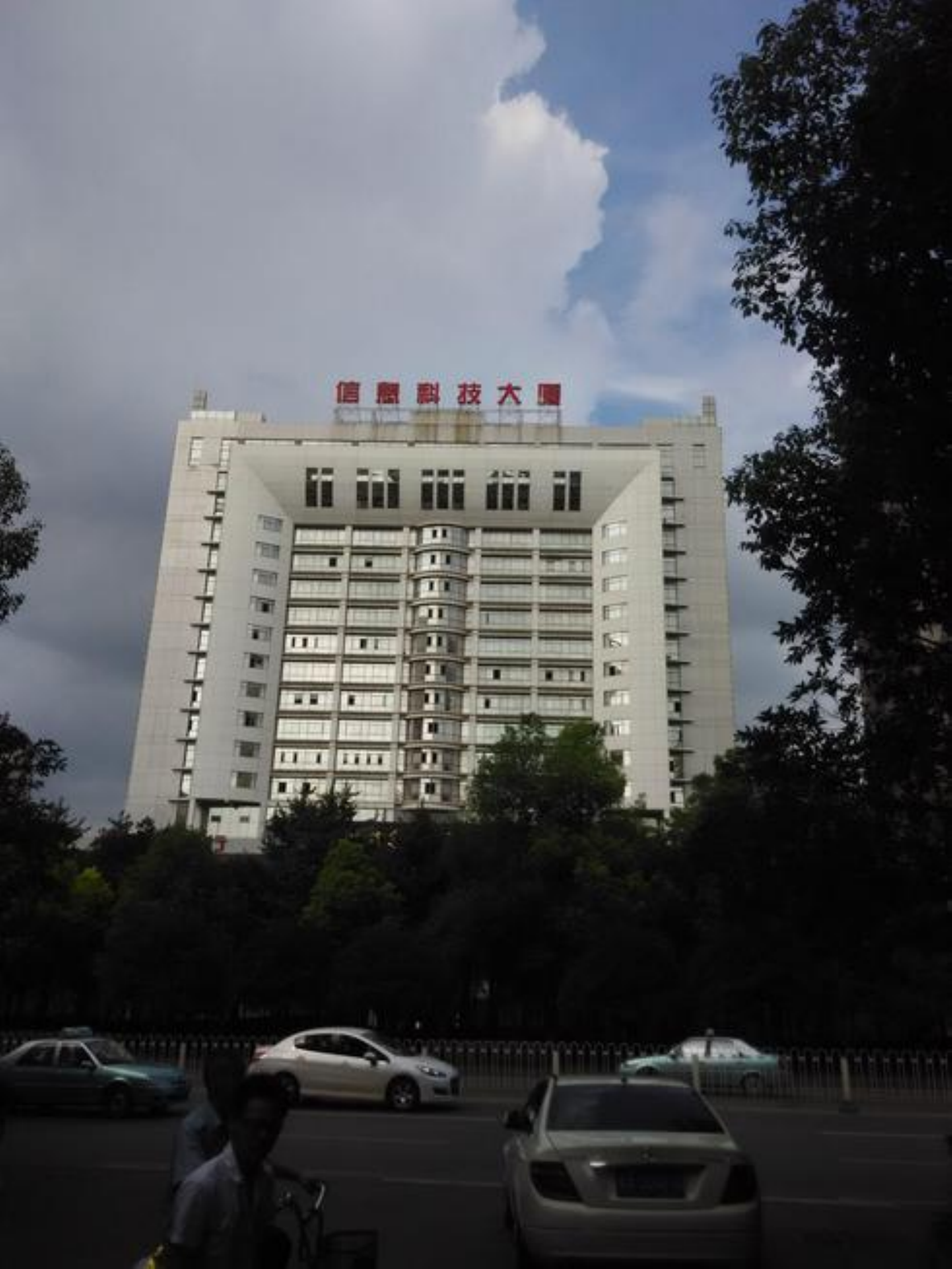}
    \subcaption{ICDAR2017 Competition on Reading Chinese Text in the Wild (RCTW) Dataset~\cite{ICDAR_RCTW17}}
    \label{fig:ICDAR2017_RCTW}
  \end{minipage}
  \begin{minipage}[b]{\hsize}
    \centering    
    \includegraphics[width=.115\hsize]{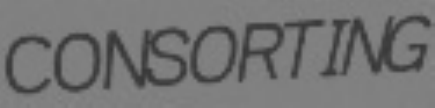}
    \includegraphics[width=.115\hsize]{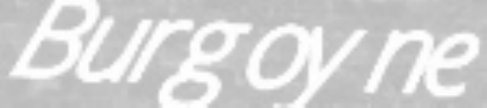}
    \includegraphics[width=.115\hsize]{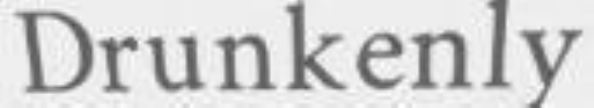}
    \includegraphics[width=.115\hsize]{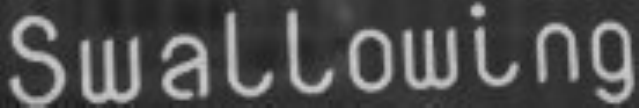}
    \includegraphics[width=.115\hsize]{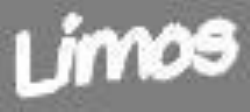}
    \includegraphics[width=.115\hsize]{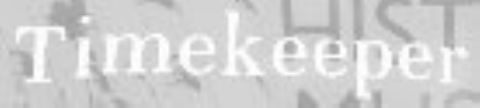}
    \includegraphics[width=.115\hsize]{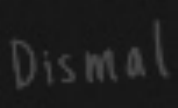}
    \includegraphics[width=.115\hsize]{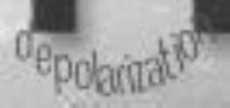}
    \subcaption{MJSynth Dataset~\cite{Jaderberg_NIPS_DLWorkshop2014}}
    \label{fig:MJSynth}
  \end{minipage}
  \begin{minipage}[b]{\hsize}
    \centering    
    \includegraphics[width=.115\hsize]{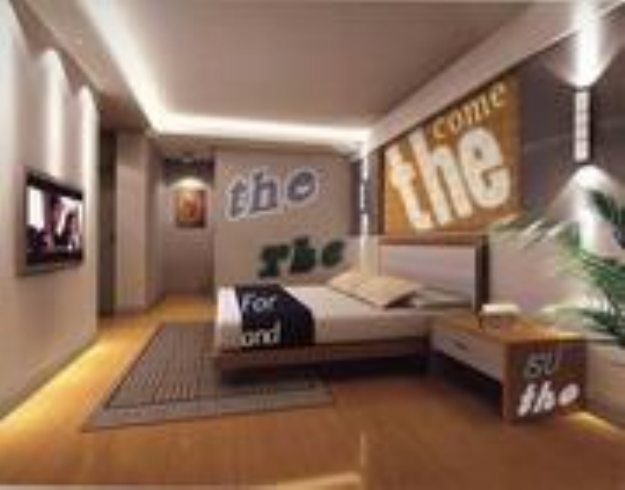}
    \includegraphics[width=.115\hsize]{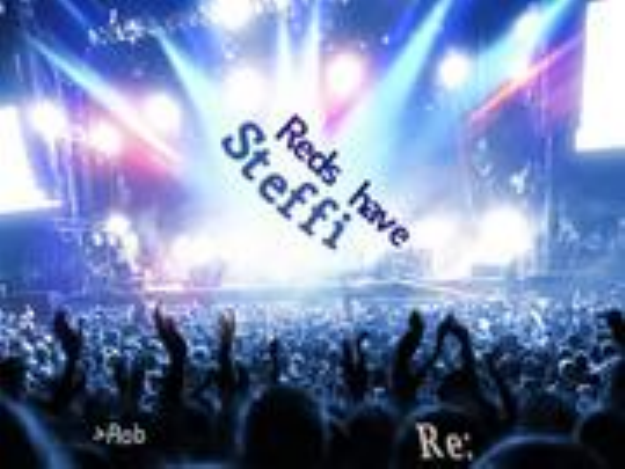}
    \includegraphics[width=.115\hsize]{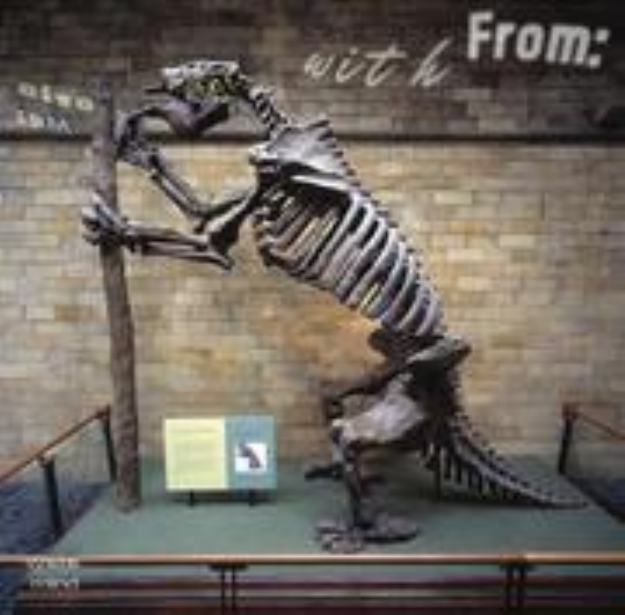}
    \includegraphics[width=.115\hsize]{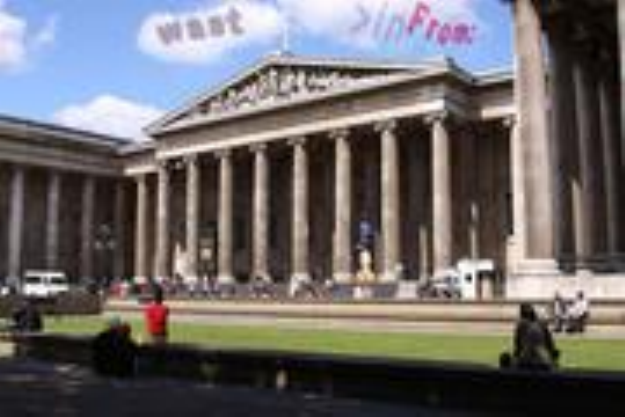}
    \includegraphics[width=.115\hsize]{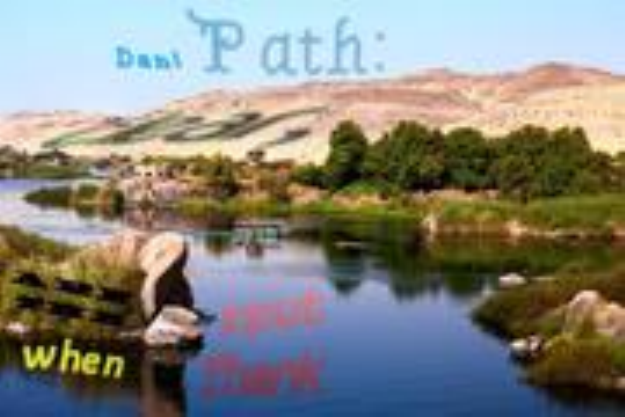}
    \includegraphics[width=.115\hsize]{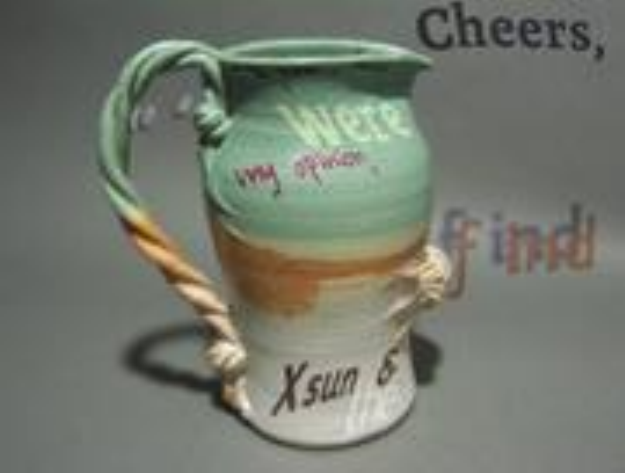}
    \includegraphics[width=.115\hsize]{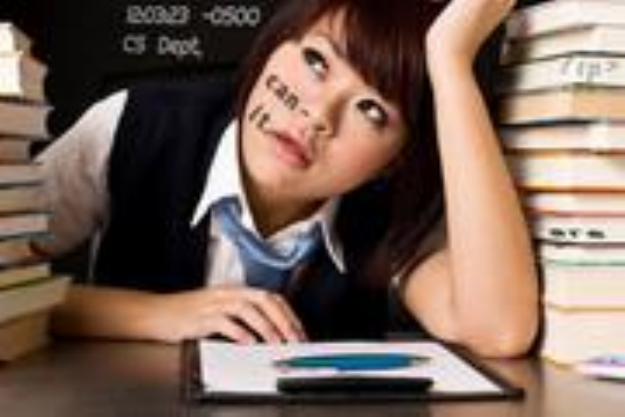}
    \includegraphics[width=.115\hsize]{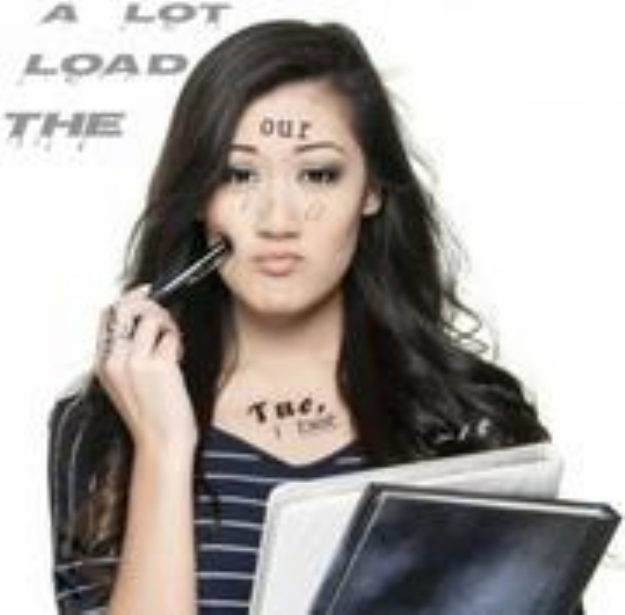}
    \subcaption{SynthText in the Wild Dataset (SynthText)~\cite{Gupta_CVPR2016}}
    \label{fig:SynthText}
  \end{minipage}
  
  \caption{Sample images of databases \#3.} 
  \label{fig:db_sample3}
\end{figure}

\section{Overview of Publicly Available Datasets}

Publicly available datasets are summarized in Table~\ref{tbl:db_summary}.
Their sample images are shown in Figs.~\ref{fig:db_sample1}--\ref{fig:db_sample3}.
They consist of 21 datasets\footnote{The datasets of ICDAR Robust Reading Competitions ``Born Digital Images'' (2011-2015), ``Focused Scene Text'' (2011-2015) and ``Text in Videos'' (2013-2015) are counted as a single dataset for each.}.
Nine of them are related to ICDAR Robust Reading Competitions (2003-2005 and 2011-2015) / Challenges (2017),
ten are other general datasets (out of ten, three focus on character, digit and cropped word images for each), and two are fully synthesized.


The first fully ground-truthed dataset for scene text detection and recognition tasks was provided in 2003 for the first ICDAR RRC~\cite{ICDAR_RRC2003,ICDAR_RRC2003_IJDAR2005}.
The dataset for the scene text detection task contained about 500 images captured with a variety of digital cameras intentionally focusing on words in the images.
Keeping its concept, the dataset was updated in 2011~\cite{ICDAR2011_RRC_challenge2} and 2013~\cite{ICDAR2013_RRC} which are later referred as ICDAR RRC ``Focused Scene Text.''
Though these datasets were used long time as the \textit{de facto standard} for benchmarking, they are almost at the end of their lives.
Primal reasons include their quality and size;
word images of high quality are less challenging to detect and recognize, and 500 images are too small. 

To meet such demands, more challenging datasets have been created.
Street View Text (SVT) dataset~\cite{Wang_ECCV2010,Wang_ICCV2011}, released in 2010, harvests word images from Google Street View.
The word images have variability in appearance and are often low resolution.
Natural Environment OCR (NEOCR) dataset~\cite{Nagy_CBDAR2012}, released in 2011,
provides more challenging text images including blurred, partially occluded, rotated and circularly laid out text.
MSRA Text Detection 500 (MSRA-TD500) database~\cite{Yao_CVPR2012}, released in 2012, contains text images in various angles.
Though the datasets mentioned above contain text images intentionally focused in capturing,
ICDAR RRC ``Incidental Scene Text'' dataset, released in 2015, provides
those captured without intentionally focused.
As a result of not focused, images contained in the dataset are of low quality; they are often out of focus, blurred and low resolution.
The creation of the dataset is encouraged by improvement of imaging technology.
That is, while in the past, word images were assumed to be captured with a digital camera, capturing images with a wearable device become realistic.
COCO-Text dataset~\cite{Veit_arXiv2016,ICDAR_COCO2017}, released in 2016, is
text annotation of MS COCO dataset~\cite{Lin_arXiv2014} constructed for object recognition.
Hence, text in the dataset is not intentionally focused.
Downtown Osaka Scene Text (DOST) dataset~\cite{Iwamura_IWRR2016,ICDAR_DOST2017}, released in 2017, contains sequential images captured with an omni-directional camera.
Use of the omni-directional camera ensures text images are completely free from human intention.
Regarding the dataset size, generally speaking, datasets released more recently contain more data.

Another direction to enhance datasets was to handle scene text in videos (as sequential images).
Compared to static images, videos contain more information.
For example, even if text in a single frame image of a video is hard to read due to blur, we may be able to read it by watching it for a while.
This implies that we can expect more robust detection and recognition of scene text in videos by employing slightly different approaches to those in static images.
ICDAR RRC ``Text in Videos'' dataset~\cite{ICDAR2013_RRC,ICDAR2015_RRC}, released in 2013 and extended in 2015, is the first dataset for scene text detection and recognition in videos.
YouTube Video Text (YVT) dataset~\cite{Nguyen_WACV2014}, released in 2014, harvests image sequences from YouTube videos.
DOST dataset~\cite{Iwamura_IWRR2016,ICDAR_DOST2017} mentioned above is also a video dataset.

While a video is one constructed by aligning static images toward time,
aligning static images toward space yields multiple view images.
French Street Name Signs (FSNS) dataset~\cite{Smith_IWRR2016}, released in 2016, provides French street name signs of up to four views.
In this challenge, similar to video, it is expected to increase recognition performance by using the information contained in the multi-view images.
DOST dataset~\cite{Iwamura_IWRR2016,ICDAR_DOST2017} is also considered as a dataset containing multi-view images.

A recent trend of datasets is to treat scene text of non-English, non-Latin and multiple languages.
Back in 2011, KAIST~\cite{Jung_ETRI2011} and NEOCR~\cite{Nagy_CBDAR2012} datasets containing Korean and German text in addition to English, respectively, are released.
ICDAR RRC ``Text in Videos'' dataset~\cite{ICDAR2013_RRC,ICDAR2015_RRC} contains French and Spanish text in addition to English.
MSRA-TD500~\cite{Yao_CVPR2012} and ICDAR2015 TRW~\cite{ICDAR_TRW2015} datasets contain Chinese and English.
ICDAR2017 RCTW dataset~\cite{ICDAR_RCTW17} contains Chinese only.
FSNS dataset~\cite{Smith_IWRR2016} contains French.
DOST dataset~\cite{Iwamura_IWRR2016,ICDAR_DOST2017} contains Japanese and English.
ICDAR2017 Competition on Multi-lingual Scene Text Detection and Script Identification (MLT) dataset~\cite{ICDAR_MLT2017} contains text of nine languages: Arabic, Bangla, Chinese, English, French, German, Italian, Japanese and Korean.
The tasks include ``joint text detection and script identification'' in addition to text detection and cropped word recognition.

Three datasets focus on character, digit and cropped word images for each.
Chars74k Dataset~\cite{deCampos_VISAPP2009} focusing on character images collects 74k English character images as well as Kannada characters.
Street View House Numbers (SVHN) dataset~\cite{svhn} focusing on digit images collects 630k digits of house numbers from Google Street View.
IIIT5K dataset~\cite{Mishra_BMVC2012} collects 5,000 cropped word images.
In addition, while not treating scene text,
ICDAR RRC ``Born Digital Images'' dataset~\cite{ICDAR2011_RRC_challenge1,ICDAR2013_RRC,ICDAR2015_RRC}, released in 2011, contains text images collected from Web and email images has substantial relationship.

Last but not least, synthesized datasets are expected to play very important roles.
MJSynth dataset~\cite{Jaderberg_NIPS_DLWorkshop2014}, released in 2014, contains 8M cropped word images rendered by a synthetic data engine using 1,400 fonts and variety of combinations of shadow, distortion, coloring and noise.
SynthText in the Wild dataset (SynthText)~\cite{Gupta_CVPR2016}, released in 2016, contains 800k scene text images naturally rendered.
Using these datasets, it is shown that even without real datasets in training, scene text can be detected and recognized very well.


\section{Conclusion and information sources}

This article gave an overview of publicly available datasets in scene text detection and recognition.
Some useful information sources are as follows.
\begin{itemize}
\item ICDAR Robust Reading Competition Portal:\\
\url{http://rrc.cvc.uab.es/}

\item The IAPR TC11 Dataset Repository:\\
\url{http://www.iapr-tc11.org/mediawiki/index.php?title=Datasets}
\end{itemize}

\subsection*{\ackname}
This work is partially supported by JSPS KAKENHI \#17H01803.

\bibliographystyle{splncs}
\bibliography{paper}

\begin{thebibliography}{10}

\bibitem{Jaderberg_NIPS_DLWorkshop2014}
Jaderberg, M., Simonyan, K., Vedaldi, A., Zisserman, A.:
\newblock Synthetic data and artificial neural networks for natural scene text
  recognition.
\newblock In: Proc. NIPS Deep Learning Workshop. (2014)

\bibitem{Gupta_CVPR2016}
Gupta, A., Vedaldi, A., Zisserman, A.:
\newblock Synthetic data for text localisation in natural images.
\newblock In: Proc. IEEE Conference on Computer Vision and Pattern Recognition.
  (2016)

\bibitem{ICDAR2013_RRC}
Karatzas, D., Shafait, F., Uchida, S., Iwamura, M., {Gomez i Bigorda}, L.,
  Mestre, S.R., Mas, J., Mota, D.F., Almazan, J.A., {de las Heras}, L.P.:
\newblock {ICDAR} 2013 robust reading competition.
\newblock In: Proc. International Conference on Document Analysis and
  Recognition. (2013)  1115--1124

\bibitem{ICDAR2015_RRC}
Karatzas, D., Gomez-Bigorda, L., Nicolaou, A., Ghosh, S., Bagdanov, A.,
  Iwamura, M., Matas, J., Neumann, L., Chandrasekhar, V.R., Lu, S., Shafait,
  F., Uchida, S., Valveny, E.:
\newblock {ICDAR} 2015 robust reading competition.
\newblock In: Proc. International Conference on Document Analysis and
  Recognition. (2015)  1156--1160

\bibitem{ICDAR_RRC2003}
Lucas, S.M., Panaretos, A., Sosa, L., Tang, A., Wong, S., Young, R.:
\newblock {ICDAR} 2003 robust reading competitions.
\newblock In: Proc. International Conference on Document Analysis and
  Recognition. Volume~2. (2003)  682--687

\bibitem{ICDAR_RRC2003_IJDAR2005}
Lucas, S.M., Panaretos, A., Sosa, L., Tang, A., Wong, S., Young, R., Ashida,
  K., Nagai, H., Okamoto, M., Yamamoto, H., Miyao, H., Zhu, J., Ou, W., Wolf,
  C., Jolion, J.M., Todoran, L., Worring, M., Lin, X.:
\newblock {ICDAR} 2003 robust reading competitions: Entries, results and future
  directions.
\newblock International Journal on Document Analysis and Recognition
  \textbf{7}(2-3) (2005)  105--122

\bibitem{ICDAR_RRC2005}
Lucas, S.M.:
\newblock {ICDAR} 2005 text locating competition results.
\newblock In: Proc. International Conference on Document Analysis and
  Recognition. Volume~1. (2005)  80--84

\bibitem{Wolf_IJDAR2006}
Wolf, C., Jolion, J.M.:
\newblock Object count/area graphs for the evaluation of object detection and
  segmentation algorithms.
\newblock International Journal of Document Analysis and Recognition
  \textbf{8}(4) (September 2006)  280--296

\bibitem{Everingham_IJCV2014}
Everingham, M., Eslami, S.M.A., Gool, L.V., Williams, C.K.I., Winn, J.,
  Zisserman, A.:
\newblock The pascal visual object classes challenge: A retrospective.
\newblock International Journal of Computer Vision \textbf{111}(1) (June 2014)
  98--136

\bibitem{Bernardin_EJIVP2008}
Bernardin, K., Stiefelhagen, R.:
\newblock Evaluating multiple object tracking performance: The clear mot
  metrics.
\newblock EURASIP Journal on Image and Video Processing \textbf{2008} (May
  2008)

\bibitem{Kasturi_TPAMI2009}
Kasturi, R., Goldgof, D., Soundararajan, P., Manohar, V., Garofolo, J., Bowers,
  R., Boonstra, M., Korzhova, V., Zhang, J.:
\newblock Framework for performance evaluation of face, text, and vehicle
  detection and tracking in video: Data, metrics, and protocol.
\newblock IEEE Transactions on Pattern Analysis and Machine Intelligence
  \textbf{31}(2) (2009)  319--336

\bibitem{Nguyen_WACV2014}
Nguyen, P.X., Wang, K., Belongie, S.:
\newblock Video text detection and recognition: Dataset and benchmark.
\newblock In: Proc. IEEE Winter Conference on Applications of Computer Vision.
  (2014)

\bibitem{ICDAR2011_RRC_challenge1}
Karatzas, D., Mestre, S.R., Mas, J., Nourbakhsh, F., Roy, P.P.:
\newblock {ICDAR} 2011 robust reading competition challenge 1: Reading text in
  born-digital images (web and email).
\newblock In: Proc. International Conference on Document Analysis and
  Recognition. (2011)  1485--1490

\bibitem{Jung_ETRI2011}
Jung, J., Lee, S., Cho, M.S., Kim, J.H.:
\newblock Touch {TT}: Scene text extractor using touchscreen interface.
\newblock ETRI Journal \textbf{33}(1) (2011)  78--88

\bibitem{Clavelli_DAS2010}
Clavelli, A., Karatzas, D., Llad{\'{o}}s, J.:
\newblock A framework for the assessment of text extraction algorithms on
  complex colour images.
\newblock In: Proc. International Workshop on Document Analysis Systems. (2010)

\bibitem{Wang_ICCV2011}
Wang, K., Babenko, B., Belongie, S.:
\newblock End-to-end scene text recognition.
\newblock In: Proc. International Conference on Computer Vision. (2011)
  1457--1464

\bibitem{ICDAR2011_RRC_challenge2}
Shahab, A., Shafait, F., Dengel, A.:
\newblock {ICDAR} 2011 robust reading competition challenge 2: Reading text in
  scene images.
\newblock In: Proc. International Conference on Document Analysis and
  Recognition. (2011)  1491--1496

\bibitem{Veit_arXiv2016}
Veit, A., Matera, T., Neumann, L., Matas, J., Belongie, S.:
\newblock {COCO-Text}: Dataset and benchmark for text detection and recognition
  in natural images.
\newblock \texttt{arXiv:1601.07140 [cs.CV]} (2016)

\bibitem{ICDAR_COCO2017}
Gomez, R., Shi, B., Gomez, L., Neumann, L., Veit, A., Matas, J., Belongie, S.,
  Karatzas, D.:
\newblock {ICDAR}2017 robust reading challenge on {COCO-Text}.
\newblock In: Proc. International Conference on Document Analysis and
  Recognition. (2017)

\bibitem{Lin_arXiv2014}
Lin, T.Y., Maire, M., Belongie, S., Bourdev, L., Girshick, R., Hays, J.,
  Perona, P., Ramanan, D., Zitnick, C.L., Doll^^c3^^a1r, P.:
\newblock Microsoft coco: Common objects in context.
\newblock \texttt{ arXiv:1405.0312 [cs.CV]} (2014)

\bibitem{Smith_IWRR2016}
Smith, R., Gu, C., Lee, D.S., Hu, H., Unnikrishnan, R., Ibarz, J., Arnoud, S.,
  Lin, S.:
\newblock End-to-end interpretation of the french street name signs dataset.
\newblock In: Proc. International Workshop on Robust Reading. (2016)  411--426

\bibitem{Iwamura_IWRR2016}
Iwamura, M., Matsuda, T., Morimoto, N., Sato, H., Ikeda, Y., Kise, K.:
\newblock Downtown osaka scene text dataset.
\newblock In: Proc. International Workshop on Robust Reading. (2016)  440--455

\bibitem{ICDAR_DOST2017}
Iwamura, M., Morimoto, N., Tainaka, K., Bazazian, D., Gomez, L., Karatzas, D.:
\newblock {ICDAR}2017 robust reading challenge on omnidirectional video.
\newblock In: Proc. International Conference on Document Analysis and
  Recognition. (2017)

\bibitem{ICDAR_MLT2017}
Nayef, N., Yin, F., Bizid, I., Choi, H., Feng, Y., Karatzas, D., Luo, Z., Pal,
  U., Rigaud, C., Chazalon, J., Khlif, W., Luqman, M.M., Burie, J.C., lin Liu,
  C., Ogier, J.M.:
\newblock {ICDAR}2017 robust reading challenge onmulti-lingual scene text
  detection and scriptidentification ^^e2^^80^^93 {RRC-MLT}.
\newblock In: Proc. International Conference on Document Analysis and
  Recognition. (2017)

\bibitem{deCampos_VISAPP2009}
de~Campos, T.E., Babu, B.R., Varma, M.:
\newblock Character recognition in natural images.
\newblock In: Proc. International Conference on Computer Vision Theory and
  Applications. (2009)

\bibitem{Wang_ECCV2010}
Wang, K., Belongie, S.:
\newblock Word spotting in the wild.
\newblock In: Proc. European Conference on Computer Vision: Part I. (2010)
  591--604

\bibitem{Nagy_CBDAR2012}
Nagy, R., Dicker, A., Meyer-Wegener, K.:
\newblock {NEOCR}: A configurable dataset for natural image text recognition.
\newblock In: Camera-Based Document Analysis and Recognition. Volume 7139 of
  Lecture Notes in Computer Science. (2012)  150--163

\bibitem{svhn}
Netzer, Y., Wang, T., Coates, A., Bissacco, A., Wu, B., Ng, A.Y.:
\newblock Reading digits in natural images with unsupervised feature learning.
\newblock In: Proc. NIPS Workshop on Deep Learning and Unsupervised Feature
  Learning. (2011)

\bibitem{Yao_CVPR2012}
Yao, C., Bai, X., Liu, W., Ma, Y., Tu, Z.:
\newblock Detecting texts of arbitrary orientations in natural images.
\newblock In: Proc. IEEE Conference on Computer Vision and Pattern Recognition.
  (2012)  1083--1090

\bibitem{Mishra_BMVC2012}
Mishra, A., Alahari, K., Jawahar, C.V.:
\newblock Scene text recognition using higher order language priors.
\newblock In: Proc. British Machine Vision Conference. (2012)

\bibitem{ICDAR_TRW2015}
Zhou, X., Zhou, S., Yao, C., Cao, Z., Yin, Q.:
\newblock {ICDAR} 2015 text reading in the wild competition.
\newblock arXiv preprint (2015)

\bibitem{ICDAR_RCTW17}
Shi, B., Yao, C., Liao, M., Yang, M., Xu, P., Cui, L., Belongie, S., Lu, S.,
  Bai, X.:
\newblock {ICDAR}2017 competition on reading chinesetext in the wild
  ({RCTW}-17).
\newblock In: Proc. International Conference on Document Analysis and
  Recognition. (2017)

\end{thebibliography}
\end{document}